%% file: main.tex
\documentclass[11pt]{article}

\usepackage[margin=1in]{geometry}

\usepackage{authblk}
\usepackage{amsmath,amssymb,amsthm}
\usepackage{mathtools}
\usepackage{nicefrac}
\usepackage{stmaryrd}
\SetSymbolFont{stmry}{bold}{U}{stmry}{m}{n}
\usepackage{microtype}
\usepackage[inline]{enumitem}
\usepackage{graphicx}
\usepackage{booktabs}
\usepackage{tikz}
\usepackage{wrapfig}
\usepackage{thmtools}
\usepackage[numbers,sort&compress]{natbib}
\usepackage[colorlinks=true,allcolors=blue!60!black]{hyperref}
\usepackage{cleveref}
\setcitestyle{numbers,square,comma,sort&compress}
\usetikzlibrary{arrows.meta,decorations.pathreplacing,positioning,
                calc,angles,intersections,patterns}

\usepackage{amartya_ltx}

\newcommand{\OurDimName}{Independent Class Packing}
\newcommand{\OurDimAcronym}{ICP}

\newcommand{\icp}{\ensuremath{\textsc{\OurDimAcronym}}}

\makeatletter
\@for\@thmname:=thm,defn,lem,corollary,claim,proposition,remark,example,%
  question,inequality,assumption\do{%
  \@for\@thmpre:=,end\do{%
    \expandafter\let\csname\@thmpre\@thmname\endcsname\@undefined}%
  \expandafter\let\csname c@\@thmname\endcsname\@undefined
  \expandafter\let\csname the\@thmname\endcsname\@undefined}
\makeatother

\newtheorem{thm}{Theorem}
\newtheorem{defn}{Definition}
\newtheorem{lem}{Lemma}
\newtheorem{corollary}{Corollary}
\newtheorem{claim}{Claim}

\newtheorem{remark}{Remark}

\usepackage{hyperref}
\hypersetup{
  colorlinks=true,
  linkcolor=blue,
  filecolor=magenta,
  urlcolor=cyan,
  citecolor=blue
}
\crefname{thm}{theorem}{theorems}              \Crefname{thm}{Theorem}{Theorems}
\crefname{lem}{lemma}{lemmas}                  \Crefname{lem}{Lemma}{Lemmas}
\crefname{defn}{definition}{definitions}       \Crefname{defn}{Definition}{Definitions}
\crefname{proposition}{proposition}{propositions}
\Crefname{proposition}{Proposition}{Propositions}
\crefname{corollary}{corollary}{corollaries}
\Crefname{corollary}{Corollary}{Corollaries}
\crefname{claim}{claim}{claims}                \Crefname{claim}{Claim}{Claims}
\crefname{remark}{remark}{remarks}             \Crefname{remark}{Remark}{Remarks}
\crefname{example}{example}{examples}          \Crefname{example}{Example}{Examples}
\crefname{assumption}{assumption}{assumptions}
\Crefname{assumption}{Assumption}{Assumptions}
\crefname{question}{question}{questions}       \Crefname{question}{Question}{Questions}
\crefname{inequality}{inequality}{inequalities}
\Crefname{inequality}{Inequality}{Inequalities}

\title{New Lower Bounds for Private PAC Learning}

\author[1]{Prathamesh Dharangutte }
\author[1]{Amartya Sanyal}
\affil[1]{Department of Computer Science, University of Copenhagen}
\affil[ ]{\texttt{\{prathamesh.dharangutte, amsa\}@di.ku.dk}}

\begin{document}

\date{}
\maketitle

\begin{abstract}

We study realizable PAC learning under approximate differential privacy, and give a framework for proving lower bounds on its sample complexity. We construct classes with identical VC and Littlestone dimensions but different private sample complexities at common privacy and accuracy
parameters, showing that these dimensions do not determine the sample
complexity, even for improper learners. To address this we introduce the Independent Class Packing ($\icp$) parameter, the largest total sample complexity of classes in $\cG$ that can be packed independently into a class $\cH$, and prove that every $(\epsilon, \delta)$-DP, $(\alpha, \beta)$-PAC learner for $\cH$, possibly improper, uses $\Om{\icp(\cH, \cG)/(\alpha\epsilon)}$ examples. Using this framework we prove tight lower bounds for axis-aligned rectangles, union of intervals and linear subspaces and improved lower bounds for halfspaces and convex polygons. Finally we also show that for product classes, and more generally when the same classes
pack into $\cH$ and cover its domain so that target labels
can be used directly, $\icp$ characterizes private PAC
sample complexity up to logarithmic factors.

\end{abstract}

\section{Introduction}
\label{sec:introduction}
\input{contents/01_introduction}

\subsection{Related Work}
\label{sec:related-work}
\input{contents/02_related_work}

\input{appendix/prelim}

\section{Same Dimensions, Different Private Sample Complexity}
\label{sec:classes}
\input{contents/03_product_union_classes}

\input{contents/independent_class_packings}

\input{contents/threshold_packing}

\input{contents/tree_packing}

\input{contents/upper_bounds}

\section*{Acknowledgements}

The authors repeatedly interacted with Claude Opus 5.5 and GPT-6 Astra in verifying and polishing the proofs and drawing tikz figures.
The authors take full responsibility for all the content in this work. Prathamesh Dharangutte and Amartya Sanyal are supported by Villum Fonden via the Young Investigator program (72069). Amartya Sanyal also acknowledges the Novo Nordisk Foundation for support via the Startup
grant (NNF24OC0087820).

\bibliography{references}

\appendix

\section{Proofs for \Cref{sec:classes}}
\label{app:classes}
\input{appendix/icp_product_union}

\input{appendix/compare_icp}

\input{appendix/independent_class_packings_proofs}

\section{Proofs for \Cref{sec:packing_thresholds}}
\label{app:threshold-lower-bound}
\input{appendix/04_threshold_lower_bound_proofs}

\input{appendix/upper_bound_proofs}
\input{appendix/interval_upper_bound}

\end{document}

%% file: contents/01_introduction.tex
A learning algorithm trained using sensitive data should not reveal
much about any individual example. Differential privacy~(DP)
\citep{DworkKMMN2006,DworkEtAl2006} makes this requirement precise;
private PAC learning~\citep{KasiviswanathanEtAl2011} asks how many
examples a PAC learner needs when its output must also satisfy DP.
Privacy can make learning substantially more expensive, and a central
question is which properties of a hypothesis class determine this
cost. Without privacy, the sample complexity of PAC learning is
characterised by the VC dimension \citep{BlumerEtAl1989}. In their
seminal result,~\citet{AlonEtAl2022} showed that under approximate
differential privacy, a class is learnable if and only if its
Littlestone dimension is finite. This settled the qualitative question
of when a hypothesis class is privately learnable. However, the known
upper and lower bounds on the private sample complexity in terms of
the VC and Littlestone dimensions are far apart.

We study realizable PAC learning~\citep{Valiant1984} and allow improper
learners. Let $\cH \subseteq
\bc{0,1}^{\cX}$. An $n$-sample learner receives \(n\) i.i.d. examples
$\bc{\br{x_i, h^\star(x_i)}}_{i=1}^n$, where $h^\star \in \cH$, and
outputs a Boolean function. It is an $\br{\alpha, \beta}$-PAC learner
if its population error is at most $\alpha$ with probability at least
$1 - \beta$ uniformly over all $h^\star$ and all distributions. It is
$\br{\epsilon, \delta}$-DP if, for all samples $S, S'$ of size \(n\)
that differ in one example, and every event $E$,
\[
\Pr\bs{\cM(S) \in E} \leq e^{\epsilon}\Pr\bs{\cM(S') \in E} + \delta.
\]

We write $\mpriv{\cH}{\epsilon}{\delta}{\alpha}{\beta}$ for the
smallest sample size a learner satisfying both conditions needs. Let
$v = \VC(\cH)$ and $D = \ldim(\cH)$, and let $\logstar
x$ be the iterated logarithm \emph{i.e.,} the number of base-two
logarithms needed to bring $x$ below two.  For constant privacy and
utility parameters, the best known general lower bound is $\Om{v +
\logstar D}$ \citep{AlonEtAl2022,BlumerEtAl1989}. A recent work by
\citet{moran2026} gives the best known general upper bound, $\bigO{D\log
D}$. Finding a combinatorial characterization that closes this gap is
a central open problem, recently restated by
\citet{NissimStemmerTsfadia2026}.

Lower bounds for private learning are typically proved in one of two
ways. The first finds a simple problem hidden inside the larger class
such that we know a lower bound on the sample complexity of the simple
problem, and argues that every learner for the larger class must
require it. For example, \citet{AlonEtAl2022} use a threshold class.
By a quantitative form of Shelah's theorem, every class of Littlestone
dimension $D$ contains a threshold class of length about $\log D$, and
privately learning thresholds on $N$ points requires $\Om{\logstar
N}$ examples, which gives their bound $\Om{\logstar D}$.  The second
measures how many hypotheses are needed to~\emph{represent} the class
to realise a packing style bound. \citet{BeimelNS19} introduced
probabilistic representations and the associated probabilistic
representation dimension $\prdim$. Roughly this measures how compactly
a class can be represented by random sets of hypotheses and they
showed that every purely private learner needs
$\Om{\prdim(\cH)/(\alpha\epsilon)}$
examples.~\citet{FeldmanXiao2014} related $\prdim$ to randomized
one-way communication complexity. They showed that $\prdim$ is at
least of the order of the Littlestone dimension but can be much
larger, e.g. the class of lines over $\bZ_p^2$ has Littlestone
dimension $2$ and representation dimension $\Theta(\log p)$. They also
gave an approximate-private learner for this class whose sample
complexity, for fixed privacy and utility parameters, does not depend
on $p$. Thus, approximate privacy can be much cheaper than pure
privacy. Nevertheless, \(\prdim\) still gives a lower bound under
approximate privacy but only up to $\log(1/\delta)$: the journal
version of \citet{BeimelNS19} gives $\Om{\min\bc{\prdim(\cH),
\log(\epsilon/\delta)}/(\alpha\epsilon)}$. At constant
accuracy,~\citet{GhaziKM21} prove a bound of the same form in the more
general setting of user-level privacy. Additionally, the non-private
VC lower bound also applies to private learners. Combined with the
threshold lower bound, it gives the general lower bound $\Om{v +
\logstar D}$.

Many natural classes contain several hard learning problems on
disjoint parts of their domain. These problems are independent when
every combination of their target concepts can be realized by a single
hypothesis of the larger class. The bounds above need not capture
their combined private learning cost. This is the main observation our
lower bound leverages. We express this bound through the
\emph{Independent Class Packing parameter} $\icp(\cH, \cG)$. For a
collection \(\cG\) of classes, \(\icp(\cH,\cG)\) is the largest sum of
the sample complexities, at constant accuracy, of classes from \(\cG\)
that can be packed independently in \(\cH\). Our proof reduces the
private learning of \(\cH\) to learning the packed classes and
establishes an additive lower bound on the required sample size. The
hard problems used in earlier lower bounds provide the packed classes.
Packing thresholds of length $L$ adds the cost $\Om{\logstar L}$ for
each packed threshold class, and for small $\delta$, packing classes
with a large representation dimension adds the bound of
\citet{BeimelNS19} across these classes.

For the parameters that characterize other learning models, additivity
over independent parts is often direct. For independently packed
classes \(\cG_1,\ldots,\cG_k\) in \(\cH\), the VC dimension satisfies
\(\VC(\cH)\geq\sum_i\VC(\cG_i)\), so the corresponding non-private
lower bounds add. Similarly, \(\ldim(\cH)\geq\sum_i\ldim(\cG_i)\),
giving additive lower bounds for online learning. For private
learning, however, the known Littlestone-based lower bound depends on
\(\logstar\ldim(\cH)\). Assume \(\ldim(\cG_i)=D\) for all \(i\).
Using only this inequality in the known private lower bound gives
\(\Om{\logstar(kD)}\), rather than the sum \(\Om{k\logstar D}\). Our framework
establishes an additive lower bound directly in terms of the packed
classes’ private learning costs.

\begin{table}[t]
\centering
\small
\begin{tabular*}{\linewidth}{@{\extracolsep{\fill}}lccc@{}}
\toprule
Class
  & Previous lower bound & Our lower bound & Upper bound \\
\midrule
{\footnotesize Axis-aligned boxes}~$\cB_{N,d}$
  & \bcite{$\Om{d+\logstar N}$}
          {\citep{AlonEtAl2022,BlumerEtAl1989}}
  & \bcite{$\Om{d\logstar N}$}
          {\Cref{cor:four_classes}}
  & \bcite{$\softO{d\logstar N}$}
          {\citep{CohenEtAl2023,SadigurschiStemmer2021}} \\
\addlinespace[6pt]
{\footnotesize Unions of intervals}~$\cI_{N,k}$
  & \bcite{$\Om{k+\logstar N}$}
          {\citep{AlonEtAl2022,BlumerEtAl1989}}
  & \bcite{$\Om{k\logstar\frac Nk}$}
          {\Cref{cor:four_classes}}
  & \bcite{$\softO{k\logstar N}$}
          {\Cref{thm:interval-upper}} \\
\addlinespace[6pt]
{\footnotesize Affine halfspaces}~$\cH_{N,d}$
  & \bcite{$\Om{d+\logstar N}$}
          {\citep{AlonEtAl2022,BlumerEtAl1989}}
  & \bcite{$\Om{d\logstar N}$}
          {\Cref{cor:four_classes}}
  & \bcite{$\softO{d^{2.5}\logstar N}$}
          {\citep{karni2026private}} \\
\addlinespace[6pt]

{\footnotesize Convex $k$-gons}~$\cC_{N,k}$
  & \bcite{$\Om{k+\logstar N}$}
          {\citep{AlonEtAl2022,BlumerEtAl1989}}
  & \bcite{$\Om{k\logstar\frac{\sqrt N}{k}}$}
          {\Cref{cor:four_classes}}
  & \bcite{$\softO{k\log N}$}
          {\citep{KaplanMMS22}} \\
\addlinespace[6pt]
{\footnotesize Linear subspaces}~$\cS_{d,D}$
  & \bcite{$\Om{d}$}
          {\citep{BlumerEtAl1989}}
  & \bcite{$\Om{d\log(1/\delta)}$}
          {\Cref{lem:subspace-packing}}
  & \bcite{$\bigO{d\log(1/\delta)}$}
          {\citep{moran2026}} \\
\addlinespace[6pt]
{\footnotesize Unions of tree paths}~$\cT_{d,k}$
  & \bcite{$\Om{d}$}
          {\citep{BlumerEtAl1989}}
  & \bcite{$\Om{d\log(1/\delta)}$}
          {\Cref{cor:tree_path_packing}}
  & \bcite{$\bigO{dk}$}
          {} \\
\bottomrule
\end{tabular*}
\normalsize
\caption[Private PAC sample bounds]{\footnotesize Private PAC sample
bounds for axis-aligned boxes $\cB_{N,d}$ and affine halfspaces
$\cH_{N,d}$ on $\bs N^d$, unions of at most $k$ intervals $\cI_{N,k}$
on $\bs N$,  convex $k$-gons $\cC_{N,k}$ on $\bs N^2$, linear
subspaces $\cS_{d,D}$ of dimension at most $d$ of $\bF_p^D$, and
unions $\cT_{d,k}$ of at most $d$ root-to-leaf paths in the complete
binary tree of depth $2k$. Privacy and utility parameters and additive
constants in $\logstar$ are suppressed, except $\delta$ in the last
two rows. These two rows are shown for $\log(1/\delta) \leq \min\bc{d,
\log p}$ and $\log(1/\delta) \leq \min\bc{d, k}$, respectively. The
parameter ranges are stated in
\Cref{cor:four_classes,cor:tree_path_packing} and
\Cref{lem:subspace-packing}.  The upper bound for $\cT_{d,k}$ follows
from the exponential mechanism.}
\label{tab:natural-classes}
\end{table}

\subsection{Our contributions.}
We first show that the VC and Littlestone dimensions do not determine
the private sample complexity of a class as two classes with the same
VC and Littlestone dimensions can have different private sample
complexities. The underlying reason is that the cost of private
learning adds up over independent hard problems inside a class in a
way that existing bounds do not capture. The \OurDimAcronym{}
parameter captures this.  It gives a lower bound for every class, and
this lower bound is tight up to logarithmic factors for several
natural classes. All our lower bounds hold for improper learners.

\medskip
\noindent\textbf{A separation at equal \(\VC\) and \(\ldim\)} For
every \(1\leq v\leq D\), we construct classes \(\ProdClass_{v,D}\) and
\(\UnionClass_{v,D}\), both with dimensions \((v,D)\), such that for
the same privacy and utility parameters \[
  \begin{aligned}
    \mprivshort{\ProdClass_{v,D}}
    &=\widetilde\Theta\br{v\logstar\br{\max\bc{2,D/v}}},\\
    \mprivshort{\UnionClass_{v,D}}
    &=\widetilde\Theta\br{v+\logstar D}.
  \end{aligned}
\] Here \(\widetilde{\Theta}\) hides factors polylogarithmic in the
rates~(see~\Cref{thm:two-families} for full details).  Taking $v =
\logstar D$, the sample complexities differ by $\softOm{\logstar D}$,
so no function of $(v, D)$ can determine the private sample complexity
up to polylogarithmic factors. However, this does not rule out an
upper bound \(\poly{v,\logstar D}\) that holds for every class.

\medskip
\noindent\textbf{A stronger lower bound.} We propose the \OurDimName\
(\OurDimAcronym{}) parameter in~\Cref{def:icp}. Our main result,
\Cref{thm:lower_bound_icp}, shows that all $(\epsilon, \delta)$-DP,
$(\alpha, \beta)$-PAC learner of $\cH$ requires
\[
  n=\Om{
    \frac{\icp_{0.1,\delta/\epsilon}(\cH,\cG)}{\epsilon\alpha}
  }~\text{samples}.
\]

\noindent If $\cH\in\cG$, the one-component packing of $\cH$ into
itself only compares its sample complexity at different privacy and
accuracy parameters. The interesting
regime is when $\cG$ is restricted to simple classes with known sample
complexity, such as thresholds, so that packing several of them into
$\cH$ gives explicit lower bounds that add over the packed classes.
The strength of these bounds therefore depends on what is known about
the sample complexity of the packed classes, and any improved lower
bound for a simple class immediately gives improved lower bounds for
every class into which several copies of it can be packed.

This lower bound matches the upper bounds for \(\ProdClass_{v,D}\) and
\(\UnionClass_{v,D}\), up to logarithmic factors, via packing
thresholds.  For \(\ProdClass_{v,D}\), the packing contains all
threshold components, so their sample complexities add in
\OurDimAcronym{}.  For \(\UnionClass_{v,D}\), the Boolean and the
threshold components cannot be packed independently together. The
separate VC and threshold lower bound already give \(v+\logstar D\)
which matches the upper bound~\Cref{sec:classes}.  More generally, let
$\cG_{\leq 1}$ be the collection of all $\VC = 1$  classes (for example thresholds). Then we have that
for every class $\cH$ with VC dimension $v$ and Littlestone dimension
$D$,
\begin{equation}
  \Om{v + \logstar D}
  \lesssim \icp_{0.1,\delta/\epsilon}(\cH,\cG_{\leq 1})
  \lesssim \softO{v\logstar\br{D/v}}.\qquad(\Cref{lem:icp_vc_ldim})
\end{equation}

\medskip
\noindent\textbf{Applications to natural classes.} The packing idea
applies well beyond \(\ProdClass_{v,D}\) and \(\UnionClass_{v,D}\).  A
simple packing also gives multiplicative lower bounds for
\begin{enumerate*}[label=(\roman*), itemjoin={,~ }, itemjoin*={, and~}]
\item axis-aligned boxes
\item affine halfspaces on finite grids
\item unions of intervals
\item convex polygons in the plane \end{enumerate*}.  For boxes on
$[N]^d$ we obtain $\Om{d\logstar N}$, and for unions of $k$ intervals
on $[N]$ we obtain $\Om{k\logstar(N/k)}$. These improve the earlier
bounds \(\Om{d+\logstar N}\) and \(\Om{k+\logstar N}\), respectively.
They match the known upper bound for boxes and our upper bound for
unions of intervals, up to logarithmic factors. Resolving the gap for
halfspaces and polygons remains open.

For small $\delta$, we also add the representation-dimension lower
bound of \citet{BeimelNS19,GhaziKM21} across packed classes. At fixed
$\epsilon,\alpha,\beta$, this bound contributes at most
$\bigO{\log(1/\delta)}$ when applied to the whole class, whereas each
packed class can contribute this amount. Packing lines into linear
subspaces and paths into unions of tree paths gives
$\Om{d\log(1/\delta)}$, improving the $\Om d$ bounds from VC. The subspace bound is tight. For unions of
paths, it is tight when $\log(1/\delta)=\Theta(k)$. In this regime,
packing paths gives $\Om{dk}$, whereas every threshold packing yields
a lower bound of at most $\bigO{d\log(k+1)}$ through $\icp$
(\Cref{lem:tree_threshold_packing}). Thus the improvement over
threshold packings grows without bound with
$k$.~\Cref{tab:natural-classes} summarizes the bounds for these six
classes.

\medskip
\noindent\textbf{Upper bounds in terms of \OurDimAcronym{}.} It is
natural to ask whether \OurDimAcronym{} also controls private sample
complexity from above. We show that it does when the independently
packed classes \(\cG=\bc{\cG_i}\) also cover $\cH$~(\Cref{def:cover}).
Here covering means that every hypothesis of $\cH$ can be represented
on the whole domain by a fixed Boolean combination of hypotheses from
\(\bc{\cG_i}\). For such coverings, \Cref{thm:boolean_aggr}, using the closure algorithm of~\citet{AlonEtAl2020}, gives a
learner using
\[\softO{\VC(\cH)^3\icp_{0.1,\delta/\epsilon}(\cH,\cG)/\alpha\epsilon}
\quad\text{samples,}\] which is within a factor $\VC(\cH)^3$ of our
lower bound. For example, the class of boxes anchored at the origin
packs one threshold class per coordinate and is covered by the
conjunction of these coordinate thresholds. We remove this extra
$\VC(\cH)^3$ factor when the target is a disjunction of components on
disjoint parts of the domain, as in product classes like
\(\ProdClass_{v,D}\). Each component can then be learned directly from
the labels on its own part thereby matching the lower bound up to
logarithmic factors.

\medskip
\noindent\textbf{A threshold lower bound for larger \(\delta\).}
For improper learning of thresholds on $N$ points,
\citet{AlonEtAl2022} prove that $m=\Om{\logstar N}$ examples are
necessary when $\delta=\bigO{1/(m^2\log m)}$, at fixed constant
privacy parameter $\epsilon$, accuracy and confidence. We prove the
same lower bound under the weaker restriction $\delta\leq c/m$, where
$c>0$ is universal (\Cref{thm:threshold-m-rows} and \Cref{cor:threshold_pac}). In
particular, this includes every choice $\delta(m)=o(1/m)$ for
sufficiently large $m$.

Together, these results suggest that a characterization of private
sample complexity should account for the combined cost of
independently packed classes, as $\icp$ does, rather than depend only
on the VC and Littlestone dimensions of individual classes. Whether
$\icp$ with a fixed collection of simple classes gives such a
characterization, which requires matching upper bounds for every
class, is the main question that we leave open.

\medskip\noindent\textbf{Paper structure} \Cref{sec:classes} gives the
separating classes.  \Cref{sec:icp} proves the general lower bound.
Then in \Cref{sec:packing_thresholds}, we first prove the improved
threshold lower bound in \Cref{sec:threshold-lower-bound} and then in
\Cref{sec:applications} apply it to boxes, unions of intervals,
halfspaces on a finite grid and convex polygons in a plane, and in
\Cref{sec:small_delta} extend these bounds to small $\delta$.
\Cref{sec:beyond_thresholds} gives packings of classes other than
thresholds, and \Cref{sec:upper_bounds} gives upper bounds in terms of
\icp{} using coverings.  Detailed proofs are deferred to
appendix.

\subsection{Technical overview}

We first define independent packings and the \OurDimAcronym{}
parameter, then describe the main challenges behind our results.
An independent packing places classes on pairwise disjoint parts
of the domain $\cX$, so that every choice of one hypothesis from
each class can be realised by a single hypothesis in $\cH$.\looseness-1

\begin{definition}
    [Packable classes]
    \label{def:packing_classes}
    Let $\cH \subseteq \bc{0,1}^\cX,~\cG_i \subseteq \bc{0,1}^{\cY_i}$
    for $i \in \bs{k}$ be nonempty classes. We say $\br{\cG_1, \cdots,
    \cG_k}$ is independently packable in $\cH$ if there exist $k$
    injective maps $\Phi_i: \cY_i \rightarrow \cX$ with pairwise
    disjoint images such that for every $ \br{g_1, \cdots, g_k} \in
    \prod_{i=1}^k \cG_i$,  some $h \in \cH$ satisfies
    \[
    h\br{\Phi_i(y)} = g_i(y) \; \text{for all} \; i \in [k], y \in \cY_i.
    \]

\end{definition}

\noindent  We denote such a packable tuple of classes by $\br{\cG_1,
\cdots, \cG_k} \sqsubseteq \cH$. We now define our main combinatorial
parameter which characterises our lower bound.

\begin{definition}
    [\OurDimName{} (\OurDimAcronym{})]
    \label{def:icp}
    Fix privacy parameters $\epsilon > 0$ and $0 \leq \delta < 1$, a
    constant accuracy parameter $\alphazero \leq 1/4$ and a
    collection $\cG$ of nonempty binary classes, and for each $\cG_i
    \in \cG$ let $\mpriv{\cG_i}{\epsilon}{\delta}{\alphazero}{1/4}$ be
    as defined in \Cref{sec:prelims}.
    For any hypothesis class $\cH$ on  $\cX$, define
    \[
    \icp_{\epsilon,\delta}(\cH, \cG) := \sup_{\substack{(\cG_1, \cdots, \cG_k) \sqsubseteq \cH \\ \cG_i\in\cG}} \sum_{i = 1}^k \mpriv{\cG_i}{\epsilon}{\delta}{\alphazero}{1/4}.
    \]
\end{definition}
\noindent\textbf{Lower bounds via adding independent costs}
The main idea behind \Cref{thm:lower_bound_icp} is to hide one hard
database for each packed class inside a single database for $\cH$. The
packing condition allows the target concepts on the disjoint parts of
the domain to be chosen independently. Our aim is to force a constant
fraction of mistakes in expectation on each part that has too few
examples, and then add these mistakes.
We first work with empirical error on fixed databases. By resampling,
a private PAC learner for $\cH$ gives a private algorithm with small
empirical error on every realizable database (\Cref{lem:resample}).
For each packed class \(\cG_i\), we find a distribution over
realizable databases on which every private algorithm has a constant
expected empirical error when the database size is below the class’s
private sample complexity (\Cref{lem:hard_dist}). As the same
distribution works for every algorithm, we can draw the hard database
for each $\cG_i$ independently.

Fix all hard databases but the $i$-th. With the other databases
fixed, running the learner and restricting its predictions to the
$i$-th part gives an algorithm for $\cG_i$. This algorithm is
private because changing one example in its input changes only
one example in the combined database.
Since its input is drawn from the hard distribution for $\cG_i$, it
mislabels more than an an $\bar{\alpha}$ fraction of its examples
(\Cref{lem:hard_dist}). Averaging over the other databases and adding
these errors over all parts gives more than $\bar{\alpha}Q$ mistakes
in expectation, where $Q$ is the total number of examples in the hard
databases (\Cref{lem:direct_sum}).

Suppose that the hard databases form the whole database that the
learner receives, which has $Q$ examples. Its empirical accuracy
guarantee bounds the expected number of mistakes by $(\alpha+\beta)Q$:
it makes at most $\alpha Q$ mistakes when it succeeds, but may
mislabel all $Q$ examples when it fails. To contradict the lower bound
$\bar{\alpha}Q$, we therefore need the failure probability to be small
as well. We first amplify the learner's confidence to reduce its
failure probability below $\bar{\alpha}/2$ at the cost of constant
factors in error (\Cref{lem:resample}). For sufficiently small
$\alpha$, its expected number of mistakes is then below
$\bar{\alpha}Q$, giving the contradiction. We can fill the input with
such databases whenever its size is smaller than the sum of the
component sample complexities. This gives an additive lower bound for
each packing and taking the supremum over packings gives the $\icp$
lower bound.

\medskip\noindent\textbf{The factors $1/\alpha$ and $1/\epsilon$:} To obtain the
factor $1/\alpha$, we place the hard databases inside a larger
database whose remaining examples are copies of a fixed padding point,
keeping the larger database realisable (\Cref{lem:padding}). The \(Q\)
hard examples still force $\Om{Q}$ expected mistakes on the hard
databases. Combining this with the accuracy guarantee on the full
database of size \(m\) gives
\(m=\Om{Q/\alpha}\).
The factor $1/\epsilon$ comes from a standard reduction of repeating
each example \(\Theta(1/\epsilon)\) times. By group privacy, this
converts a learner at privacy $(\epsilon,\delta)$ into one at privacy
$(0.1,\delta/\epsilon)$ using fewer examples (\Cref{lem:duplicate}).
This gives the following theorem.

\begin{restatable}[Lower bound via~\OurDimAcronym]{thm}{thmLowerBoundICP}
    \label{thm:lower_bound_icp}
    Fix $0 < \alphazero \leq 1/4$, let $\cH$ be a non-trivial
    (\Cref{def:non_trivial}) binary class, $\cG$ be a collection of
    nonempty binary classes over finite domains. Let $0 < \epsilon \leq 0.1$, $0 \leq
    \delta \leq \alphazero\epsilon/64$, $0 < \alpha \leq
    \alphazero/(1536e)$ and $0 < \beta \leq 1/3$. Then every $(\epsilon,
    \delta)$-DP, $(\alpha, \beta)$-PAC learner for $\cH$ requires
    \[
    n = \Om{\frac{\icp_{0.1,\delta/\epsilon}(\cH, \cG)}{\alpha\epsilon}}~\enskip\text{samples}.
    \]
\end{restatable}
\noindent\textbf{A threshold lower bound for larger $\delta$:}
For the threshold packings in our applications, we extend the
$\Om{\logstar N}$ lower bound of \citet{AlonEtAl2022} to the larger
range $\delta \leq c/m$ (\Cref{thm:threshold-m-rows}). After applying
Ramsey's theorem, their proof constructs databases $S_x$ that differ
only in one point $x$ and obtains this fixed level $r$, such that the
probability that $A(S_x)$ labels a point $y$ by $1$ lies below
$r-\Om{1/m}$ when $y<x$ and above $r+\Om{1/m}$ when $y>x$. Then, they
conduct a binary search on \(y\) to find \(x\) by estimating this
probability and comparing it with \(r\). However, to make the
comparisons accurate, they need to average $\Theta(m^2\log m)$ outputs
on each database, which then due to composition requires $\delta =
\bigO{1/(m^2\log m)}$. We avoid this by comparing the prediction
probabilities directly  without averaging outputs. As in their proof,
for an ordered set $U=\bc{x_1<\cdots<x_{m+1}}$, we form a database by
omitting $x_i$ and assigning nearly equal numbers of zero and one
labels. Let $a_i(U)$ be the probability that the learner's output
labels the omitted point by $1$. Then, we show that accuracy and
privacy force a later value of $a_i(U)$ to exceed an earlier one by a
constant. Hence some adjacent increase $a_{\ell+1}(U)-a_\ell(U)$ is at
least $\kappa/m$, for a universal constant $\kappa>0$. Ramsey's
theorem then gives a large set on which the first such index is the
same for every $(m+1)$-point subset $U$.

We then fix $m-1$ points and vary the one at this index. For two
choices $x<y$, the neighboring databases $S_x,S_y$ satisfy \(
\Pr\bs{A(S_x)(y)=1}-\Pr\bs{A(S_y)(x)=1}\geq\kappa/m.\) Using this and
the DP inequality, the discrete Hilbert inequality bounds the number
of possible choices by $\exp(\bigO{m})$ when $\delta\leq c/m$
(\Cref{lem:skew-matrix}). Combining this with the quantitative Ramsey
bound as done by~\citet{AlonEtAl2022} gives $m=\Om{\logstar N}$.

\medskip\noindent\textbf{Applications to Geometric Classes:} We consider the following classes.
\begin{enumerate}[itemsep=0.05em, leftmargin=2.5em, label=(\roman*)]
    \item \emph{Axis-aligned Boxes:} For $N \geq 3$ and $d \geq 1$, let $\cB_{N,d}$ be the class of axis-aligned boxes on $[N]^d$, that is, the functions $x \mapsto \one\bc{x \in \prod_{j=1}^d [a_j, b_j]}$ with integers $1 \leq a_j \leq b_j \leq N$.
    \item \emph{Halfspace:} Let $\cH_{N,d}$ be the class of affine halfspaces on $[N]^d$, i.e., the functions $x \mapsto \one\bc{\ip{w}{x} \leq \theta}$ with $w \in \bR^d$ and $\theta \in \bR$.
    \item \emph{Union of Intervals:} For $N \geq 2k$, let $\cI_{N,k}$ be the class of unions of at most $k$ intervals on $[N]$.
    \item \emph{Convex Polygon:} For integers $N, k \geq 1$, let $\cC_{N,k}$ be the class of convex $k$-gons on $[N]^2$ studied in \cite{KaplanMMS22}, that is, the indicators, restricted to $[N]^2$, of intersections of at most $k$ halfplanes $\bc{z \in \bR^2 : \ip{w}{z} \leq \theta}$.
\end{enumerate}

\begin{figure}[t]
\centering
\begingroup
\definecolor{packingblue}{RGB}{32,91,133}
\definecolor{packinggray}{RGB}{116,121,126}
\scalebox{0.85}{%
\begin{tikzpicture}[
  x=1cm,y=1cm,
  font=\footnotesize,
  packing axis/.style={draw=black!48,line width=.45pt,-{Stealth[length=3.5pt]}},
  packing boundary/.style={draw=packingblue,line width=.8pt},
  packing one/.style={circle,draw=packingblue,fill=packingblue,
    inner sep=0pt,minimum size=4.6pt,line width=.55pt},
  packing zero/.style={circle,draw=black!68,fill=white,
    inner sep=0pt,minimum size=4.6pt,line width=.55pt},
  packing label/.style={text=packingblue,fill=white,inner sep=1.4pt},
  packing title/.style={anchor=west,font=\small\bfseries},
  packing brace/.style={draw=black!45,line width=.4pt,
    decorate,decoration={brace,mirror,amplitude=3pt}}
]
\path[use as bounding box] (-.05,-.2) rectangle (15.65,11.75);

\begin{scope}[shift={(0,6.2)}]
  \node[packing title] at (.12,5.18) {(a) Boxes};
  \begin{scope}[shift={(.83,.87)},x=.84cm,y=.61cm]
    \fill[packingblue!9] (0,0) rectangle (3,4);
    \draw[packing boundary] (0,0) rectangle (3,4);
    \draw[packing axis] (-.18,0)--(6.55,0)
      node[right,text=black] {$\Phi_1$};
    \draw[packing axis] (0,-.18)--(0,6.55);
    \node[anchor=west] at (.20,6.20) {$\Phi_2$};
    \foreach \r in {1,...,6} {
      \ifnum\r<4 \node[packing one] at (\r,0) {};
      \else \node[packing zero] at (\r,0) {};\fi
      \ifnum\r<5 \node[packing one] at (0,\r) {};
      \else \node[packing zero] at (0,\r) {};\fi
    }
    \draw[black!58,line width=.6pt] (-.07,-.1)--(.07,.1)
      (-.07,.1)--(.07,-.1);
    \node[below left,inner sep=2pt] at (0,0) {$u$};
    \node[packing label,below] at (3,-.15) {$t_1=3$};
    \node[packing label,above right] at (.20,4) {$t_2=4$};
    \draw[black!28,densely dashed,line width=.4pt] (3.5,.3)--(3.5,-.3);
    \draw[black!28,densely dashed,line width=.4pt] (-.23,4.5)--(.23,4.5);
  \end{scope}
\end{scope}

\begin{scope}[shift={(8.1,6.2)}]
  \node[packing title] at (.12,5.18) {(b) Halfspaces};
  \begin{scope}[shift={(.83,.87)},x=.84cm,y=.61cm]
    \fill[packingblue!9] (0,0)--(3.5,0)--(0,4.5)--cycle;
    \draw[packing boundary] (-.26,4.8343)--(3.82,-.4114);
    \draw[packing axis] (-.18,0)--(6.55,0)
      node[right,text=black] {$\Phi_1$};
    \draw[packing axis] (0,-.18)--(0,6.55);
    \node[anchor=west] at (.20,6.20) {$\Phi_2$};
    \foreach \r in {1,...,6} {
      \ifnum\r<4 \node[packing one] at (\r,0) {};
      \else \node[packing zero] at (\r,0) {};\fi
      \ifnum\r<5 \node[packing one] at (0,\r) {};
      \else \node[packing zero] at (0,\r) {};\fi
    }
    \draw[black!58,line width=.6pt] (-.07,-.1)--(.07,.1)
      (-.07,.1)--(.07,-.1);
    \node[below left,inner sep=2pt] at (0,0) {$u$};
    \node[packing label,below] at (3.5,-.35) {$t_1=3$};
    \node[packing label,above right] at (.18,4.65) {$t_2=4$};
  \end{scope}
\end{scope}

\begin{scope}[shift={(0,.65)}]
  \node[packing title] at (.12,5.18) {(c) Unions of intervals};
  \begin{scope}[shift={(.57,2.65)},x=.71cm,y=1cm]
    \draw[packing axis] (-.25,0)--(8.75,0) node[right,text=black] {$[N]$};
    \foreach \start/\t/\i in {0/1/1,3/2/2,6/1/3} {
      \pgfmathsetmacro{\last}{\start+\t-1}
      \fill[packingblue!13] (\start-.16,-.13) rectangle (\last+.16,.13);
      \draw[packing boundary,line width=1.3pt]
        (\start-.12,0)--(\last+.12,0);
      \draw[packing brace] (\start-.22,-.45)--(\start+2.22,-.45)
        node[midway,below=5pt,text=black] {$\Phi_{\i}$};
      \node[packing label] at (\start+1,.58) {$t_{\i}=\t$};
    }
    \foreach \j in {0,3,4,6} \node[packing one] at (\j,0) {};
    \foreach \j in {1,2,5,7,8} \node[packing zero] at (\j,0) {};
    \draw[black!22,densely dashed,line width=.4pt] (2.5,-.23)--(2.5,.24);
    \draw[black!22,densely dashed,line width=.4pt] (5.5,-.23)--(5.5,.24);
  \end{scope}
  \node[text=black!72,align=center] at (3.67,1.15)
    {One interval keeps the chosen prefix\\of each block.};
\end{scope}

\begin{scope}[shift={(8.1,.65)}]
  \node[packing title] at (.12,5.18) {(d) Convex polygons};
  \begin{scope}[shift={(3.85,.90)},x=.70cm,y=.185cm]
    \begin{scope}
      \clip (-4.5,-2.5) rectangle (4.55,21);
      \fill[packingblue!9]
        (-4.5,17.25)--(-.642857,-2.035714)--(2.1,3.45)
        --(4.55,20.6)--(4.55,21)--(-4.5,21)--cycle;
      \draw[packingblue!55,densely dashed,line width=.45pt]
        (-4.5,17.25)--(.4,-7.25);
      \draw[packingblue!55,densely dashed,line width=.45pt]
        (-4.5,-9.75)--(4.55,8.35);
      \draw[packingblue!55,densely dashed,line width=.45pt]
        (.7,-6.35)--(4.55,20.6);
      \draw[packing boundary]
        (-4.5,17.25)--(-.642857,-2.035714)--(2.1,3.45)--(4.55,20.6);
      \draw[black!62,line width=.6pt,smooth,domain=-4.4:4.4,samples=80]
        plot (\x,{\x*\x});
    \end{scope}
    \foreach \x in {-4,-1,0,2} \node[packing one] at (\x,{\x*\x}) {};
    \foreach \x in {-3,-2,1,3,4} \node[packing zero] at (\x,{\x*\x}) {};
    \draw[packing brace] (-4.23,-3.10)--(-1.77,-3.10)
      node[midway,below=5pt,text=black] {$\Phi_1$};
    \draw[packing brace] (-1.23,-3.10)--(1.23,-3.10)
      node[midway,below=5pt,text=black] {$\Phi_2$};
    \draw[packing brace] (1.77,-3.10)--(4.23,-3.10)
      node[midway,below=5pt,text=black] {$\Phi_3$};
    \node[text=black!68] at (-3.2,20.6) {Parabola};
    \node[text=packingblue,align=center] at (.25,15.5)
      {Intersection of\\the halfplanes};
  \end{scope}
\end{scope}

\node[packing one] at (5.65,.12) {};
\node[anchor=west] at (5.84,.12) {label $1$};
\node[packing zero] at (8.25,.12) {};
\node[anchor=west] at (8.44,.12) {label $0$};
\end{tikzpicture}
}
\endgroup
\caption{Threshold packings for the four geometric classes. Filled points form the chosen prefixes and open points form the remaining suffixes. The polygon panel is shown after change of coordinates.}
\label{fig:geometric_packings}
\end{figure}

We obtain the lower bounds for the four classes in
\Cref{tab:natural-classes} by packing thresholds. A packing of $k$
copies of a threshold class into $\cH$ consists of $k$ disjoint
sequences of $L$ points each, such that for every choice of cutoffs
$t_1, \cdots, t_k \in \bc{0, \cdots, L}$, a single hypothesis of $\cH$
labels exactly the first $t_i$ points of the $i$-th sequence by $1$.
For boxes and halfspaces on $[N]^d$, take one sequence of length $N-1$
along each coordinate direction from the corner $(1,\cdots,1)$,
leaving out the corner so that the $d$ sequences are disjoint. For
boxes, fix the lower endpoint in every coordinate at $1$ and choose
the cutoff independently. Changing the cutoff in coordinate $i$
changes the prefix kept on the $i$-th sequence without
affecting the others. For halfspaces, translate the corner to the
origin and consider $\sum_{j=1}^d a_j x_j\leq1$. On the $i$-th
sequence, only $x_i$ is nonzero, so this reduces to $a_i x_i\leq1$.
Choosing $a_i$ sets the cutoff on this sequence
without affecting the others.

For unions of at most $k$ intervals on $[N]$, take $k$ disjoint
consecutive blocks, each of length $\floor{N/k}$.  Any choice of
prefixes is realized by taking their union, which uses at most one
interval per block. For convex $k$-gons on $[N]^2$, we use the integer
points on the parabola $z_2=z_1^2$ and divide them into $k$
consecutive blocks of length about $\sqrt N/k$.  The useful property
of the parabola is that a halfplane above a line through two of its
points excludes the points between them and contains those outside. By
placing the two intersections just before and just after the unwanted
suffix of a block, we obtain a halfplane that keeps its chosen prefix
and every point of the other blocks. Intersecting these halfplanes
realizes all the chosen prefixes simultaneously. Thus the four
constructions pack $d$ or $k$ threshold classes whose costs add by
\Cref{thm:lower_bound_icp}, giving the multiplicative lower bounds of
\Cref{cor:four_classes}. See \Cref{fig:geometric_packings} for an
illustration of the packings.

\medskip\noindent\textbf{Packing beyond thresholds:} Threshold packings can give weak lower bounds even when the class has large representation dimension. For example, the class of lines over $\bZ_p^2$, studied by \citet{FeldmanXiao2014}, has VC and Littlestone dimension $2$ and representation dimension $\Theta(\log p)$. Every threshold packing of this class has at most two components, each of length at most $2$, so it can yield a lower bound of at most $\bigO{1/(\alpha\epsilon)}$ through $\icp$ (\Cref{thm:lower_bound_icp}). On the other hand, \Cref{lem:prdim_lower}, together with the exponential mechanism and the learner of \citet{FeldmanXiao2014}, shows that the sample complexity of lines is $\Theta(\min\bc{\log p, \log(1/\delta)}/(\alpha\epsilon))$ for constant $\beta$. We then pack several copies of lines into the class $\cS_{d,D}$ of linear subspaces of dimension at most $d$ of $\bF_p^D$. We split the first $3\floor{d/3}$ coordinates into blocks of size three, and map a point $(x, y)$ of the $i$-th copy to the vector that equals $(x, y, 1)$ on the $i$-th block and zero elsewhere. A line $y = a_i x + b_i$ then corresponds to the two-dimensional subspace $\bc{(x, y, z) : y = a_i x + b_i z}$ of the $i$-th block, and the direct sum of these subspaces has dimension at most $d$ and contains the image of $(x, y)$ exactly when $(x, y)$ lies on the $i$-th line. This gives $\Om{d\min\bc{\log p, \log(\epsilon/\delta)}/(\alpha\epsilon)}$ (\Cref{lem:subspace-packing}). For fixed $\epsilon,\alpha,\beta$, this bound is tight up to constants when $p\geq\epsilon/\delta$, by the learner of \citet{moran2026}.

Our second example is the class $\cT_{d,k}$ of unions of at most $d$ root-to-leaf paths in the complete binary tree of depth $2k$. We choose $d$ distinct nodes at depth $k$ and map a tree of depth $k$ onto the subtree below each of them, so that any choice of one path in each subtree is realized by extending these paths to the root and taking their union. See \Cref{fig:path_packing} for an illustration with $d=k=2$. For this class, every threshold packing gives at most $\bigO{d\log(k+1)/(\alpha\epsilon)}$,  whatever lower bound is used for a single threshold class (\Cref{lem:tree_threshold_packing}). Indeed, such a packing has at most $\VC(\cT_{d,k}) = d$ components, and their total length is at most $d(2k+1)$, as a union of $d$ paths labels at most this many nodes by $1$. Instead, packing $d$ copies of the class of paths, which, like a threshold class, has VC dimension $1$ but has representation dimension $\Theta(k)$, gives $\Om{d\min\bc{k, \log(\epsilon/\delta)}/(\alpha\epsilon)}$ (\Cref{cor:tree_path_packing}). The representation dimension bounds from two packings agree up to constants for $\delta/\epsilon \geq 1/(k+1)$. The path packing is stronger by an unbounded factor once $\delta/\epsilon \leq (k+1)^{-\omega(1)}$, and for $\delta \leq \epsilon e^{-k}$ it gives $\Om{dk/(\alpha\epsilon)}$, which matches the exponential mechanism for constant $\beta$. To see the gap more concretely, consider constant $\epsilon$, $\alpha$ and $\beta$, with $k = \ceil{\log_2 d}$ and $\delta = d^{-2}$. The path packing shows that the sample complexity is $\Theta(d\log d)$, every threshold packing gives at most $\bigO{d\log\log d}$, and the previously known bounds give only $\Om{d}$.

\begin{figure}[!b]
\centering
\begingroup
\definecolor{packingblue}{RGB}{32,91,133}
\begin{tikzpicture}[
 x=0.8cm,y=0.8cm,font=\footnotesize,
 one/.style={circle,draw=packingblue,fill=packingblue,
   inner sep=0pt,minimum size=4.6pt,line width=.55pt},
 zero/.style={circle,draw=black!68,fill=white,
   inner sep=0pt,minimum size=4.6pt,line width=.55pt},
 quiet/.style={text=black!65},
 embedded domain/.style={draw=black!38,densely dashed,line width=.45pt},
 edge/.style={draw=black!32,line width=.5pt},
 chosen/.style={draw=packingblue,line width=1pt},
 brace/.style={draw=black!45,line width=.4pt,
   decorate,decoration={brace,mirror,amplitude=3pt}}
]
\path[use as bounding box] (-.35,-.30) rectangle (7.75,4.75);
\foreach \level in {0,...,4} {
 \pgfmathtruncatemacro{\last}{2^\level-1}
 \foreach \j in {0,...,\last} {
  \coordinate (n-\level-\j)
    at ({(\j+.5)*7.04/(2^\level)},{4.50-.78*\level});
 }
}
\foreach \center in {.88,4.4} {
 \path[fill=black!2] (\center,3.35)
   --(\center-1.05,1.20)--(\center+1.05,1.20)--cycle;
 \draw[embedded domain] (\center,3.35)
   --(\center-1.05,1.20)--(\center+1.05,1.20)--cycle;
}
\foreach \level in {1,...,4} {
 \pgfmathtruncatemacro{\previous}{\level-1}
 \pgfmathtruncatemacro{\last}{2^\level-1}
 \foreach \j in {0,...,\last} {
  \pgfmathtruncatemacro{\parent}{floor(\j/2)}
  \draw[edge] (n-\previous-\parent)--(n-\level-\j);
 }
}
\draw[chosen] (n-0-0)--(n-1-0)--(n-2-0)--(n-3-0)--(n-4-1);
\draw[chosen] (n-0-0)--(n-1-1)--(n-2-2)--(n-3-5)--(n-4-10);
\foreach \level in {0,...,4} {
 \pgfmathtruncatemacro{\last}{2^\level-1}
 \foreach \j in {0,...,\last} \node[zero] at (n-\level-\j) {};
}
\foreach \level/\j in {0/0,1/0,1/1,2/0,2/2,3/0,3/5,4/1,4/10}
  \node[one] at (n-\level-\j) {};
\node[anchor=east] at ($(n-2-0)+(-.20,.25)$) {$c_1$};
\node[anchor=east] at ($(n-2-2)+(-.20,.25)$) {$c_2$};
\foreach \y/\lab in {4.5/0,2.94/k,1.38/2k} {
 \draw[black!35,line width=.4pt] (7.13,\y)--(7.25,\y);
 \node[quiet,anchor=west] at (7.30,\y) {$\lab$};
}
\draw[brace] (-.17,1.00)--(1.93,1.00)
  node[midway,below=5pt] {$\Phi_1(X_k)$};
\draw[brace] (3.35,1.00)--(5.45,1.00)
  node[midway,below=5pt] {$\Phi_2(X_k)$};
\node[one] at (2.30,-.05) {};
\node[anchor=west] at (2.48,-.05) {label $1$};
\node[zero] at (4.05,-.05) {};
\node[anchor=west] at (4.23,-.05) {label $0$};
\end{tikzpicture}
\endgroup
\caption{Packing $d=2$ copies of the class of root-to-leaf paths of a tree of depth $k=2$ into $\cT_{d,k}$. The nodes $c_1$ and $c_2$ at depth $k$ are fixed, and $\Phi_i$ maps the tree $X_k$ of depth $k$ onto the subtree below $c_i$ (dashed). Choosing one path in each subtree, the union of the two root-to-leaf paths that extend these choices labels exactly the filled nodes by one.}
\label{fig:path_packing}
\end{figure}

\paragraph{Upper bounds via covering:} In \Cref{sec:upper_bounds} we
complement \Cref{thm:lower_bound_icp} with upper bounds. For our upper
bounds, we assume that every hypothesis \(h\in\cH\) can be built
from one hypothesis from each of the packed classes. We express this
through coverings (\Cref{def:cover}). A covering fixes, for each $i$,
a part $\cX_i$ of the domain and a map $\Psi_i:\cX_i\to\cY_i$. Each
$g_i\in\cG_i$ then gives a hypothesis $\widetilde g_i$ on $\cX$, which
equals $g_i\circ\Psi_i$ on $\cX_i$ and a fixed function $o_i$ outside
$\cX_i$. The tuple $\br{\cG_1,\cdots,\cG_k}$ covers $\cH$ if, for a
fixed Boolean function $\cF$, every $h\in\cH$ equals
$\cF\br{\widetilde g_1,\cdots,\widetilde g_k}$ for some choice of
$g_1\in\cG_1,\cdots,g_k\in\cG_k$.
To prove the upper bound, we first show~(\Cref{lem:transfer}) that a
private learner for $\cG_i$ gives a private learner for
\(\widetilde{g_i}\) on \(\cX\) with only constant blowup in sample
complexity. Now, we discuss two cases. First suppose that the parts
$\cX_i$ cover $\cX$ and that every $h\in\cH$ agrees on each point in
$\cX_i$ with some $g_i\in\cG_i$.  This is satisfied by the product
classes we discussed above. Then each learner can be trained directly
on the examples in its own part, and all learners can share one
sample. Let $s_i$ be the cost of $\cG_i$ in
$\icp_{0.1,\delta/\epsilon}$. We allow the $i$-th learner to have
error proportional to $s_i$, so that every learner needs about $\sum_j
s_j/(\alpha\epsilon)$ examples. As in the product learner of
\Cref{sec:classes}, changing one example affects at most two of these
learners, and we obtain an upper bound $\softO{\sum_i
s_i/(\alpha\epsilon)}$ (\Cref{thm:direct}). If the same tuple also
packs into $\cH$, then $\sum_i s_i \leq
\icp_{0.1,\delta/\epsilon}(\cH,\cG)$, so this matches
\Cref{thm:lower_bound_icp} up to logarithmic factors.

\begin{restatable}{thm}{thmtightproduct}
\label{cor:product_icp}
Let \(\cH\) be a non-trivial binary class and \(\cG\) a collection of
hypothesis classes. Suppose a tuple $\br{\cG_1,\ldots,\cG_k}$ with
$\cG_i\in\cG$ satisfies the conditions in \Cref{thm:direct} and is
independently packable into $\cH$. Let $0<\epsilon\leq0.1$, $0\leq\delta\leq\alphazero\epsilon/64$, $0<\beta\leq1/3$ and $0<\alpha\leq\alphazero/(1536e)$. Then,
\[
 \mpriv{\cH}{\epsilon}{\delta}{\alpha}{\beta}
 =
 \widetilde\Theta\!\br{
 \frac{\icp_{0.1,\delta/\epsilon}(\cH,\cG)}
      {\alpha\epsilon}
 }.
\] In particular, this applies to the general product class.
\end{restatable}

For a general Boolean aggregation \(\cF\), the labels of $\cH$ need
not be labels for the classes $\cG_i$. For example, if $\cF$ is AND, a
label $0$ only says that some $\widetilde g_i$ is $0$, not which one.
We use the closure algorithm of \citet{AlonEtAl2020}. It
learns the components one at a time, each from a fresh block of
examples. In each round, it selects labels for the next component so
that, combined by \(\cF\) with the components learned so far and
suitable hypotheses for the remaining ones, they agree with most of
the observed labels. It then learns the next component privately from
these new labels. For small enough $\delta$, this gives an upper bound
that is larger than the lower bound by at most a factor $\VC(\cH)^3$,
up to logarithmic factors (\Cref{thm:boolean_aggr}).

%% file: contents/02_related_work.tex
Our lower bounds build primarily on the threshold lower bound of
\citet{AlonEtAl2022} and the representation-dimension bounds of
\citet{BeimelNS19,GhaziKM21}. For thresholds on an ordered domain of
size $N$, \citet{AlonEtAl2022} proved that improper private learning
requires $\Om{\logstar N}$ examples. Using a quantitative form of
Shelah's bound, they derive a $\Om{\logstar D}$ lower bound for a
class of Littlestone dimension $D$. Together with the non-private VC
lower bound of \citet{BlumerEtAl1989}, it gives $\Om{v+\logstar D}$.  For thresholds,
\citet{CohenEtAl2023} obtained an upper bound with nearly matching
dependence on $N$ using interior-point methods. For general classes,
\citet{BunLivniMoran2020} first showed the qualitatiev result that
online learnability implies learnability under approximate DP. However
their sample complexity scaled doubly exponentially in $D$.
\citet{GhaziEtAl2021} gave a learner using $\softO{D^6}$ examples,
\citet{Lyu2025} improved this to $\softO{D^5}$, and \citet{moran2026}
recently obtained $\bigO{D\log D}$. Nevertheless, these upper bounds
are still far from the lower bound of \(\Om{v+\logstar D}\).

\citet{GolowichLivni2021,LiWangYe2025}, and \citet{Lyu2025}
give mistake upper bounds for private online learning that depend on
the Littlestone dimension. Complementary lower bounds, including
bounds with linear dependence on this dimension, were proved by
\citet{DmitrievEtAl2024}, \citet{CohenEtAl2024}, \citet{LiWangYe2024},
and \citet{Sanyal2026}. However, these online mistake lower bounds do
not directly imply sample-complexity lower bounds for private PAC
learning.

Another line of lower bounds was developed for pure differential
privacy.~\citet{BeimelNS19} introduced probabilistic representations
and showed that \(\prdim\) characterizes the sample complexity of
pure-private learning.~\citet{FeldmanXiao2014} related representation
dimension to randomized one-way communication complexity. They showed
that it is at least of the order of the Littlestone dimension and can
be arbitrarily larger, giving separations between pure and approximate
private learning.  Although their~\citep{BeimelNS19} characterisation
was meant for pure DP, the lower-bound argument can be extended to
approximate DP, as stated in \Cref{lem:prdim_lower}; it gives
$\Om{\min\bc{\prdim(\cH),\log(\epsilon/\delta)}/(\alpha\epsilon)}$.
\citet{GhaziKM21} prove a lower bound of the same form at constant
accuracy in the more general setting of user-level privacy.

In proper learning, the output must belong to the class being learned.
For proper private learning, \citet{BunEtAl2015} proved a lower bound
for multidimensional thresholds that is multiplicative in the
dimension and the iterated logarithm of the domain size.
\citet{BunNissimStemmer2016} studied private multi-learning, where
each example carries labels from several target concepts and the
learner must output an accurate hypothesis for each concept. Our
setting is an improper learning problem for a  single binary
classification problem containing independently packed classes on
disjoint parts of its domain. Similar in spirit
to~\citet{BunEtAl2015}, we show that the component learning costs add
under this guarantee.

\citet{AlonEtAl2020} established closure properties for private
learning by constructing learners for Boolean combinations from
learners for the component classes. Our covering upper bounds uses their algorithm. Recently,~\citet{karni2026private}
improved the sample bounds for learning Boolean combinations using
proper component learners and obtained improved upper bounds for
halfspaces. For boxes, we get the upper bound
from~\citet{SadigurschiStemmer2021}, improved
by~\citet{CohenEtAl2023}, \citet{KaplanMMS22} gave private algorithms
for learning polygons in the plane, and \citet{moran2026} gave the
learner for linear subspaces. For classes of VC dimension one,
\citet{Yan2025} obtained a private learner with sample complexity
$\softO{\logstar\br{\max\bc{2,D}}}$, nearly matching the known lower bound in
its dependence on $D$.\looseness-1

%% file: appendix/prelim.tex
\section{Preliminaries}
\label{sec:prelims}

A class is a nonempty set $\cH \subseteq \bc{0,1}^{\cX}$ of hypotheses
on a domain $\cX$. A database of size $n$ is a sequence
$S = ((x_1, y_1), \cdots, (x_n, y_n)) \in (\cX \times \bc{0,1})^n$.
The empirical error of a function $h : \cX \to \bc{0,1}$ on $S$ is
\(\err_S(h) = \frac{1}{n}\abs{\bc{i \in [n] : h(x_i) \neq y_i}}\).
For a distribution $\mu$ over $\cX \times \bc{0,1}$, the population
error of $h$ is $\err_\mu(h) = \Pr_{(x,y) \sim \mu}\bs{h(x) \neq y}$.
The database $S$ is realizable by $\cH$ if $\err_S(h) = 0$ for some
$h \in \cH$, and $\mu$ is realizable by $\cH$ if $\err_\mu(h) = 0$
for some $h \in \cH$. We write $S \sim \mu^n$ for a database of $n$
independent examples from $\mu$. For our lower bounds, we assume
$\cH$ is non-trivial in the following sense and each packed class
contains at least two hypotheses.

\begin{definition}
    [Non-trivial class]
    \label{def:non_trivial}
    A class $\cH \subseteq \bc{0,1}^{\cX}$ is non-trivial if there
    exist $h_1, h_2 \in \cH$ and $x, x' \in \cX$ such that
    $h_1(x) = h_2(x)$ and $h_1(x') \neq h_2(x')$.
\end{definition}

\noindent Next we define the notion private pac learning. For this, we
recall the standard definitions of differential privacy and PAC
learning. Two databases of the same size are neighbors if they differ in exactly one example. For $\epsilon > 0$ and $\delta \in [0,1)$, a randomized algorithm $\cM$ is $(\epsilon, \delta)$-differentially private, or $(\epsilon, \delta)$-DP, if for all neighbors $S, S'$ and every event $E$,
\[
\Pr\bs{\cM(S) \in E} \leq e^{\epsilon}\Pr\bs{\cM(S') \in E} + \delta.
\]
It is purely private if $\delta = 0$. A learner is a randomised
algorithm that maps a database to a function \(\cX\to\bc{0,1}\). In
this work, we only consider improper learners, where the output need
not belong to the class being learned. For $\alpha, \beta \in (0,1)$,
the learner $\cM$ is an $(\alpha, \beta)$-PAC learner for $\cH$ with
$n$ examples if, for every distribution $\mu$ realizable by $\cH$,
\[
\Pr_{S \sim \mu^n, \cM}\bs{\err_\mu(\cM(S)) > \alpha} \leq \beta.
\]
Then, an \((\epsilon,\delta)\)-DP, \((\alpha, \beta)\)-PAC learner for
\(\cH\) is a learner that is both \((\epsilon, \delta)\)-DP and
\((\alpha, \beta)\)-PAC for \(\cH\). The private sample complexity
$\mpriv{\cH}{\epsilon}{\delta}{\alpha}{\beta}$ is the least $n$ such
that some $(\epsilon, \delta)$-DP, $(\alpha, \beta)$-PAC learner for
$\cH$ with $n$ examples exist. If $\cH$ contains two hypotheses, then $\mpriv{\cH}{0.1}{\delta}{\alpha}{1/4}\geq2$ for $\delta,\alpha\leq1/4$. At a point $x$ where they differ, at most one example cannot distinguish the point masses on $(x,0)$ and $(x,1)$, as $e^{0.1}/4+\delta<3/4$. Next, we state some standard facts
about DP \citep{DworkEtAl2006}.

\begin{lemma}[Post processing \cite{DworkEtAl2006}]
\label{lem:postprocess}
    Let $\cM:\cX\rightarrow \cO$ be a $(\epsilon,\delta)$-DP
    algorithm. Let $f:\cO\rightarrow \cO'$ be an arbitrary randomized
    mapping, then $f\circ \cM: \cX \rightarrow \cO'$ is also
    $(\epsilon,\delta)$-DP.
\end{lemma}

\begin{lemma}[Basic composition
     \cite{DworkEtAl2006}]\label{lem:composition} Let $d \in \cX$ be a
     dataset and $\cM_1, \cdots, \cM_k$ be $k$ algorithms such that
     $\cM_i: \cX \rightarrow \cO_i$ preserves
     $(\epsilon_i,\delta_i)$-DP. Then the composed algorithm $\cM: \cX
     \rightarrow \prod_{i=1}^k \cO_i$ defined as $\cM(d) = (\cM_1(d),
     \cdots, \cM_k(d))$ preserves $\br{\sum_{i=1}^k{\epsilon_i},
     \sum_{i=1}^k{\delta_i}}$-DP.
\end{lemma}

\begin{lemma}[Group privacy]
\label{lem:group_privacy}
    Let $\cM:\cX\rightarrow \cO$ be an $(\epsilon,\delta)$-differentially private algorithm, and let $d, d' \in \cX$ be datasets that differ in $r$ examples. Then for every event $E \subseteq \cO$,
    \[
    \Pr\bs{\cM(d) \in E} \leq e^{r\epsilon}\Pr\bs{\cM(d') \in E} + \delta\sum_{s=0}^{r-1}e^{s\epsilon}.
    \]
    In particular, $\cM$ satisfies $\br{r\epsilon,
    re^{(r-1)\epsilon}\delta}$-DP for datasets that differ in at most
    $r$ examples.
\end{lemma}

\begin{lemma}[Parallel composition]\label{lem:parallel_composition}
    Let $d = (d_1, \cdots, d_k) \in \cX$ be a dataset split into $k$ blocks at fixed positions. For $i \in [k]$, let $\cM_i$ be an algorithm that takes as input the block $d_i$ and the outputs $o_1, \cdots, o_{i-1}$ of $\cM_1, \cdots, \cM_{i-1}$, and produces an output $o_i \in \cO_i$, such that for all fixed $o_1, \cdots, o_{i-1}$, the algorithm $\cM_i(\cdot\,; o_1, \cdots, o_{i-1})$ preserves $(\epsilon,\delta)$-DP. Then the algorithm $\cM: \cX \rightarrow \prod_{i=1}^k \cO_i$ defined as $\cM(d) = (o_1, \cdots, o_k)$, with fresh independent randomness at each step, preserves $(\epsilon, \delta)$-DP.
\end{lemma}

\medskip\noindent\textbf{Dimensions.} A set $Z \subseteq \cX$ is shattered by $\cH$ if every labeling of $Z$ is realized by some $h \in \cH$, and the VC dimension $\VC(\cH)$ is the largest size of a shattered set. A mistake tree of depth $d$ is a complete binary tree of depth $d$ whose internal nodes are labeled by points of $\cX$; a root-to-leaf path corresponds to the labels $0$ or $1$ given to the points along it, according to whether it continues to the left or to the right child. The tree is shattered by $\cH$ if every root-to-leaf path is realized by some $h \in \cH$, and the Littlestone dimension $\ldim(\cH)$ is the largest depth of a shattered mistake tree. As a shattered mistake tree of depth $d$ needs $2^d$ distinct hypotheses, $\ldim(\cH) \leq \log_2\abs{\cH}$.

A probabilistic representation of $\cH$ is a distribution $\cQ$ over finite sets of functions $\cX \to \bc{0,1}$ such that, for every $h \in \cH$ and every distribution $\mu$ over $\cX \times \bc{0,1}$ realized by $h$, a random set $\cR \sim \cQ$ contains a function with population error at most $1/4$ with probability at least $3/4$. Its size is the largest value of $\log\abs{\cR}$ over the support of $\cQ$, and the probabilistic representation dimension $\prdim(\cH)$ is the least size of a probabilistic representation of $\cH$ \citep{BeimelNS19}.~\citet{FeldmanXiao2014} show that $\prdim(\cH)$ is at least a universal constant times $\ldim(\cH)$.

%% file: contents/03_product_union_classes.tex
In this section, we construct two simple classes which have the same
VC dimension and Littlestone dimension but different private sample
complexities. Our result uses the standard threshold class on
\(\bs{N}=\bc{1,\ldots,N}\),
\[
  \Thr{N}
  =
  \bc{h_t:h_t(x)=\one\bc{x\leq t},\ t\in\bc{0,\ldots,N}}.
\]
It has VC dimension one and Littlestone dimension
\(\floor{\log_2(N+1)}\) (\Cref{lem:threshold-dimensions}). We now
construct the two classes:~\emph{Product class} \(\ProdClass_{v,D}\)
and~\emph{Union class} \(\UnionClass_{v,D}\) both of which have VC
dimension \(v\) and Littlestone dimension \(D\).  \citet{AlonEtAl2022}
showed that, \[m=\mpriv{\Thr{N}}{0.1}{\delta}{1/16}{1/16}=\Om{\logstar
N}\] whenever \(\delta\leq c/(m^2\log m)\), for a sufficiently small
universal constant \(c>0\). This lower bound holds even for improper
learners. We ask whether a class can contain several independent
thresholds whose costs add up. We compare a product class, in which
their positions are chosen independently, with a union class, in which
the target belongs to only one component. Fix integers \(1\leq v\leq
D\). Choose \(r_i\in\bc{\floor{D/v},\ceil{D/v}}\) with \(\sum_{i=1}^v
r_i=D\), and set \(N_i=2^{r_i}\).

\noindent\textbf{Product class \(\ProdClass_{v,D}\).}
Let \(X_i=\bc{i}\times\bs{N_i}\) and
\(X_{\mathrm P}=\bigcup_{i=1}^vX_i\). Define
\begin{equation}
  \ProdClass_{v,D}
  =
  \bc{
    h_t:h_t(i,x)=\one\bc{x\leq t_i},\
    0\leq t_i\leq N_i\text{ for every }i
  }.
  \label{eq:product-class}
\end{equation}

\noindent\textbf{Union class \(\UnionClass_{v,D}\).} Let \(A\) contain
\(v\) points, let \(B=\bs{2^D-1}\), and take them to be disjoint. Let
\(\cF_v\) be the full Boolean class on \(A\), extended by zero on
\(B\) and consider \(\Thr{2^D-1}\) to be the threshold class on \(B\)
extended by zero on \(A\). Define
\begin{equation}
  \UnionClass_{v,D}
  =
  \cF_v\cup\Thr{2^D-1}.
  \label{eq:union-class}
\end{equation}
In~\(\ProdClass_{v,D}\), a hypothesis chooses one threshold
independently on each \(X_i\) and in \(\UnionClass_{v,D}\), a target
is either an arbitrary labeling of \(A\) or a threshold on \(B\). Let
\(
  L_{\mathrm P}
  =
  \logstar\!\br{\max\bc{2,\frac Dv}},
  L_{\mathrm U}
  =
  \logstar\!\br{\max\bc{2,D}}.
\)

\begin{restatable}[Product versus union]{theorem}{productuniontheorem}
\label{thm:two-families}
There exist universal constants
\(\epszero,\alphazero,\betazero\in(0,1)\) and \(c_0>0\) such that, for
every \(1\leq v\leq D\), with \(\delta_{v,D}=c_0/\br{v+L_{\mathrm
U}}, \) both \(\ProdClass_{v,D}\) and \(\UnionClass_{v,D}\) have VC
dimension \(v\) and Littlestone dimension \(D\). At privacy parameters
\((\epszero,\delta_{v,D})\) and accuracy parameters
\((\alphazero,\betazero)\), their sample complexities are
\(
  \widetilde\Theta(vL_{\mathrm P})\) and \(
  \widetilde\Theta(v+L_{\mathrm U}),
\)
respectively.
\end{restatable}

For the product class, one point from each \(X_i\) can be shattered,
while two points from the same \(X_i\) cannot. Its VC dimension is
therefore \(v\). The Littlestone dimensions of the \(v\) thresholds
add to \(\sum_i r_i=D\). For the union class, \(\cF_v\) gives VC
dimension \(v\), and the threshold component gives Littlestone
dimension \(D\). The reverse Littlestone bound follows from
\(
  \abs{\UnionClass_{v,D}}
  =
  2^v+2^D-1
  <
  2^{D+1}.
\)

In~\Cref{app:classes}, we prove \(\Om{vL_{\mathrm P}}\) lower bound for the product class by packing thresholds.
For the union class, VC lower bound gives \(\Om v\), while
the threshold component gives \(\Om{L_{\mathrm U}}\). Together these
give \(\Om{v+L_{\mathrm U}}\).

For the product upper bound, learn one threshold on each \(X_i\) with
target error \(\alpha/v\), and combine the outputs. Examples outside
\(X_i\) are replaced by a fixed zero-labeled point; changing one
original example then affects at most two threshold learners. For the
union upper bound, learn one candidate from \(\cF_v\), one candidate
from the threshold component, and privately select between them using
held-out data. These constructions give the corresponding upper
bounds. The full proofs are in \Cref{app:classes}.

To make the separation explicit, let \(T_0=2\), \(T_{j+1}=2^{T_j}\),
and set \(v_s=s\) and \(D_s=sT_{s-1}\).  Then \(L_{\mathrm
P}=L_{\mathrm U}=\Theta(s)\). The product class requires \(\Om{s^2}\)
examples, whereas the union class has an \(\bigO{s\log^2s}\)-sample
learner. Their sample complexities differ by \(\Om{s/\log^2s}\), which
is larger than every fixed power of \(\log(v_s+\logstar D_s)\). Thus
VC and Littlestone dimensions do not determine private sample
complexity, even up to polylogarithmic factors.

%% file: contents/independent_class_packings.tex
\section{Independent Class Packings}
\label{sec:icp}

In this section we prove our main result, that $\icp$ of a class $\cH$ with respect to $\cG$ gives a lower bound on the private sample complexity.

\thmLowerBoundICP*

\noindent The proof follows the outline in \Cref{sec:introduction}. \Cref{lem:hard_dist} gives, for each packed class, a distribution over databases on which every private algorithm with too few examples has large empirical error. \Cref{lem:padding} shows that the hard instances of all packed classes together form a realizable database for $\cH$, which stays realizable after adding copies of a padding point, and shows how to obtain a padding point by replacing one packed class $\cG_j$ with a subclass $\cG_j^{(y,b)}$. \Cref{lem:direct_sum} compares the errors forced on the hard instances with the error guarantee of an algorithm for $\cH$. \Cref{lem:resample,lem:duplicate} turn a learner for $\cH$ into the private algorithm used in \Cref{lem:direct_sum}.

We first show that, for every database size below the private learning
sample complexity, there is one distribution over realizable databases
on which every private algorithm has large expected empirical error.

\begin{restatable}[Hard distribution]{lem}{lemHardDistribution}
    \label{lem:hard_dist}
    Let $\epsilon \in (0,1]$, $ \alphazero \in (0,1/4]$, $\delta \in [0,
    \alphazero / 8]$ and $\bar{\alpha} = \alphazero / (8e)$. Let $\cG
    \subseteq \bc{0,1}^\cY$ be a class over a finite domain. Then, for
    every $1 \leq q < \mpriv{\cG}{\epsilon}{\delta}{\alphazero}{1/4}$,
    there exists a distribution $\mu_{\cG,q}$ over databases of size
    $q$ realizable by $\cG$ such that for every $(\epsilon,\delta)$-DP
    algorithm $\cM$ on $q$ records\looseness-1
    \[
    \bE_{S \sim \mu_{\cG, q}, \cM} \bs{ \err_{S}(\cM(S)) } > \bar{\alpha}.
    \]
\end{restatable}

\begin{proof}[Proof of~\Cref{lem:hard_dist}]
    We first show that every $(\epsilon,\delta)$-DP algorithm $\cM$ on
    $q$ records has a database $S$ realizable by $\cG$ with
    $\bE_{\cM}\bs{\err_S(\cM(S))} > \bar{\alpha}$. Suppose instead
    that some such algorithm has expected empirical error at most
    \(\bar\alpha\) on every realisable database of size \(q\). For any
    distribution $\mu$ realizable by $\cG$ and $S \sim \mu^q$, \Cref{lem:emp_to_pop_error} gives
    \[
    \bE \bs{\err_{\mu}(\cM(S))} \leq e^\epsilon \bar{\alpha} + \delta \leq \frac{\alphazero}{4}.
    \]
    By Markov's inequality, $\Pr\bs{\err_{\mu}(\cM(S)) > \alphazero}
    \leq 1/4$. Thus, $\cM$ is an $(\epsilon, \delta)$-DP, $(\alphazero,
    1/4)$-PAC learner for $\cG$ using $q$ samples, contradicting the
    assumption $q < \mpriv{\cG}{\epsilon}{\delta}{\alphazero}{1/4}$.

     We now use the minimax theorem to find one distribution that
works for every $\cM$. We can represent each output hypothesis by its
labels on $\cY$, which preserves privacy and empirical error.  An
algorithm is then described by a finite table of probabilities
$\cM\br{f\mid S}=\Pr\bs{\cM(S)=f}$, with a row for every labeled
database $S$ of size $q$ and a column for every $f\in\bc{0,1}^{\cY}$.
Let $\cT$ be the set of these tables satisfying
$(\epsilon,\delta)$-DP. Each row comprises non-negative entries and
sums to one, and privacy gives finitely many non-strict linear
inequalities in the entries. These conditions imply that $\cT$ is
compact and convex.

Let $\cR$ be the finite set of databases of size $q$ realizable
by $\cG$, and set $u(\cM,S)=\bE_{\cM}\bs{\err_S(\cM(S))}$.
By the first part of the proof, $\max_{S\in\cR}u(\cM,S)>\bar\alpha$
for every $\cM\in\cT$.  Since $\cR$ is finite, this maximum is
continuous in the probabilities specifying $\cM$. Its minimum over the
compact set $\cT$ is therefore attained and is strictly greater than
\(\bar\alpha\). Let $\Delta(\cR)$ be the set of distributions over
$\cR$.  The quantity $\bE_{S\sim\nu}u(\cM,S)$ is continuous and linear
in each of $\cM$ and $\nu$. Both $\cT$ and $\Delta(\cR)$ are compact
and convex. Maximizing over $\nu\in\Delta\br{\cR}$ is equivalent to
choosing a database that maximizes $u(\cM,S)$, so the minimax theorem
gives
\[
\min_{\cM\in\cT}\max_{S\in\cR}u(\cM,S)
=
\max_{\nu\in\Delta(\cR)}\min_{\cM\in\cT}
\bE_{S\sim\nu}u(\cM,S).
\]
The left side is greater than $\bar\alpha$. Taking $\mu_{\cG,q}$ to be
a distribution attaining this maximum gives
\[
\bE_{S\sim\mu_{\cG,q},\cM}\bs{\err_S(\cM(S))}>\bar\alpha,
\]
for every $(\epsilon,\delta)$-DP algorithm $\cM$ on $q$ records,
as required.

\end{proof}

\noindent The hard distribution depends on the class $\cG$ and the database size $q$, which suffices as we fix $q_i$ for each packed class $\cG_i$. To fill the rest of a database built from these hard instances, we need points that can be added without affecting realizability, which we call \emph{padding} points~(\Cref{def:padding_point}).

\begin{definition}
    [Padding point]
    \label{def:padding_point}
    For $\cH \subseteq \bc{0,1}^{\cX}$, let $(\cG_1, \cdots, \cG_k) \sqsubseteq \cH$ via $\Phi_1, \cdots, \Phi_k$. A labeled example $(x_0, y_0) \in \cX \times \bc{0,1}$ is a padding point if every $(g_1, \cdots, g_k) \in \prod_{i=1}^k \cG_i$ has an extension $h \in \cH$ with $h \circ \Phi_i = g_i$ for all $i \in [k]$ and $h(x_0) = y_0$.
\end{definition}

\begin{restatable}[Padding]{lem}{lemPadding}
    \label{lem:padding}
    Let $\epsilon \in (0,1]$, $\alphazero \in (0, 1/4]$, $0 \leq \delta \leq\alphazero/8$. For $\cH \subseteq \bc{0,1}^{\cX}$ and $\cG_i \subseteq \bc{0,1}^{\cY_i}$ for all $i \in [k]$, let $(\cG_1, \cdots, \cG_k) \sqsubseteq \cH$ with $\Phi_i$ as the corresponding maps and let $s_i := \mpriv{\cG_i}{\epsilon}{\delta}{\alphazero}{1/4}$. For a database $S_i$ over $\cY_i$, we write $\Phi_i(S_i)$ for the database obtained by replacing every example $(y, \ell)$ of $S_i$ with $(\Phi_i(y), \ell)$. For $y \in \cY_j$ and $b \in \bc{0,1}$, we write $\cG_j^{(y,b)} = \bc{g \in \cG_j : g(y) = b}$, and for a real number $a$, we write $a_+ = \max\bc{a, 0}$.
    \begin{enumerate}[]
        \item For every integer $1 \leq Q \leq \sum_{i = 1}^k (s_i -
        1)_{+}$, there exist integers $q_1, \cdots, q_k$ such that $0
        \leq q_i \leq (s_i - 1)_{+}$ and $\sum_{i = 1}^k q_i = Q$. For
        each \(i\) with $q_i > 0$, let $S_i$ be any database of size
        $q_i$ realizable by $\cG_i$, and let $S_i$ be empty when $q_i
        = 0$. Then $\cH$ realizes the database $\Phi_1(S_1) \cdots
        \Phi_k(S_k)$.
        \item If $(x_0, y_0)$ is a padding point (\Cref{def:padding_point}) then appending any number of copies of $(x_0, y_0)$ to the database in (1) keeps it realizable.
        \item Let $j \in [k]$ and $y \in \cY_j$.
        \begin{enumerate}
            \item At least one of $\cG_j^{(y,0)}$ and $\cG_j^{(y,1)}$ is nonempty.
            \item If $\cG_j^{(y,b)} \neq \emptyset$, replacing $\cG_j$
            with $\cG_j^{(y,b)}$ preserves packability in $\cH$ via
            the same maps $\Phi_1,\cdots,\Phi_k$ and $(\Phi_j(y), b)$
            is a padding point for it.
            \item There is $b\in \bc{0,1}$ with $\cG_j^{(y,b)} \neq \emptyset$ and
            \[
            \mpriv{\cG_j^{(y,b)}}{\epsilon}{\delta}{\alphazero/8}{1/4} \geq \frac{1}{4} \br{ s_j - \frac{101}{\alphazero \epsilon} }.
            \]
        \end{enumerate}

    \end{enumerate}
\end{restatable}

\begin{proof}[Proof of~\Cref{lem:padding}]
We first prove (1). For $i=1,\cdots,k$, let
$q_i=\min\bc{(s_i-1)_+,Q-\sum_{i'<i}q_{i'}}$.
As $Q\leq\sum_i(s_i-1)_+$, these choices satisfy
$0\leq q_i\leq(s_i-1)_+$ and $\sum_iq_i=Q$.
For every $i$, choose $g_i\in\cG_i$ realizing $S_i$, where any
$g_i\in\cG_i$ works if $S_i$ is empty. By
\Cref{def:packing_classes}, there is $h\in\cH$ with
$h\circ\Phi_i=g_i$ for every $i$. Thus $h$ realizes
$\Phi_1(S_1)\cdots\Phi_k(S_k)$.

For (2), the definition of a padding point lets us choose this
extension with $h(x_0)=y_0$. This $h$ also realizes any number of
added copies of $(x_0,y_0)$.

We now prove (3). Fix $j\in[k]$ and $y\in\cY_j$. As $\cG_j$ is
nonempty, at least one of $\cG_j^{(y,0)}$ and $\cG_j^{(y,1)}$ is
nonempty. Replacing $\cG_j$ with any nonempty $\cG_j^{(y,b)}$
preserves packability via the same maps, as it is a subclass of
$\cG_j$. Every corresponding extension satisfies
$h(\Phi_j(y))=g_j(y)=b$, so $(\Phi_j(y),b)$ is a padding point.

For each nonempty $\cG_j^{(y,b)}$, let
$n_b=\mpriv{\cG_j^{(y,b)}}{\epsilon}{\delta}{\alphazero/8}{1/4}$
and $C=101$. If one subclass is empty, the other equals $\cG_j$,
so $n_b\geq s_j$ and (3c) follows. Suppose both are nonempty.
It suffices to show that
\begin{equation}
s_j\leq2(n_0+n_1)+\frac{C}{\alphazero\epsilon},
\end{equation}
as (3c) then follows by choosing $b$ with $n_b=\max\bc{n_0,n_1}$.

For $b\in\bc{0,1}$, fix an $(\epsilon,\delta)$-DP,
$(\alphazero/8,1/4)$-PAC learner $\cA_b$ for $\cG_j^{(y,b)}$
using $n_b$ examples. Let $n=\ceil{24\ln(64)/(\epsilon\alphazero)}$.
Our learner $\cA$ uses $N=2n_0+2n_1+n$ examples. It runs each
$\cA_b$ twice on disjoint blocks of size $n_b$, with independent
randomness, obtaining $h_{b,1},h_{b,2}$. On the remaining block $V$
of size $n$, it applies \Cref{lem:priv_selection_union} to the four
candidates and returns the selected hypothesis.

\emph{Privacy.} Changing a training example affects the output only
through one $(\epsilon,\delta)$-DP learner, followed by
post-processing. Changing an example in $V$ leaves the candidates
fixed, and selection is $(\epsilon,0)$-DP by
\Cref{lem:priv_selection_union}. Thus $\cA$ is $(\epsilon,\delta)$-DP.

\emph{Utility.} Let $c\in\cG_j$ be the target and $\mu$ a distribution
on $\cY_j$, and w.l.o.g. assume $c(y)=0$.
Each of $h_{0,1},h_{0,2}$ has error at most $\alphazero/8$ with
probability at least $3/4$, independently. Thus some candidate has
this error except with probability at most $1/16$. The block $V$ is
independent of the candidates. If a candidate has error at most
$\alphazero/8$, \Cref{lem:priv_selection_union}, with $r=4$,
$\alpha=\alphazero$ and $\eta=3/16$, selects a hypothesis of error at
most $\alphazero$ except with probability at most $3/16$.
By a union bound, $\cA$ fails with probability at most
$1/16+3/16=1/4$, so $s_j\leq N=2(n_0+n_1)+n$.
Since $24\ln64<100$ and $\epsilon\alphazero\leq1$,
$n\leq101/(\epsilon\alphazero)$, which completes the proof.
\end{proof}

Next, we place the hard instances of the packed classes in one database for $\cH$ and bound the total number of mistakes a private algorithm makes on it.

\begin{restatable}{lem}{lemDirectSum}
\label{lem:direct_sum}
Assume the setting and notation of \Cref{lem:padding}, with each
$\cY_i$ finite, and set $\bar\alpha=\alphazero/(8e)$.  Let $Q,M$ be
positive integers satisfying \(Q \leq M\) and \(Q \leq \sum_{i=1}^k
(s_i-1)_+\). If $Q<M$, also assume the packing has a padding point.
Let $\gamma,\eta\in[0,1]$. If an $(\epsilon,\delta)$-DP algorithm has
empirical error at most $\gamma$ with probability at least $1-\eta$ on
every database of size $M$ realizable by $\cH$, then
\[
(\bar\alpha-\eta)Q<\gamma M.
\]
\end{restatable}

\begin{proof}[Proof of~\Cref{lem:direct_sum}]
    Let $\cM$ be an algorithm satisfying the assumptions. By (1) of
    \Cref{lem:padding}, choose integers $q_1,\cdots,q_k$ with
    $0 \leq q_i \leq (s_i-1)_+$ and $\sum_i q_i=Q$.
    If $Q<M$, fix a padding point $(x_0,y_0)$.
    For each $i$ with $q_i>0$, draw $S_i$ independently from
    $\mu_{\cG_i,q_i}$ in \Cref{lem:hard_dist}, and let $S_i$ be empty
    when $q_i=0$. Let $S$ be the concatenation
    $\Phi_1(S_1)\cdots\Phi_k(S_k)$. If $Q<M$, append $M-Q$ copies
    of $(x_0,y_0)$ to $S$.
    By \Cref{lem:padding}, $S$ has size $M$ and is realizable by $\cH$.

    Fix $i$ with $q_i>0$. On any database $S_i$ of size $q_i$, let
    $\cM_i$ draw the other databases independently as above, construct
    $S$, run $\cM$ to obtain $h$, and return $h\circ\Phi_i$.
    For every fixed choice of
    the other databases, changing one example of $S_i$ changes one
    example of $S$. Thus $\cM_i$ is $(\epsilon,\delta)$-DP.
    The pair $(S_i,\cM_i(S_i))$ has the same distribution as
    $(S_i,h\circ\Phi_i)$ for $h=\cM(S)$. By \Cref{lem:hard_dist},
    $\bE\bs{\err_{S_i}(h\circ\Phi_i)}>\bar\alpha$.
    Summing over $i$ with $q_i>0$ gives
    \[
    \bE\bs{\sum_{i:q_i>0} q_i\err_{S_i}(h\circ\Phi_i)}
    > \bar\alpha Q.
    \]

    The sum counts the mistakes on the hard blocks, so it is at most
    the number of mistakes $h$ makes on $S$. On the success event,
    this is at most $\gamma M$. On the failure event, there are at most
    $Q$ mistakes on the hard blocks. Since the failure probability is
    at most $\eta$, we have
    \[
    \bE\bs{\sum_{i:q_i>0} q_i\err_{S_i}(h\circ\Phi_i)}
    \leq \gamma M+\eta Q.
    \]
    Combining the two bounds completes the proof.
\end{proof}

Before proving \Cref{thm:lower_bound_icp}, we state two standard
results.  \Cref{lem:resample} converts a private PAC guarantee to an
empirical guarantee on every realizable database using the resampling
lemma of \citet{BunEtAl2015} (\Cref{lem:secrecy_of_sample}), and also
reduces the learner's failure probability.  \Cref{lem:duplicate} is a
reduction that trades off privacy for sample size. This provides the
source of the factor $1/\epsilon$ in \Cref{thm:lower_bound_icp}. The
proof of \Cref{lem:resample} is in \Cref{app:icp-proofs}.  We also use
the standard lower bound of $\Om{1/(\alpha\epsilon)}$ for non-trivial
classes when $\delta \leq \epsilon$ (\Cref{lem:non_trivial_bound}).

\begin{restatable}{lem}{lemResample}
\label{lem:resample}
Let $0<\epsilon\leq1$, and $\cM$ be an
$(\epsilon,\delta)$-DP, $(\alpha,\beta)$-PAC learner for $\cH$
using $n\geq1$ examples.
\begin{enumerate}
    \item For every $M\geq9n$, there is an $(\epsilon,\delta)$-DP
    mechanism on $M$ examples with empirical error at most $\alpha$
    on every realizable database, except with probability at most
    $\beta$.
    \item If $\beta\leq1/3$, then for every $\eta\in(0,1)$ and odd
    integer $T\geq72\ln(1/\eta)$, there is an $(\epsilon,\delta)$-DP,
    $(12\alpha,\eta)$-PAC learner for $\cH$ using $Tn$ examples.
\end{enumerate}
\end{restatable}

\begin{restatable}{lem}{lemDuplicate}
\label{lem:duplicate}
Let $0<\epsilon\leq0.1$, $\ell=\floor{0.1/\epsilon}$,
and let $\cM$ be an $(\epsilon,\delta)$-DP mechanism on $\ell m$
examples, where $m\geq1$. Repeating each of the $m$ input examples
$\ell$ times and running $\cM$ gives a $(0.1,\delta/\epsilon)$-DP
mechanism $\cM'$ with the same empirical error and failure
probability guarantees on realizable databases.
\end{restatable}

We now prove \Cref{thm:lower_bound_icp}.

\begin{proof}[Proof of~\Cref{thm:lower_bound_icp}]
Let $\cM$ be an $(\epsilon, \delta)$-DP, $(\alpha, \beta)$-PAC learner
for $\cH$ using $n$ examples.

\medskip
\noindent\textbf{Parameters.}
    We first fix
    \(
    \bar{\alpha} = \alphazero/(64e), \eta = \bar{\alpha}/2,\) and let $T$ be the smallest positive odd integer with $T \geq 72\ln(1/\eta)$, as in \Cref{lem:resample}, which is the number of independent copies we will use to reduce the failure probability of a learner from $1/3$ to $\eta$. Throughout the proof, \Cref{lem:hard_dist,lem:padding,lem:direct_sum} are used with privacy parameters $0.1$ and $\delta/\epsilon$, which are the privacy parameters of the mechanism we construct in the first step.
    Fix a tuple $(\cG_1, \cdots, \cG_k) \sqsubseteq \cH$ with $\cG_i \in \cG$ for all $i \in [k]$, let $s_i = \mpriv{\cG_i}{0.1}{\delta/\epsilon}{\alphazero}{1/4}$ and $\widetilde{s_i} = \mpriv{\cG_i}{0.1}{\delta/\epsilon}{\alphazero/8}{1/4}$ for all $i \in [k]$. We have $\widetilde{s_i} \geq s_i$, as any $(\alphazero/8, 1/4)$-PAC learner is also $(\alphazero, 1/4)$-PAC. We will lower bound $n$ by a constant multiple of
$\sum_i s_i/(\alpha\epsilon)$. Let
$c=\min\bc{\bar\alpha/(69120\,T),\alphazero/(576(10C+4\alphazero))}$,
where $C=101$.
The constant $C$ comes from (3c) of \Cref{lem:padding}, and the two terms in the minimum come from the two subcases of the case $k = 1$ below. By assumption, $\alpha \leq \alphazero/(1536e) = \bar\alpha/24$, and  $\alpha \leq 1/30$.

\medskip\noindent\textbf{Step 1: A mechanism with privacy parameters $0.1$ and $\delta/\epsilon$.}  As $\cH$ is non-trivial, $\delta \leq \epsilon$ and $\alpha < 1/30$, \Cref{lem:non_trivial_bound} applied to $\cM$ gives $n > 1/(288\alpha\epsilon) \geq 0.1/\epsilon$. Let $\ell = \floor{ 0.1/\epsilon }$, so that $1 \leq \ell \leq n$ and $\ell \geq 0.05/\epsilon$.
The second part of \Cref{lem:resample} applied to $\cM$ reduces the failure probability from $\beta \leq 1/3$ to $\eta$, at the cost of a factor $T$ in the number of examples and a factor $12$ in the error. This gives an $(\epsilon,\delta)$-DP, $(12\alpha, \eta)$-PAC learner $\cA$ for $\cH$ on $Tn$ examples. Let $m = \ceil{ 9Tn/\ell }$
so that $\ell m \geq 9Tn$. The first part of \Cref{lem:resample}, applied to $\cA$ with databases of size $\ell m$, turns the guarantee of $\cA$ for i.i.d.\ samples into a guarantee for every realizable database, which we need as the databases of the second step are not i.i.d.\ samples. The mechanism that draws $Tn$ examples uniformly with replacement from a database of size $\ell m$ and runs $\cA$ is $(\epsilon,\delta)$-DP, and on every database of size $\ell m$ realizable by $\cH$ it has empirical error at most $12\alpha$ with probability at least $1-\eta$.

Let $\cM'$ be the mechanism that, on a database of size $m$, draws
$Tn$ examples uniformly with replacement from it and runs $\cA$.
Drawing uniformly from a database is the same as drawing uniformly
from the database that repeats each of its examples $\ell$ times. Thus
$\cM'$ has the same output distribution as repeating each input
example $\ell$ times and applying the preceding mechanism. By
\Cref{lem:duplicate}, $\cM'$ is $(0.1, \delta/\epsilon)$-DP and, on
every database of size $m$ realizable by $\cH$, has empirical error at
most $12\alpha$ with probability at least $1-\eta$. Finally, as $n
\geq \ell$ and $1/\ell \leq 20\epsilon$, we have $m \leq 9Tn/\ell + 1
\leq 18Tn/\ell \leq 360\,\epsilon Tn$.

\paragraph{Step 2: Databases from hard instances.} We now run this mechanism on databases built from hard instances of the packed classes. We use the hard distributions of \Cref{lem:hard_dist} at accuracy $\alphazero/8$ rather than $\alphazero$ and privacy parameters $(0.1, \delta / \epsilon)$, to obtain empirical error of at least $\bar{\alpha}$.
These parameter choices also let us apply \Cref{lem:padding} and obtain a realizable database. Applying \Cref{lem:direct_sum} with database size $m$ to the $(0.1, \delta/\epsilon)$-DP mechanism with $\gamma = 12\alpha$ and $\eta = \bar{\alpha}/2$ compares the expected error larger than $\bar{\alpha}$ on the hard blocks with the error at most $12\alpha$ on the whole database. As $\eta < \bar{\alpha}$, it gives, for every database of size $m$ constructed as in \Cref{lem:direct_sum},
\begin{equation}
\label{eq:Q_M_relation}
Q < \frac{24\alpha}{\bar{\alpha}} m.
\end{equation}
Since $\alpha \leq \bar{\alpha}/24$, the right-hand side is at most $m$. Thus the hard blocks can fill at most a $24\alpha/\bar{\alpha}$ fraction of the database, and the remaining rows must be filled with copies of a padding point.

Next, we show that for every packable tuple $(\cG'_1,\cdots,\cG'_{k'}) \sqsubseteq \cH$ that has a padding point,
\begin{equation}
\label{eq:n_bound}
n > \frac{\bar{\alpha}}{8640\,T\alpha\epsilon}\,Q',\; \text{where} \;
Q'=\sum_i \br{ \mpriv{\cG'_i}{0.1}{\delta/\epsilon}{\alphazero/8}{1/4} - 1}_{+}.
\end{equation}

If $Q' = 0$ we are done. We have $m > Q'$; otherwise (1) of \Cref{lem:padding} gives integers $q_i$ with $\sum_i q_i = m$, no padding is required and from \Cref{eq:Q_M_relation} we get $m < m$, a contradiction. As $Q' < m$, we first obtain a realizable database of size $m$ with padding points from \Cref{lem:padding} and applying \Cref{eq:Q_M_relation} to this database gives $Q' < 24\alpha m/\bar\alpha \leq 8640\,T\alpha\epsilon n/\bar\alpha$, where the last inequality uses $m \leq 360\,\epsilon Tn$.

\paragraph{Step 3: Wrapping up.} We start by considering a fixed tuple $(\cG_1,\dots,\cG_k)$. Dropping the components with $s_i=0$ keeps the tuple packable and does not change $\sum_i s_i$, so we may assume $s_i>0$, and hence $s_i\geq2$ by \Cref{sec:prelims}, for every $i$ (if all $s_i=0$ there is nothing to prove). Choose $j\in\argmin_i s_i$ and fix $y\in\cY_j$.
Let $b$ be given by (3c) of \Cref{lem:padding}, applied with
privacy parameters $(0.1,\delta/\epsilon)$ and accuracy
$\alphazero$. Write
$\widetilde{s_j}^{(y,b)}
=\mpriv{\cG_j^{(y,b)}}{0.1}{\delta/\epsilon}{\alphazero/8}{1/4}$.
By (3b) of \Cref{lem:padding}, replacing $\cG_j$ with
$\cG_j^{(y,b)}$ preserves packability and gives the padding
point $(\Phi_j(y),b)$. Applying \Cref{eq:n_bound} to this tuple gives

\begin{equation}
\label{eq:n_bound_padded}
n > \frac{\bar\alpha}{8640\,T\alpha\epsilon}\bs{ \sum_{i\ne j}\br{ \widetilde{s_i} - 1}_{+}
+\br{ \widetilde{s_j}^{(y,b)} - 1 }_{+} }.
\end{equation}

We first consider two cases based on $k$.

\paragraph{Case $k\ge2$.} Recall $\cG_j^{(y,b)}$ is only required for its padding point. We simply ignore the term $(\widetilde{s_j}^{(y,b)} - 1)_{+}$. For $i\neq j$ we have $(\widetilde{s_i} - 1)_+ \geq \widetilde{s_i} - 1 \geq s_i - 1 \geq s_i/2$ because $s_i \geq 2$, and $\sum_{i \neq j}s_i \geq (1-1/k) \sum_i s_i \geq \frac{1}{2}\sum_i s_i$ (we choose $j$ with the smallest $s_j$). So the quantity is at least $\frac{1}{4} \sum_i s_i$, and
\[n > \frac{\bar{\alpha}}{34560\,T\alpha\epsilon}\sum_i s_i \geq \frac{c}{\epsilon\alpha} \sum_i s_i.
\]

\paragraph{Case $k=1$.} From (3c) of \Cref{lem:padding}, which compares $\cG_j^{(y,b)}$ at accuracy $\alphazero/8$ with $\cG_1$ at accuracy $\alphazero$ for privacy parameters $0.1$ and $\delta/\epsilon$, we have $\widetilde{s_1}^{(y,b)} \geq \frac{1}{4} \br{ s_1 - 10C/\alphazero }$.
Hence
\[
(\widetilde{s_1}^{(y,b)} - 1)_+ \geq \widetilde{s_1}^{(y,b)} - 1 \geq \frac{1}{4} s_1 - \frac{1}{4} \br{\frac {10C}{\alphazero}+4}.
\]

\begin{itemize}[leftmargin=*]
\item If $s_1 \ge 2\br{ \frac{10C}{\alphazero} + 4 }$, then $(\widetilde{s_1}^{(y,b)} - 1)_{+} \geq s_1/8$, so from \Cref{eq:n_bound_padded} $n>\frac{\bar\alpha}{69120\,T\alpha\epsilon} s_1 \ge \frac{c}{\epsilon\alpha} s_1$.
\item Otherwise, the loss $10C/\alphazero$ in (3c) of \Cref{lem:padding} may exceed $s_1/4$, so \Cref{eq:n_bound_padded} may be vacuous, but $s_1$ is then bounded by a constant depending only on $\alphazero$, and \Cref{lem:non_trivial_bound} suffices. Indeed, as $\cH$ is non-trivial, $\delta \leq \epsilon$ and $\alpha \leq 1/4$, from \Cref{lem:non_trivial_bound}, $n>\frac{1}{288\,\alpha\epsilon}$, which is larger than $\frac{\alphazero}{576(10C+4\alphazero)}\cdot\frac{s_1}{\epsilon\alpha}\ge\frac{c}{\epsilon\alpha} s_1$
  as $s_1 < 2 \br{ \frac{10C}{\alphazero} + 4 }$.
\end{itemize}

Thus, $n \geq \frac{c}{\epsilon\alpha} \sum_i s_i$. Taking the supremum over packable
tuples gives
$n = \Om{ \frac{\icp_{0.1,\delta/\epsilon}(\cH, \cG)}{\epsilon\alpha} }$.
\end{proof}

%% file: contents/threshold_packing.tex
\section{Lower Bounds via Threshold Packing}
\label{sec:packing_thresholds}

We now use $\icp$ to prove lower bounds for various classes by packing thresholds.
Several familiar classes contain many disjoint ordered sets on which one hypothesis can choose threshold positions independently. Such a class $\cH$ packs several copies of a threshold class, and \Cref{thm:lower_bound_icp} adds up their sample complexities. \Cref{sec:threshold-lower-bound} proves the lower bound for a single threshold class that we add up. \Cref{sec:applications} constructs threshold packings for boxes, unions of intervals, halfspaces and convex polygons, which give better and in some cases tight lower bounds, and \Cref{sec:small_delta} uses the same packings with a different lower bound for threshold classes when $\delta$ is small.

\subsection{A Threshold Lower Bound for a Given Sample Size}
\label{sec:threshold-lower-bound}
\input{contents/04_threshold_lower_bound}

\subsection{Applications to Geometric Classes}
\label{sec:applications}

We now construct threshold packings for four classes. Recall that the class $\cB_{N,d}$ of axis-aligned boxes on $[N]^d$ consists of the indicators of products $\prod_{j=1}^d [a_j, b_j]$ with $1 \leq a_j \leq b_j \leq N$, and the class $\cH_{N,d}$ of affine halfspaces on $[N]^d$ consists of the functions $x \mapsto \one\bc{\ip{w}{x} \leq \theta}$. The class $\cI_{N,k}$ consists of unions of at most $k$ intervals on $[N]$, and the class $\cC_{N,k}$ of convex $k$-gons on $[N]^2$ consists of the indicators of intersections of at most $k$ halfplanes, restricted to the grid.

Recall that for an integer $N \geq 1$, $\Thr{N} = \bc{h_t : h_t(x) = \one\bc{x \leq t}, \; 0 \leq t \leq N}$ is the class of thresholds on $[N]$. For integers $L, k \geq 1$, we write $(\Thr{L})^k$ for the tuple with $k$ entries, each equal to $\Thr{L}$. Throughout this section we fix $\alphazero = 1/192$.
We first show that each of the four classes packs several copies of a threshold class.

\begin{lemma}[Threshold packings]
    \label{lem:threshold_packings}
    \begin{enumerate}[]
        \item For $N \geq 3$ and $d \geq 1$, $(\Thr{N-1})^d \sqsubseteq \cB_{N,d}$.
        \item For $N \geq 2k$ and $L = \floor{ N/k }$, $(\Thr{L})^k \sqsubseteq \cI_{N,k}$.
        \item For $N \geq 3$ and $d \geq 1$, $(\Thr{N-1})^d \sqsubseteq \cH_{N,d}$.
        \item For $N, k \geq 1$ with $\floor{ \sqrt{N} } \geq 2k$ and $L = \floor{ \floor{ \sqrt{N} } / k }$, $(\Thr{L})^k \sqsubseteq \cC_{N,k}$.
    \end{enumerate}
    In all four cases the length of the thresholds is at least $2$.
\end{lemma}

\begin{proof}

By \Cref{def:packing_classes}, to show $(\Thr{L})^k \sqsubseteq \cH$ for a class $\cH$, we need injective maps $\Phi_1, \cdots, \Phi_k$ from $[L]$ to the domain with pairwise disjoint images, and, for every choice of threshold positions $t_1, \cdots, t_k \in \bc{0, \cdots, L}$, a hypothesis $h \in \cH$ with
\[
h(\Phi_i(r)) = \one\bc{r \leq t_i} \quad \text{for all } i \in [k] \text{ and } r \in [L].
\]
 In each case below, the maps place $k$ ordered sequences of length $L$ in the domain, and $h$ is chosen so that it cuts the $i$-th sequence after its $t_i$-th point without affecting the other sequences.

\medskip
\noindent\textbf{Boxes.} Each sequence starts at the corner $u = (1, \cdots, 1) \in [N]^d$ and moves along one coordinate direction, and the box extends from $u$ by $t_i$ in direction $i$. Formally, with $e_i$ as the $i$-th standard basis vector,
\[
\Phi_i(r) = u + re_i, \; \text{for} \; i \in [d] \; \text{and} \; r \in [N-1].
\]
These points lie in $[N]^d$, and the images are pairwise disjoint, as $\Phi_i(r)$ differs from $u$ exactly in coordinate $i$. Given $t_1, \cdots, t_d$, let $h$ be the indicator of the box $\prod_{j=1}^d [1, 1 + t_j]$. Every coordinate of $\Phi_i(r)$ other than the $i$-th equals $1 \in [1, 1 + t_j]$, and the $i$-th equals $1 + r$, so $h(\Phi_i(r)) = \one\bc{1 + r \leq 1 + t_i} = \one\bc{r \leq t_i}$.

\medskip
\noindent\textbf{Unions of intervals.} Each sequence is a block of consecutive points, and the union keeps a prefix of each block. Formally, partition the first $kL$ points of $[N]$ into the consecutive blocks $\bc{(i-1)L + 1, \cdots, iL}$, $i \in [k]$, and let
\[
\Phi_i(r) = (i-1)L + r \; \text{for} \; r \in [L].
\]
The images are the blocks, which are pairwise disjoint. Given $t_1, \cdots, t_k$, let $h$ be the indicator of
\[
\bigcup_{i : t_i > 0} \bc{(i-1)L + 1, \cdots, (i-1)L + t_i},
\]
which is a union of at most $k$ intervals. Its restriction to the $i$-th block is the prefix of length $t_i$, so $h(\Phi_i(r)) = \one\bc{r \leq t_i}$.

\medskip
\noindent\textbf{Halfspaces.} We use the same sequences as for boxes, and choose the halfspace so that its boundary crosses the $i$-th sequence between its $t_i$-th and $(t_i + 1)$-th point. Formally, let $\Phi_i(r) = u + re_i$ as for boxes. Given $t_1, \cdots, t_d$, let
\[
h(x) = \one\bc{\sum_{i=1}^d \frac{x_i - 1}{t_i + 1/2} \leq 1},
\]
which is an affine halfspace, with $w_i = 1/(t_i + 1/2)$ and $\theta = 1 + \sum_i w_i$. At $x = \Phi_i(r)$, every summand except the $i$-th vanishes, so $h(\Phi_i(r)) = \one\bc{r/(t_i + 1/2) \leq 1} = \one\bc{r \leq t_i}$, as $r$ and $t_i$ are integers.

\medskip\noindent\textbf{Convex polygons.} We place the sequences on the parabola $z_2 = z_1^2$. A halfplane above a line through two points of the parabola contains exactly the points of the parabola outside the segment between them. So such a halfplane can remove any set of consecutive points of the parabola and no other point of it, and an intersection of $k$ such halfplanes can remove a suffix of each of $k$ blocks of consecutive points.

Formally, let $p(z) = (z, z^2)$. For $z \in [\floor{ \sqrt{N} }]$, we have $z^2 \leq N$, so these points lie in $[N]^2$. Let $\Phi_i(r) = p((i-1)L + r)$ for $i \in [k]$ and $r \in [L]$. These points are well defined as $kL \leq \floor{ \sqrt{N} }$, and the images are pairwise disjoint as $p$ is injective. Given $t_1, \cdots, t_k$, for every $i$ with $t_i < L$ let $a_i = (i-1)L + t_i + 1/2$ and $b_i = iL + 1/2$, and let
\[
\Pi_i = \bc{(z_1, z_2) \in \bR^2 : z_2 \geq (a_i + b_i) z_1 - a_i b_i}
\]
be the halfplane above the line through $p(a_i)$ and $p(b_i)$. For every $i$ with $t_i = L$, let $\Pi_i = \bc{(z_1, z_2) \in \bR^2 : z_2 \geq -1}$, which contains $[N]^2$. Let $h$ be the indicator of $\Pi_1 \cap \cdots \cap \Pi_k$ restricted to $[N]^2$, so $h \in \cC_{N,k}$. For $t_i < L$ and every $z$, the point $p(z)$ lies in $\Pi_i$ if and only if $z^2 \geq (a_i + b_i)z - a_i b_i$, that is, $(z - a_i)(z - b_i) \geq 0$. For an integer $z$ this holds if and only if $z \leq (i-1)L + t_i$ or $z \geq iL + 1$. Thus $\Pi_i$ contains every point $p(z)$ except those with $z$ in the suffix $\bc{(i-1)L + t_i + 1, \cdots, iL}$ of the $i$-th block. These suffixes lie in pairwise disjoint blocks, so for $r \in [L]$ the point $\Phi_i(r)$ lies in every $\Pi_{i'}$ with $i' \neq i$, and it lies in $\Pi_i$ if and only if $r \leq t_i$. Hence $h(\Phi_i(r)) = \one\bc{r \leq t_i}$.

\end{proof}
\noindent We now combine \Cref{lem:threshold_packings} and \Cref{cor:threshold_sc} with \Cref{thm:lower_bound_icp}.

\begin{theorem}[Private lower bounds]
    \label{cor:four_classes}
    Let $c_\delta$ be the constant of \Cref{cor:threshold_sc}. Let $0 < \epsilon \leq 0.1$, $0 \leq \delta \leq \alphazero\epsilon/64$, $0 < \alpha \leq \alphazero/(1536e)$ and $0 < \beta \leq 1/3$, and let $n$ be the number of examples used by an $(\epsilon, \delta)$-DP, $(\alpha, \beta)$-PAC learner, possibly improper, for the class in each item below.
    \begin{enumerate}[]
        \item For $\cB_{N,d}$ with $N \geq 3$, if $\delta \leq c_\delta\epsilon/(1 + \logstar N)$, then $n = \Om{d(1 + \logstar N)/(\alpha\epsilon)}$.
        \item For $\cI_{N,k}$ with $N \geq 2k$, if $\delta \leq c_\delta\epsilon/(1 + \logstar(N/k))$, then $n = \Om{k(1 + \logstar(N/k))/(\alpha\epsilon)}$.
        \item For $\cH_{N,d}$ with $N \geq 3$, if $\delta \leq c_\delta\epsilon/(1 + \logstar N)$, then $n = \Om{d(1 + \logstar N)/(\alpha\epsilon)}$.
        \item For $\cC_{N,k}$ with $\floor{ \sqrt{N} } \geq 2k$, if $\delta \leq c_\delta\epsilon/(1 + \logstar(\sqrt{N}/k))$,\newline then $n = \Om{k(1 + \logstar(\sqrt{N}/k))/(\alpha\epsilon)}$.
    \end{enumerate}
\end{theorem}

\begin{proof}
By \Cref{lem:threshold_packings}, each class $\cH$ in the four items satisfies $(\Thr{L})^k \sqsubseteq \cH$ for an integer $L \geq 2$, where $L = N - 1$ and $k = d$ in (1) and (3), $L = \lfloor N/k \rfloor$ in (2), and $L = \lfloor \lfloor \sqrt{N} \rfloor / k \rfloor$ in (4). We first show that, for such a class and $\delta \leq c_\delta\epsilon/(1 + \logstar L)$,
\begin{equation}
\label{eq:threshold_packing_lb}
n = \Om{\frac{k(1 + \logstar L)}{\alpha\epsilon}},
\end{equation}
by applying \Cref{thm:lower_bound_icp} with the collection $\cG = \{\Thr{L}\}$.

The conditions of \Cref{thm:lower_bound_icp} on $\epsilon$, $\delta$, $\alpha$ and $\beta$ are part of our assumptions. The class $\cH$ is non-trivial (\Cref{def:non_trivial}): let $\Phi_1, \cdots, \Phi_k$ be the maps witnessing $(\Thr{L})^k \sqsubseteq \cH$; as $L \geq 2$, the hypotheses of $\cH$ that extend $(h_1, h_0, \cdots, h_0)$ and $(h_2, h_0, \cdots, h_0)$ agree at $\Phi_1(1)$, where both are $1$, and differ at $\Phi_1(2)$. Let $s = \mpriv{\Thr{L}}{0.1}{\delta/\epsilon}{\alpha_0}{1/4}$. As $\delta/\epsilon \leq \alpha_0/64 \leq 1/64$ and $\delta/\epsilon \leq c_\delta/(1 + \logstar L)$, \Cref{cor:threshold_sc} applies with $\delta/\epsilon$ in place of $\delta$, and gives $s \geq \Om{1 + \logstar L}$. As $(\Thr{L})^k$ is a packable tuple with entries in $\cG$, \Cref{def:icp} gives $\icp_{0.1,\delta/\epsilon}(\cH, \cG) \geq ks$. Hence \Cref{thm:lower_bound_icp} gives
\[
n = \Om{\frac{\icp_{0.1,\delta/\epsilon}(\cH, \cG)}{\alpha\epsilon}} = \Om{\frac{ks}{\alpha\epsilon}} = \Om{\frac{k(1 + \logstar L)}{\alpha\epsilon}},
\]
which is \Cref{eq:threshold_packing_lb}.

Let $M$ be $N$ in (1) and (3), $N/k$ in (2), and $\sqrt N/k$ in (4). Now, in each item $L\leq M<L+2\leq2^L$, using $L\geq2$, and in (4) also $\sqrt{N}<\lfloor\sqrt{N}\rfloor+1$ and $\lfloor\sqrt{N}\rfloor/k<L+1$. As $\logstar$ is nondecreasing and $\logstar(2^L)=1+\logstar L$, the condition $\delta\leq c_\delta\epsilon/(1+\logstar M)$ implies $\delta\leq c_\delta\epsilon/(1+\logstar L)$, and $1+\logstar M\leq2+\logstar L\leq2(1+\logstar L)$. So \Cref{eq:threshold_packing_lb} gives the bound in each item.
\end{proof}

\subsection{\texorpdfstring{Threshold Packings for Small $\delta$}{Threshold Packings for Small delta}}
\label{sec:small_delta}

The bounds of \Cref{cor:four_classes} use the threshold lower bound of \Cref{cor:threshold_sc}, which gives $\Om{1 + \logstar L}$ for a threshold class of length $L$. This dependence on $L$ is very weak, and it does not improve as $\delta$ decreases. For small $\delta$, the lower bound in \Cref{lem:prdim_lower} could be larger. In this subsection we identify the range of $\delta$ in which this happens, and use the packings of \Cref{lem:threshold_packings} to obtain lower bounds multiplicative in $\log(1/\delta)$. As in \Cref{sec:applications}, we fix $\alphazero = 1/192$.

For the packed classes we use the following lower bound, which follows from \cite{BeimelNS19}. The argument in \cite{BeimelNS19} gives $\min\bc{\prdim(\cH), \log(\epsilon/\delta)} \leq 8\alpha\epsilon n + \bigO{1}$, and \Cref{lem:non_trivial_bound} gives $n > 1/(288\alpha\epsilon)$. Together they give the following.

\begin{lemma}[\cite{BeimelNS19}]
    \label{lem:prdim_lower}
    Let $\cH$ be a non-trivial class on a finite domain, $0 < \epsilon \leq 1$, $0 \leq \delta \leq \epsilon$, $0 < \alpha \leq 1/4$ and $0 < \beta \leq 1/3$. Every $(\epsilon, \delta)$-DP, $(\alpha, \beta)$-PAC learner for $\cH$, possibly improper, using $n$ examples satisfies
    \[
    n = \Om{ \frac{\min\bc{\prdim(\cH), \log(\epsilon/\delta)}}{\alpha\epsilon} },
    \]
    where $\prdim(\cH)$ is the representation dimension.
\end{lemma}

\paragraph{The range of $\delta$.}
In \Cref{sec:applications}, each packed threshold class of length $L$ contributes $\Om{1+\logstar L}$. As $\log_2(L+1)\geq\max\bc{1,\logstar L}\geq(1+\logstar L)/2$, the representation-dimension bound for this class is at least of the same order whenever $\log_2(1/\delta)\geq1+\logstar L$, that is,
\(
\delta\leq2^{-(1+\logstar L)}.
\)
In this range the ratio of the displayed lower-bound rates is of order $\min\bc{\log(1/\delta),\log(L+1)}/(1+\logstar L)$, which increases as $\delta$ decreases, until $\delta\leq1/(L+1)$, where the minimum equals $\log(L+1)$ and this factor reaches its pure-private value. The bounds in this subsection hold for all $\delta\leq\min\bc{\alphazero\epsilon/64,\epsilon^2}$, and improve asymptotically on \Cref{cor:four_classes} when the ratio above grows. For threshold classes at privacy parameter $0.1$, this bound reads as follows, by \Cref{lem:prdim_lower} and the bound $2$ of \Cref{sec:prelims}.

\begin{corollary}[Threshold classes for small $\delta$]\label{cor:threshold_small_delta}
Let $L\geq1$, $\alphazero\in(0,1/4]$ and $0<\delta\leq1/100$. Then
\[
\mpriv{\Thr{L}}{0.1}{\delta}{\alphazero}{1/4}
=\Om{\min\bc{\log(1/\delta),\log(L+1)}}.
\]
\end{corollary}
\noindent Using the same packing as \Cref{lem:threshold_packings}, we obtain the following bounds.

\begin{theorem}[Private lower bounds for four classes for small $\delta$]
    \label{lem:four_classes_small_delta}
    Let $0 < \epsilon \leq 0.1$, $0 < \delta \leq \min\bc{\alphazero\epsilon/64, \epsilon^2}$, $0 < \alpha \leq \alphazero/(1536e)$ and $0 < \beta \leq 1/3$, and let $n$ be the number of examples used by an $(\epsilon, \delta)$-DP, $(\alpha, \beta)$-PAC learner, possibly improper, for the class in each item below.
    \begin{enumerate}[]
        \item For $\cB_{N,d}$ with $N \geq 3$, $n = \Om{d\min\bc{\log(1/\delta), \log N}/(\alpha\epsilon)}$.
        \item For $\cI_{N,k}$ with $N \geq 2k$, $n = \Om{k\min\bc{\log(1/\delta), \log(N/k)}/(\alpha\epsilon)}$.
        \item For $\cH_{N,d}$ with $N \geq 3$, $n = \Om{d\min\bc{\log(1/\delta), \log N}/(\alpha\epsilon)}$.
        \item For $\cC_{N,k}$ with $\floor{ \sqrt{N} } \geq 2k$, $n = \Om{k\min\bc{\log(1/\delta), \log(\sqrt{N}/k)}/(\alpha\epsilon)}$.
    \end{enumerate}
\end{theorem}

\begin{proof}
By \Cref{lem:threshold_packings}, each class $\cH$ in the four items satisfies $(\Thr{L})^k \sqsubseteq \cH$ with $L \geq 2$ and $k$ as in the proof of \Cref{cor:four_classes}. Let $s = \mpriv{\Thr{L}}{0.1}{\delta/\epsilon}{\alphazero}{1/4}$. As $\delta/\epsilon \leq \alphazero/64 < 1/100$, \Cref{cor:threshold_small_delta} applies with $\delta/\epsilon$ in place of $\delta$,  giving us
\[
s = \Om{\min\bc{\log(\epsilon/\delta), \log(L+1)}}.
\]
As $s \geq 2$, exactly as in the proof of \Cref{cor:four_classes}, \Cref{thm:lower_bound_icp} with the collection $\cG = \bc{\Thr{L}}$ gives $n \geq \Om{ks/(\alpha\epsilon)}$. As $\delta \leq \epsilon^2$, we have $\epsilon/\delta \geq 1/\epsilon$, so $\log(1/\delta) = \log(1/\epsilon) + \log(\epsilon/\delta) \leq 2\log(\epsilon/\delta)$. Hence
\[
n = \Om{\frac{k\min\bc{\log(1/\delta), \log(L+1)}}{\alpha\epsilon}}.
\]
 In (1) and (3), $L + 1 = N$,  in (2), $L + 1 > N/k$, and in (4), $\sqrt{N}/k < L + 2 \leq 2(L + 1)$, so $\log(\sqrt{N}/k) \leq 1 + \log(L + 1) \leq 2\log(L + 1)$, which completes the proof.
\end{proof}

%% file: contents/04_threshold_lower_bound.tex
In this section we prove the following result.

\begin{restatable}{thm}{thresholdlowerbound}
\label{thm:threshold-m-rows}
There are universal constants \(c,C>0\) such that for all
\(N\geq1,m\geq2\), every \(\br{0.1,\delta}\)-DP algorithm \(A\) with
\(\delta\leq c/m\) that satisfies
\(\Pr\bs{\err_S(A(S))\leq1/16}\geq15/16\)
for every database \(S\) of size $m$ realizable by \(\Thr{N}\) obeys
\(
  \logstar N\leq m+\logstar m+C.
\)
\end{restatable}

In particular, any algorithm satisfying the theorem's error guarantee
needs \(\Om{\logstar N}\) samples.  The same bound holds for PAC learners, by the first part of \Cref{lem:resample}, which turns a PAC learner on $n$ examples into an algorithm with small empirical error on every realizable database of size $9n$.

\begin{corollary}
    \label{cor:threshold_pac}
    Let $c, C$ be the constants of \Cref{thm:threshold-m-rows}. For every $N \geq 1$, every $(0.1,\delta)$-DP, $(1/16,1/16)$-PAC learner for $\Thr{N}$ using $n$ examples with $\delta \leq c/(9n)$ satisfies $\logstar N \leq 18n + C$.
\end{corollary}

We first compare our argument with the earlier proof and then give the
proof sketches.

\subsubsection[Comparison with the proof of Alon et al. (2022)]
{Comparison with the proof of \citet{AlonEtAl2022}}

  Both proofs use leave-one-out
databases with nearly equal numbers of zero and one labels, and the
quantitative Ramsey theorem.  For \(U=\bc{x_1<\cdots<x_{m+1}}\), let
\(S_i\) be the corresponding labeled database after omitting \(x_i\),
and write \(a_i(U)=\Pr\bs{A(S_i)(x_i)=1}\).  The earlier proof rounds
the values \(a_i(U)\) and applies Ramsey's theorem to the resulting
vectors.  On the set produced by the theorem, the rounded value of
\(a_i(U)\) depends only on \(i\), not on the choice of \(U\).  That
proof then fixes the other points and varies one point \(x\).  For
some level \(r\), the probability that the output obtained with \(x\)
labels another point \(y\) as \(1\) is separated from \(r\) by at
least a constant times \(1/m\): it lies below \(r\) when \(y<x\) and
above \(r\) when \(y>x\).

To make these comparisons reliable enough for binary search,
\citet{AlonEtAl2022} draw \(D=\Theta(m^2\log m)\) independent outputs
for each database and average their predictions.  Applying privacy to
the \(D\) outputs multiplies the additive term \(\delta\) by \(D\), so
their argument requires \(\delta=\bigO{1/D}=\bigO{1/(m^2\log m)}\).

Our proof instead records only the first index \(\ell(U)\in\bs m\) for
which \(a_{\ell(U)+1}(U)-a_{\ell(U)}(U)\geq\kappa/m\), where
\(\kappa>0\) is a universal constant.  Ramsey's theorem gives a large
set on which \(\ell(U)\) has the same value \(\ell\).  We fix the
other points and vary the point at position \(\ell\).  For any two
choices \(x<y\), the algorithm run on the database containing \(x\) is
more likely to label \(y\) as \(1\) than the algorithm run on the
database containing \(y\) is to label \(x\) as \(1\).  The difference
is at least \(\kappa/m\).  We prove below, using this pairwise gap and
privacy, that at most \(\exp\br{\bigO m}\) points can occupy position
\(\ell\) while the other points remain fixed, provided \(\delta\leq
c/m\).  Because this argument does not combine \(D\) outputs, it
avoids the \(D\delta\) cost of composition.

\subsubsection{Proof of the threshold bound}\label{sec:threshold-proof-strategy}

The proof follows the outline above. Its new ingredient is the next lemma, which uses the pairwise gap to bound how many points can occupy position $\ell$.

\begin{restatable}{lem}{pairwisepredictiongap}
\label[lemma]{lem:skew-matrix}
Let \(\Omega\) be finite, and let \(P_1,\ldots,P_s\) be distributions on
\(\Omega\) satisfying \(P_i(E)\leq e^\epsilon P_1(E)+\delta\) for every
\(i\in\bs s\) and \(E\subseteq\Omega\).  Let
\(v_1,\ldots,v_s:\Omega\to\bs{0,1}\).  If, for some \(\Gamma>\delta\),
\(\bE_{P_i}\bs{v_j}-\bE_{P_j}\bs{v_i}\geq\Gamma\) for every \(i<j\), then
\[
  s\leq\exp\br{1+\frac{\pi e^{\epsilon/2}}{\Gamma-\delta}}.
\]
\end{restatable}

\begin{proof}[Proof sketch]
Write \(p_i\) for the probability mass function of \(P_i\), let
\(p=p_1\), and replace \(p_i\) by \(\widetilde
p_i=\min\bc{p_i,e^\epsilon p}\).  The comparison with \(P_1\) shows that
this removes at most \(\delta\) mass, so each pairwise gap decreases by at
most \(\delta\).

Since \(\Omega\) is finite, write \(\widetilde p_i=u_i p\) and view
\(u_i\) and \(v_i\) as vectors in \(L_2(P_1)\).  Thus
\(\sum_\omega\widetilde p_i(\omega)v_j(\omega)
=\ip{u_i}{v_j}\), and similarly
\(\sum_\omega\widetilde p_j(\omega)v_i(\omega)
=\ip{u_j}{v_i}\).  We also have
\(\norm{u_i}_2\leq e^{\epsilon/2}\) and \(\norm{v_i}_2\leq1\).  Hence
\(\ip{u_i}{v_j}-\ip{u_j}{v_i}\geq\Gamma-\delta\) for
every \(i<j\).

Let
\[
  T=\sum_{i<j}
  \frac{\ip{u_i}{v_j}-\ip{u_j}{v_i}}{j-i}.
\]
For each \(d=j-i\), there are \(s-d\) pairs at distance \(d\), and hence
\(T\geq(\Gamma-\delta)\sum_{d=1}^{s-1}(s-d)/d\).  The discrete Hilbert
inequality \citep{MontgomeryVaughan1974} and the norm bounds above give
\[
  T\leq\pi
  \br{\sum_i\norm{u_i}_2^2}^{1/2}
  \br{\sum_j\norm{v_j}_2^2}^{1/2}
  \leq\pi e^{\epsilon/2}s.
\]
Finally, \(\sum_{d=1}^{s-1}(s-d)/d\geq s(\ln s-1)\).  Comparing the two
bounds on \(T\) and rearranging gives
\[
  s\leq\exp\br{1+\frac{\pi e^{\epsilon/2}}{\Gamma-\delta}}.
\]
\end{proof}

\noindent We now sketch the proof of \Cref{thm:threshold-m-rows}. The full proofs of \Cref{lem:skew-matrix} and \Cref{thm:threshold-m-rows} are in \Cref{app:threshold-lower-bound}.

\begin{proof}[Proof Sketch for~\Cref{thm:threshold-m-rows}]
Without loss of generality, we work with thresholds that label smaller
indices zero and larger indices one. Randomly permuting inputs let us treat databases as unordered labeled sets. If \(N<m+1\), the result is
immediate, so we assume \(N\geq m+1\).  Set \(r_0=\floor{m/2}\) and
\(r_1=\ceil{m/2}\).  For \(U=\bc{x_1<\cdots<x_{m+1}}\), let \(S_i\) be
the threshold dataset obtained by omitting \(x_i\), with \(r_0\) zero
labels followed by \(r_1\) one labels.  Recall the definition
\(a_i(U)=\Pr\bs{A(S_i)(x_i)=1}\).

\noindent\textbf{Step 1} Accuracy on \(S_{r_0+1}\) gives a
zero-labeled point \(x_{i_-}\) and a one-labeled point \(x_{i_+}\)
that are predicted correctly w.h.p.  Applying privacy
to the corresponding leave-one-out datasets \(S_{i_-}\) and
\(S_{i_+}\) gives a universal constant \(\kappa>0\) such that
\(a_{i_+}(U)-a_{i_-}(U)\geq\kappa\). Then, as the sum below has at most \(m\) terms,
\begin{equation*}
  \frac{\kappa}{m}\leq \frac1m \br{a_{i_+}(U)-a_{i_-}(U)}
  =\frac1{m} \sum_{k=i_-}^{i_+-1}\br{a_{k+1}(U)-a_k(U)}
  \leq \max_{k\in\bs m}\br{a_{k+1}(U)-a_k(U)}.
\end{equation*}
We let \(\ell(U)\in\bs m\) be the smallest index such that
\begin{equation}
  a_{\ell(U)+1}(U)-a_{\ell(U)}(U)\geq\frac{\kappa}{m}.
  \label{eq:threshold-adjacent-gap-sketch}
\end{equation}

\noindent\textbf{Step 2} Color each \((m+1)\)-subset \(U\subseteq\bs N\)
by \(\ell(U)\).  The quantitative Ramsey theorem gives a set
\(H\subseteq\bs N\) on which the color is constant, say \(\ell\), and
\[
  \abs H\geq\frac{\iterlog{m}{N}}{C_0m\log(m+1)}.
\]
If this lower bound is smaller than \(m+1\), the result follows.
We therefore assume \(\abs H\geq m+1\).

\noindent\textbf{Step 3} Let \(B\) consist of the \(\ell-1\) smallest and
the \(m-\ell\) largest points of \(H\), and set \(J=H\setminus B\).  Every
\(y\in J\) has \(\ell-1\) points of \(B\) below it and \(m-\ell\) above it,
so it occupies position \(\ell\) in \(B\cup\bc{y}\).  Let \(S_y\) be the
dataset on this set with \(r_0\) zero labels followed by \(r_1\) one labels.

Replace each output by its values on \(\bs N\).  This makes the output space
finite without changing any prediction probabilities.  Write
\(J=\bc{y_1<\cdots<y_s}\), let \(P_i\) be the output distribution of
\(A(S_{y_i})\), and set \(v_i(h)=\one\bc{h(y_i)=1}\).

\begin{figure}[t]
\centering
\scalebox{0.9}{%
\begin{tikzpicture}[
    font=\small,
    >={Stealth[length=6pt,width=5pt]},
    dot/.style={circle,fill=black!65,minimum size=2.2pt,inner sep=0pt},
    pt/.style={circle,draw,thick,minimum size=5.5pt,inner sep=0pt},
  ]
  \node at (4.5,6.6) {$U=B\cup\bc{y_i,y_j}$};

  \draw[black!55] (0,5.1) -- (9,5.1);
  \foreach \x in {0.6,1.1,1.6} \node[dot] at (\x,5.1) {};
  \foreach \x in {7.4,7.9,8.4} \node[dot] at (\x,5.1) {};
  \node[pt,fill=blue!12,draw=blue!65!black] (yi) at (3.6,5.1) {};
  \node[pt,fill=orange!16,draw=orange!80!black] (yj) at (5.4,5.1) {};
  \node[below=3pt of yi,text=blue!65!black] {$y_i$};
  \node[below=3pt of yj,text=orange!80!black] {$y_j$};

  \draw[decorate,decoration={brace,amplitude=4pt,raise=2pt}]
    (0.3,5.4) -- (1.9,5.4)
    node[midway,above=6pt,font=\footnotesize] {$\ell-1$ fixed points};
  \draw[decorate,decoration={brace,amplitude=4pt,raise=2pt}]
    (7.1,5.4) -- (8.7,5.4)
    node[midway,above=6pt,font=\footnotesize] {$m-\ell$ fixed points};

  \draw[->,blue!65!black] (3.45,4.75) -- (2.3,3.95)
    node[midway,above left,font=\footnotesize] {delete $y_j$};
  \draw[->,orange!80!black] (5.55,4.75) -- (6.7,3.95)
    node[midway,above right,font=\footnotesize] {delete $y_i$};

  \node[text=blue!65!black] (Si) at (2.3,3.55)
    {$S_{y_i}$ on $B\cup\bc{y_i}$};
  \node[text=orange!80!black] (Sj) at (6.7,3.55)
    {$S_{y_j}$ on $B\cup\bc{y_j}$};

  \node[text=blue!65!black,below=9pt of Si] (Pi)
    {$P_i=\mathsf{Law}(A(S_{y_i}))$};
  \node[text=orange!80!black,below=9pt of Sj] (Pj)
    {$P_j=\mathsf{Law}(A(S_{y_j}))$};

  \node at (4.5,0.6) (gap)
    {$\bE_{P_i}\bs{v_j}-\bE_{P_j}\bs{v_i}\geq\kappa/m$};
  \draw[->,blue!65!black] (Pi.south) to[out=-70,in=170] (gap.west);
  \draw[->,orange!80!black] (Pj.south) to[out=-110,in=10] (gap.east);
\end{tikzpicture}
}
\caption{Deleting \(y_j\) gives \(S_{y_i}\), while deleting \(y_i\) gives
\(S_{y_j}\).}
\label{fig:threshold-pair-gap}
\end{figure}

For \(i<j\), the points \(y_i,y_j\) occupy positions \(\ell,\ell+1\) in
\(U=B\cup\bc{y_i,y_j}\).  Deleting \(y_j\) gives \(S_{y_i}\), while
deleting \(y_i\) gives \(S_{y_j}\), as shown in
\Cref{fig:threshold-pair-gap}.  Thus
\Cref{eq:threshold-adjacent-gap-sketch} gives
\begin{equation}
  \bE_{P_i}\bs{v_j}-\bE_{P_j}\bs{v_i}\geq\frac{\kappa}{m}
  \qquad(i<j).
  \label{eq:threshold-pair-gap-sketch}
\end{equation}

\noindent\textbf{Step 4} Any two datasets \(S_y\) agree on the \(m-1\)
labeled samples in \(B\) and differ only in their remaining sample, so they
are neighbors.  Hence privacy gives
\(P_i(E)\leq e^{0.1}P_1(E)+\delta\) for every \(i\in\bs s\) and every event
\(E\).  Together with \Cref{eq:threshold-pair-gap-sketch}, this verifies
the assumptions of \Cref{lem:skew-matrix} with \(\Gamma=\kappa/m\) and
\(\delta\leq\kappa/(2m)\).  The lemma gives
\(\abs J\leq\exp\br{\bigO m}\).
Since \(\abs H=\abs J+m-1\), combining this with the bound from Step~2 gives
\[
  \iterlog{m}{N}\leq\exp\br{\bigO m}.
\]
Applying another \(\logstar m+\bigO{1}\) logarithms yields
\(\logstar N\leq m+\logstar m+\bigO{1}\).
\end{proof}

 For packing thresholds, \Cref{thm:lower_bound_icp} needs the sample complexity of threshold classes at privacy parameters $(0.1,\delta/\epsilon)$, accuracy $\alphazero$ and failure probability $1/4$. The next corollary bounds it, with $\delta/\epsilon$ in place of $\delta$.

\begin{corollary}
    \label{cor:threshold_sc}
    There is a universal constant $c_\delta > 0$ such that the following holds. Let $L \geq 1$, $\alphazero \in (0, 1/192]$ and $0 \leq \delta \leq \min\bc{1/64, c_\delta/(1 + \logstar L)}$. Then
    \[
    \mpriv{\Thr{L}}{0.1}{\delta}{\alphazero}{1/4} = \Om{1 + \logstar L}.
    \]
\end{corollary}

\begin{proof}
Let $s = \mpriv{\Thr{L}}{0.1}{\delta}{\alphazero}{1/4}$. By the second part of \Cref{lem:resample} with $\eta = 1/16$, a learner achieving $s$ gives a $(0.1, \delta)$-DP, $(1/16, 1/16)$-PAC learner for $\Thr{L}$ on $201s$ examples, as $12\alphazero \leq 1/16$. If $\delta \leq c/(1890s)$, \Cref{cor:threshold_pac} applies, as $1890s \geq 9 \cdot 201s$, and gives $\logstar L \leq 18 \cdot 201s + C$, which together with $s \geq 2$ (\Cref{sec:prelims}) gives $s = \Om{1 + \logstar L}$; otherwise $s > c/(1890\delta) \geq 1 + \logstar L$ for $c_\delta = c/1890$.
\end{proof}

%% file: contents/tree_packing.tex
\section{Packing beyond thresholds}
\label{sec:beyond_thresholds}

The bounds in \Cref{lem:four_classes_small_delta} are obtained by packing threshold classes. The packed classes need not be thresholds, and for small $\delta$ other packed classes can give stronger bounds. We give two examples. The first packs several copies of lines into a class of linear subspaces. The second packs $d$ copies of the class of root-to-leaf paths of a binary tree. In both, the packed class has VC dimension at most $2$ but a large representation dimension, so the lower bound of \Cref{lem:prdim_lower} for it adds up over the packing. As in \Cref{sec:applications}, we fix $\alphazero = 1/192$.

\paragraph{Lines.} For a prime $p$, let $\mathrm{Line}_p$ be the class of the functions $f_{a,b}(x,y) = \one\bc{y = ax + b}$ on $\bZ_p^2$, for $(a,b) \in \bZ_p^2$. Feldman and Xiao \cite{FeldmanXiao2014} introduced it to separate pure from approximate private learning: its VC and Littlestone dimensions are $2$, while $\prdim(\mathrm{Line}_p) = \Theta(\log p)$, and they give an approximate private learner with sample complexity $\bigO{\log(1/\delta)\log(1/\beta)/(\alpha\epsilon)}$. This bound does not depend on $p$. Threshold packings give little for this class. A line through two points is unique, so no threshold on three points packs into $\mathrm{Line}_p$, and by \Cref{lem:packing_constraints} a threshold packing has at most two components. The exponential mechanism learns each component with $\bigO{1}$ examples, so \Cref{thm:lower_bound_icp} gives only $\Om{1/(\alpha\epsilon)}$. On the other hand, \Cref{lem:prdim_lower} gives
\[
n = \Om{\frac{\min\bc{\log p, \log(\epsilon/\delta)}}{\alpha\epsilon}}
\]
for every $(\epsilon, \delta)$-DP, $(\alpha, \beta)$-PAC learner with $\beta \leq 1/3$, $0 < \epsilon \leq 0.1, 0 < \alpha \leq 1/4$ and $0 \leq \delta \leq \min\bc{\epsilon/100, \epsilon^2}$. Together with the exponential mechanism for $\delta \leq 1/p$ and the learner of \cite{FeldmanXiao2014} for $\delta \geq 1/p$, this shows that the sample complexity of $\mathrm{Line}_p$ is $\Theta(\min\bc{\log p, \log(1/\delta)}/(\alpha\epsilon))$ for constant $\beta$.

\paragraph{Linear Subspaces.}

We obtain a lower bound for learning subspaces by packing lines.
Let $p$ be prime and $1\le d\le D$. On the domain
$\cX=\bF_p^D$, define
\[
\cS_{d,D}
=
\bc{h_V:V\subseteq\bF_p^D,\ \dim V\leq d},
\quad
h_V(x)=\one\bc{x\in V}.
\]
Here we use $V\subseteq\bF_p^D$ to denote that $V$ is a linear subspace in $\bF_p^D$.

To embed a line into a subspace, we add a coordinate.
The line $y=ax+b$ need not contain the origin, so it need not be a
linear subspace. However,
$\bc{(x,y,z)\in\bF_p^3:y=ax+bz}$ is a linear subspace of dimension 2. On the fixed image
of the map $(x,y)\mapsto(x,y,1)$, its defining equation becomes
$y=ax+b$ and preserves every label of the line.

\begin{theorem}
\label{lem:subspace-packing}
Let $p$ be prime, $3\le d\le D$, and $k=\floor{ d/3}$. Then
\(
(\mathrm{Line}_p)^k\sqsubseteq\cS_{d,D}.
\)
Additionally let $\alphazero=1/192$,
$0<\epsilon\le0.1,
0\leq\delta\le\frac{\alphazero\epsilon}{64},
0<\alpha\le\frac{\alphazero}{1536e},
0<\beta\le\frac13$.
Every $(\epsilon,\delta)$-DP, $(\alpha,\beta)$-PAC learner
for $\cS_{d,D}$, possibly improper, uses
\[
n=\Om{
  \frac{d\min\bc{\log p,\log(\epsilon/\delta)}}
       {\alpha\epsilon}
}
\quad\text{examples}.
\]
\end{theorem}

\begin{proof}
For each $i\in[k]$, take a copy of $\mathrm{Line}_p$ with domain
$\cY_i=\bF_p^2$.
Partition the first $3k$ coordinates into $k$ blocks of size three.
This is possible because $3k\le d\le D$. Define
$\Phi_i:\cY_i\to\cX$ by
\[
\Phi_i(x,y)
=
\br{
\underbrace{0,\ldots,0}_{3(i-1)},
x,y,1,
\underbrace{0,\ldots,0}_{D-3i}
},
\]
The maps are injective with pairwise disjoint images.
Fix arbitrary hypotheses $g_{a_i,b_i}\in\mathrm{Line}_p$,
for $i\in[k]$, and set
\[
V_i=\bc{(x,y,z)\in\bF_p^3:y=a_i x+b_i z},
\qquad
V=V_1\oplus\cdots\oplus V_k\oplus\bc{0}^{D-3k}.
\]
Each $V_i$ occupies block $i$, and remaining
coordinates are zero. As $\dim V_i=2$,
$\dim V=2k\le d$ and $h_V\in\cS_{d,D}$.
For any $(x,y)\in\cY_i$, the $i$-th block of
$\Phi_i(x,y)$ is $(x,y,1)$, while every other block is
$(0,0,0)\in V_j$. Therefore
\[
\Phi_i(x,y)\in V
\quad\Longleftrightarrow\quad
(x,y,1)\in V_i
\quad\Longleftrightarrow\quad
y=a_i x+b_i.
\]
Hence $h_V\circ\Phi_i=g_{a_i,b_i}$ for every $i$.
As the line hypotheses were chosen independently, these maps
witness the packing.
For any collection $\cG$ containing $\mathrm{Line}_p$,
the definition of ICP and the lower bound for lines give
\[
\begin{aligned}
\icp_{0.1,\delta/\epsilon}
(\cS_{d,D},\cG)
&\ge
k\,\mpriv{\mathrm{Line}_p}{0.1}{\delta/\epsilon}{\alphazero}{1/4}
=
\Om{
d\min\bc{\log p,\log(\epsilon/\delta)}
},
\end{aligned}
\]
where $k=\Om{d}$. The sample complexity lower bound follows from
\Cref{thm:lower_bound_icp}.
\end{proof}

\begin{remark}
A recent work by \citet{moran2026} gives a proper learner for $\cS_{d,D}$ using $\bigO{\frac{d\log\!\frac{1}{\alpha\epsilon\delta}+\log\!\frac1\beta}{\alpha\epsilon}}$
examples. Together with \Cref{lem:subspace-packing}, under it's assumptions, if
$
p\ge\epsilon/\delta,
\delta\le\min\bc{\alpha,\epsilon^2}$ and $\beta$ is a fixed constant,
the sample complexity is $\Theta\!\br{\frac{d\log(1/\delta)}{\alpha\epsilon}}$.
\end{remark}

\paragraph{Unions of paths.} For $k \geq 1$, let $X_k = \bigcup_{j=0}^k \bc{0,1}^j$ be the set of nodes of the complete binary tree of depth $k$. The root is the empty string, the children of $u$ are $u0$ and $u1$, and the leaves are the strings of length $k$. We write $u \preceq s$ if $u$ is a prefix of $s$. The class of paths is
\[
\mathrm{Path}_k = \bc{h_s : s \in \bc{0,1}^k}, \qquad h_s(u) = \one\bc{u \preceq s},
\]
on $X_k$. For $1 \leq d \leq 2^k$, the class $\cT_{d,k}$ of unions of at most $d$ paths is the class on $X_{2k}$ of functions
\[
h_A(u) = \one\bc{u \preceq s \text{ for some } s \in A}, \; A \subseteq \bc{0,1}^{2k}, \; \abs{A} \leq d.
\]
This particular class was also studied in \cite{LiLS01}. We describe the packings for this class.

\begin{lemma}[Paths]
    \label{lem:path_classes}
    Let $k \geq 1$ and $1 \leq d \leq 2^k$.
    \begin{enumerate}[]
        \item $\VC(\mathrm{Path}_k) = 1$, $\ldim(\mathrm{Path}_k) = k$ and $\prdim(\mathrm{Path}_k) = \Theta(k)$.
        \item $\VC(\cT_{d,k}) = d$, $\ldim(\cT_{d,k}) = \Theta(dk)$ and $\prdim(\cT_{d,k}) = \Theta(dk)$.
        \item $(\mathrm{Path}_k)^d \sqsubseteq \cT_{d,k}$, $\Thr{k} \sqsubseteq \mathrm{Path}_k$ and $(\Thr{k})^d \sqsubseteq \cT_{d,k}$.
    \end{enumerate}
\end{lemma}

\begin{proof}
We first prove (3). Choose distinct nodes $c_1, \cdots, c_d \in \bc{0,1}^k$ at depth $k$, and let $\Phi_i(u) = c_iu$ be the concatenation of $c_i$ and $u$, for $u \in X_k$. These maps are injective, and their images are the subtrees below the $c_i$, which are pairwise disjoint. Given $s_1, \cdots, s_d \in \bc{0,1}^k$, let $A = \bc{c_1s_1, \cdots, c_ds_d}$. The node $c_iu$ is a prefix of $c_js_j$ only if $j = i$, and it is a prefix of $c_is_i$ if and only if $u \preceq s_i$. So $h_A(\Phi_i(u)) = h_{s_i}(u)$ for all $i$ and $u$, which gives $(\mathrm{Path}_k)^d \sqsubseteq \cT_{d,k}$. For $\Thr{k} \sqsubseteq \mathrm{Path}_k$, map $j \in [k]$ to the node $0^j$ of $j$ zeros. For $t < k$, the leaf $0^t10^{k-t-1}$ gives the labels $\one\bc{j \leq t}$ on these nodes, and for $t = k$ the leaf $0^k$ does. Composing the two packings gives $(\Thr{k})^d \sqsubseteq \cT_{d,k}$.

For (1), the node $0$ is shattered, by paths through $0$ and through $1$. For two nodes, if one is a prefix of the other, no path labels the shorter one $0$ and the longer one $1$, and otherwise no path labels both $1$. So $\VC(\mathrm{Path}_k) = 1$. Asking for the label of $v0$, starting with the empty string $v$ and continuing with $v0$ if the answer is $1$ and with $v1$ otherwise, gives $k$ questions whose answers are realized by every possible sequence, so $\ldim(\mathrm{Path}_k) \geq k$. As $\abs{\mathrm{Path}_k} = 2^k$, also $\ldim(\mathrm{Path}_k) \leq k$ and $\prdim(\mathrm{Path}_k) \leq k$, by taking the whole class as the representation. Finally, $\prdim(\mathrm{Path}_k) = \Om{k}$, as the probabilistic representation dimension is at least a universal constant times the Littlestone dimension \cite{FeldmanXiao2014}.

For (2), any $d$ leaves are shattered, by taking $A$ to be the leaves labeled $1$. A shattered set contains no two nodes of which one is a prefix of the other, so a path contains at most one of its nodes, and labeling all of them $1$ needs as many paths as nodes. So $\VC(\cT_{d,k}) = d$. By (3) and \Cref{lem:packing_constraints}, $\ldim(\cT_{d,k}) \geq dk$. As $\abs{\cT_{d,k}} \leq (2^{2k} + 1)^d \leq 2^{d(2k+1)}$, both the Littlestone and the probabilistic representation dimension are at most $d(2k+1)$, and the latter is $\Om{dk}$ by \cite{FeldmanXiao2014}.
\end{proof}

We first discuss the known bounds for this class.  The exponential mechanism over its at most $2^{d(2k+1)}$ hypotheses is an $(\epsilon, 0)$-DP learner using $\bigO{(dk + \log(1/\beta))/(\alpha\epsilon)}$ examples. For the lower bound, \Cref{lem:prdim_lower} applied to the whole class $\cT_{d,k}$ gives $\Om{\min\bc{dk, \log(\epsilon/\delta)}/(\alpha\epsilon)}$, and by \Cref{thm:lower_bound_icp} with packing of $d$ $\Thr{1}$ classes we get $\Om{d/(\alpha\epsilon)}$. Unless $\log(\epsilon/\delta)$ is larger than $d$, these bounds are only $\Om{d/(\alpha\epsilon)}$.
Let $\ThrCollection = \bc{\Thr{N} : N \geq 1}$. The next lemma shows that threshold packings of $\cT_{d,k}$ give at most $\bigO{d\log(k+1)}$.

\begin{lemma}
    \label{lem:tree_threshold_packing}
    Let $k \geq 1$ and $1 \leq d \leq 2^k$. For every $0 \leq \delta < 1$,
    \(
    \icp_{0.1,\delta}(\cT_{d,k}, \ThrCollection) \leq \bigO{d\log(k+1)},
    \)
    and for $0 \leq \delta \leq \min\bc{\alphazero/64, 1/(k+1)}$, $\icp_{0.1,\delta}(\cT_{d,k}, \ThrCollection) = \Theta(d\log(k+1))$.
\end{lemma}

\begin{proof}
Let $(\Thr{N_1}, \cdots, \Thr{N_t}) \sqsubseteq \cT_{d,k}$. By \Cref{lem:packing_constraints}, $t \leq \VC(\cT_{d,k}) = d$. Choosing the threshold that labels every point $1$ in each component, the packing gives one hypothesis that labels all $\sum_i N_i$ points of the images $1$. A union of at most $d$ paths labels at most $d(2k+1)$ nodes $1$, so $\sum_i N_i \leq d(2k+1)$. The exponential mechanism over the $N + 1$ thresholds on $[N]$ is a $(0.1, 0)$-DP, and hence $(0.1, \delta)$-DP, learner for $\Thr{N}$ using $\bigO{\log(N+1)}$ examples at accuracy $\alphazero$ and failure probability $1/4$. As $t \mapsto t\log(1 + S/t)$ is nondecreasing for $S \geq 0$, concavity of the logarithm gives
\[
\begin{aligned}
\sum_{i=1}^t \mpriv{\Thr{N_i}}{0.1}{\delta}{\alphazero}{1/4}
&\leq \bigO{\sum_{i=1}^t \log(N_i + 1)} \\
&\leq \bigO{d\log\br{1 + \frac{\sum_i N_i}{d}}}
\leq \bigO{d\log(k+1)}.
\end{aligned}
\]
This holds for every threshold packing, which gives the upper bound. For the lower bound, \Cref{lem:path_classes} gives $(\Thr{k})^d \sqsubseteq \cT_{d,k}$. For $0 < \delta \leq \min\bc{\alphazero/64, 1/(k+1)}$, \Cref{cor:threshold_small_delta} gives each $\Thr{k}$ sample complexity $\Om{\min\bc{\log(1/\delta), \log(k+1)}} = \Om{\log(k+1)}$, and for $\delta = 0$ the same bound holds as pure privacy needs at least as many examples.
\end{proof}

The upper bound of \Cref{lem:tree_threshold_packing} bounds the sum of the sample complexities of the packed threshold classes, not the sample complexity of $\cT_{d,k}$. As it uses an upper bound on the sample complexity of each packed class, no better lower bound for threshold classes can make a threshold packing give more than $\bigO{d\log(k+1)/(\alpha\epsilon)}$.
Packing paths instead gives better bounds when $\delta$ is small.

\begin{theorem}
    \label{cor:tree_path_packing}
    Let $k \geq 1$, $d \leq 2^k$, $0 < \epsilon \leq 0.1$, $0 \leq \delta \leq \alphazero\epsilon/64$, $0 < \alpha \leq \alphazero/(1536e)$ and $0 < \beta \leq 1/3$. Every $(\epsilon, \delta)$-DP, $(\alpha, \beta)$-PAC learner for $\cT_{d,k}$, possibly improper, using $n$ examples satisfies
    \[
    n = \Om{\frac{d\min\bc{k, \log(\epsilon/\delta)}}{\alpha\epsilon}}.
    \]
\end{theorem}

\begin{proof}
We apply \Cref{thm:lower_bound_icp} with the collection $\cG = \bc{\mathrm{Path}_k}$ and the packing $(\mathrm{Path}_k)^d \sqsubseteq \cT_{d,k}$ of \Cref{lem:path_classes}. The class $\cT_{d,k}$ is non-trivial: two different paths agree at the root and differ at a leaf. Let $s = \mpriv{\mathrm{Path}_k}{0.1}{\delta/\epsilon}{\alphazero}{1/4}$.  \Cref{lem:prdim_lower}, applied to $\mathrm{Path}_k$ at the privacy parameters $0.1$ and $\delta/\epsilon$, accuracy $\alphazero$ and failure probability $1/4$, together with $\prdim(\mathrm{Path}_k) = \Theta(k)$ and $s \geq 2$, gives\looseness-1
\[
s \geq \Om{\min\bc{k, \log(0.1\epsilon/\delta)}} = \Om{\min\bc{k, \log(\epsilon/\delta)}},
\]
where the last step uses $\epsilon/\delta \geq 64/\alphazero \geq 100$, and for $\delta = 0$ the minimum is $k$. As $\icp_{0.1,\delta/\epsilon}(\cT_{d,k}, \cG) \geq ds$, using \Cref{thm:lower_bound_icp} we complete the proof.
\end{proof}

Compared with \Cref{lem:prdim_lower} applied to the whole class, which gives $\Om{\min\bc{dk, \log(\epsilon/\delta)}/(\alpha\epsilon)}$, the path packing replaces the minimum of $dk$ and $\log(\epsilon/\delta)$ by $d$ times the minimum of $k$ and $\log(\epsilon/\delta)$. It is larger by a factor of $d$ when $\log(\epsilon/\delta) \leq k$, and by a factor of $dk/\log(\epsilon/\delta)$ when $k \leq \log(\epsilon/\delta) \leq dk$. For $d \geq 2$, the two bounds coincide when $\delta \leq \epsilon e^{-dk}$, where both give $\Om{dk/(\alpha\epsilon)}$.

%% file: contents/upper_bounds.tex
\section{Upper Bounds via Covering}
\label{sec:upper_bounds}

In the previous sections we used packing for lower bounds by placing independent learning problems inside a hypothesis class. In this section we obtain upper bounds when $\cH$ is built from the packed classes, which we capture by coverings. For general product classes, and more generally when labels for $\cH$ can be used directly by the learners for the classes $(\cG_1,\ldots,\cG_k)$, we obtain a tight $\icp$ characterization up to logarithmic factors (\Cref{sec:general_product}). Using the learner of
\cite{AlonEtAl2020}, we then show that when $\cH$ is covered by
a boolean aggregation of classes that also pack into $\cH$,
we obtain an upper bound in terms of $\icp$ with an additional factors depending on $\VC(\cH)$ (\Cref{sec:boolean_aggregation}).
We first define covering as a representation of hypotheses.

\begin{definition}[Covering]\label{def:cover}
For integer $k$ let $\cG_i\subseteq\bc{0,1}^{\cY_i}$ for $i\in[k]$ and $\cH\subseteq\bc{0,1}^{\cX}$. A covering of $\cH$ by $(\cG_1,\cdots,\cG_k)$ consists of a boolean function $\cF:\bc{0,1}^k\to\bc{0,1}$ and, for each $i$, a subset $\cX_i\subseteq\cX$, a map $\Psi_i:\cX_i\to\cY_i$, and a function $o_i:\cX\setminus\cX_i\to\bc{0,1}$, with the following properties.
Let
\begin{equation*}
 \widetilde g_i(x)=
 \begin{cases}
  g_i(\Psi_i(x)),&x\in\cX_i,\\
  o_i(x),&x\notin\cX_i,
 \end{cases}
 \qquad
 \widetilde\cG_i=\bc{\widetilde g_i:g_i\in\cG_i}.
\end{equation*}
Then for all \(h\in \cH\), there exists \(\exists(g_1,\ldots,g_k)\in\prod_{i=1}^k\cG_i\) such that \(h(x)=\cF\br{\widetilde g_1(x),\ldots,\widetilde g_k(x)}\) for all \(x\in\cX\).
Equivalently,
$\cH\subseteq\cF(\widetilde\cG_1,\ldots,\widetilde\cG_k)$.
\end{definition}

Each $\widetilde g_i$ is a hypothesis on $\cX$ that reads $g_i$ through $\Psi_i$ on $\cX_i$ and equals the fixed function $o_i$ outside $\cX_i$. In \Cref{lem:transfer} we show that a private learner for $\cG_i$ gives a private learner for $\widetilde\cG_i$. The sets $\cX_i$ may overlap and need not cover $\cX$. The function $\cF$, the sets $\cX_i$, the maps $\Psi_i$ and the functions $o_i$ are fixed for all $h\in\cH$, and only the $g_i$ depend on $h$. Their existence is the structural assumption behind our upper bounds.

\subsection{General product classes}\label{sec:general_product}

A packing maps the $\cY_i$ into $\cX$ by injective maps $\Phi_i$ with pairwise disjoint images, so that every combination of hypotheses, one from each class, is realized by a single hypothesis in $\cH$, while a covering maps parts of $\cX$ back to the $\cY_i$ by the maps $\Psi_i$. When the images of the $\Phi_i$ partition $\cX$, the same tuple both packs and covers the following classes.

\begin{definition}[General product class]\label{prop:product}
For classes $\cG_i \subseteq \bc{0,1}^{\cY_i}$, $i \in [k]$, and injective maps $\Phi_i : \cY_i \to \cX$ whose images partition $\cX$, the product class is
\[
 \cH = \bc{h : \cX \to \bc{0,1} : h \circ \Phi_i \in \cG_i \text{ for every } i}.
\]
\end{definition}
As the images partition $\cX$, every $(g_1, \cdots, g_k) \in \prod_i \cG_i$ defines one $h \in \cH$ with $h \circ \Phi_i = g_i$, so $(\cG_1, \cdots, \cG_k) \sqsubseteq \cH$ via $\Phi_1, \cdots, \Phi_k$. Moreover, with $\cX_i = \Phi_i(\cY_i)$ and $\Psi_i = \Phi_i^{-1}$, every $h \in \cH$ agrees on $\cX_i$ with $g_i \circ \Psi_i$ for $g_i = h \circ \Phi_i \in \cG_i$, that is, $(\cG_1, \cdots, \cG_k)$ covers $\cH$ (\Cref{def:cover}) with $\cF = \mathrm{OR}$ and $o_i = 0$.

For the remainder of this section, fix $0<\alphazero\leq1/4$,
$0<\epsilon\leq0.1$, $0\leq\delta\leq\alphazero\epsilon/64$, and
$0<\alpha,\beta\leq1/3$.
We consider tuples in which every $\cG_i$ contains at least two
hypotheses, and let
$s_i=\mpriv{\cG_i}{0.1}{\delta/\epsilon}{\alphazero}{1/4}$.
Unless stated otherwise, constants may depend on the fixed $\alphazero$. We first reduce the error of a private learner (\Cref{lem:accuracy}), then turn a learner for $\cG_i$ into a learner for $\widetilde\cG_i$ (\Cref{lem:transfer} and \Cref{cor:transfer-parameters}), and finally run these learners on one sample (\Cref{thm:direct}).
To combine the learners for $\cG_i$, we need their errors to sum to
at most $\alpha$. Confidence amplification in \Cref{lem:resample}
increases the error by a constant factor. The following lemma first reduces
the error enough to absorb that factor. \Cref{lem:accuracy} is similar to Theorem~A.1 of \cite{SivakumarBG21} but our proof also applies to improper learners.

\begin{lemma}\label{lem:accuracy}
Let $\cG\subseteq\bc{0,1}^{\cY}$, where $\cY$ is nonempty, and let
$0<\epsilon\leq1$ and $0\leq\delta<1$. Suppose that $\cG$ has an
$(\epsilon,\delta)$-DP, $(1/4,1/4)$-PAC learner using $m\geq1$
examples. Then for every $0<\eta,\gamma\leq1/3$, it has an
$(\epsilon,\delta)$-DP, $(\eta,\gamma)$-PAC learner using at most
\[
 C\frac{m}{\eta}\log^2\frac{2}{\eta}\log\frac{2}{\gamma}
\]
examples, where $C$ is universal. The statement also holds for
improper learners.
\end{lemma}

Next we compare learning $\cG_i$ on $\cY_i$ with learning
$\widetilde\cG_i$ on $\cX$. Covering represents $\cH$ using the
classes $\widetilde\cG_i$ on $\cX$, whereas packing measures the
private learning costs $s_i$ of the classes $\cG_i$ on $\cY_i$.
To relate the upper bound to $\icp$, we therefore need a learner for
$\widetilde\cG_i$ whose cost is bounded by $s_i$.
As the labels outside $\cX_i$ are fixed by $o_i$, we learn only from the examples inside $\cX_i$. If this set has little probability, even arbitrary predictions give small error, and otherwise, a sufficiently large sample supplies enough examples there with high probability. This is captured in~\Cref{lem:transfer}.

\begin{lemma}\label{lem:transfer}
Fix $i\in[k]$. Let $\cX_i\subseteq\cX$,
$\Psi_i:\cX_i\to\cY_i$, and
$o_i:\cX\setminus\cX_i\to\bc{0,1}$ be fixed, and let
$\widetilde\cG_i$ be as in \Cref{def:cover}. Then
$\VC(\widetilde\cG_i)\leq\VC(\cG_i)$ and
$\mpriv{\widetilde\cG_i}{0.1}{\delta/\epsilon}{\alphazero}{1/4}\leq Cs_i$.
\end{lemma}

\noindent~\Cref{lem:transfer} gives us a learner for $\widetilde\cG_i$ with accuracy $\alphazero$ using $\bigO{s_i}$ examples. We then apply accuracy and confidence amplification using~\Cref{lem:accuracy} to this learner and adjust its privacy parameters to get the following DP-PAC guarantee for \(\widetilde\cG_i\).

\begin{corollary}\label{cor:transfer-parameters}
For $\widetilde\cG_i$ as in \Cref{lem:transfer} and
$0<\eta,\gamma\leq1/3$, there is an
$(\epsilon,\delta)$-DP, $(\eta,\gamma)$-PAC learner whose output
equals $o_i$ outside $\cX_i$ with sample complexity
\(
 \bigO{\frac{s_i}{\epsilon\eta}
       \log^2\frac{2}{\eta}\log\frac{2}{\gamma}}.
\)
\end{corollary}

Now, we are ready to give a learner for a general product class (\Cref{prop:product}).
For these classes, a label from $\cH$ is a correct label
for $\cG_i$ on its part of the domain. We can use these labels
directly, and set the labels outside that part to zero before learning.
The same reasoning works whenever every point belongs to a set on
which the observed labels agree with a hypothesis from the corresponding
class, even if these sets initially overlap.

\begin{theorem}\label{thm:direct}
Let $k\geq1$ be an integer, let $\cG_i\subseteq\bc{0,1}^{\cY_i}$
for $i\in[k]$, and let $\cH\subseteq\bc{0,1}^{\cX}$.
Suppose fixed sets $\cX_i\subseteq\cX$ cover $\cX$, and fix maps
$\Psi_i:\cX_i\to\cY_i$. The sets may overlap.
Assume that for every $h\in\cH$, we can choose $g_i\in\cG_i$ for
each $i\in[k]$ such that $h(x)=g_i(\Psi_i(x))$ for every $x\in\cX_i$.
Then, for the parameters and $s_i$ fixed above,
\[
 \mpriv{\cH}{\epsilon}{\delta}{\alpha}{\beta}
 \leq C\frac{\sum_{i=1}^k s_i}{\alpha\epsilon}
             \log^2\frac{4k}{\alpha}\log\frac{2k}{\beta},
\]
where $C$ depends only on $\alphazero$.
In particular, this holds for the general product class in \Cref{prop:product}.
\end{theorem}

\begin{proof}[Proof of~\Cref{thm:direct}]
Use the product learner of \Cref{lem:product-upper}, replacing its
threshold learners by the learners from
\Cref{cor:transfer-parameters} with the following changes.
If the sets overlap, assign each point to the first set containing it.
That is, replace them simultaneously by
$\cX_i\setminus\bigcup_{j<i}\cX_j$ and restrict each $\Psi_i$.
We continue to call these sets $\cX_i$. They partition $\cX$, and
the label assumption still holds on each set.

For each $i$, define $\widetilde\cG_i$ using this set and map, with
$o_i=0$. If $\cX\setminus\cX_i$ is nonempty, fix one point in this
complement. In the sample given to learner $i$, replace every example
outside $\cX_i$ by this fixed point with label zero. Leave the
examples inside $\cX_i$ unchanged. If $\cX_i=\cX$, no replacement
is needed. With this, no new domain point is required.

Run these learners with independent randomness, privacy
$(\epsilon/2,\delta/2)$, failure probability $\beta/k$, and error
\begin{equation}\label{eq:allocation}
 \eta_i=\frac{\alpha}{2}
       \br{\frac{s_i}{\sum_{j=1}^k s_j}+\frac1k}.
\end{equation}
Each output is zero outside its assigned set. On input $x\in\cX_i$,
return the prediction of learner $i$.

The privacy argument from \Cref{lem:product-upper} now applies.
Suppose $(x,\ell)$ is replaced by $(x',\ell')$, where
$x\in\cX_i$ and $x'\in\cX_j$. Every other learner replaces
both records by its same fixed example with label zero.
Thus at most two learners are affected.
Since each is $(\epsilon/2,\delta/2)$-DP, composition
and postprocessing give $(\epsilon,\delta)$-privacy.

For utility, each transformed sample consists of independent examples
and is realizable by $\widetilde\cG_i$, using the assumed label
identity inside $\cX_i$ and label zero outside.
Since learner $i$ predicts zero outside, its error on the transformed
distribution equals the probability of a mistake inside $\cX_i$
under the original distribution. With probability at least $1-\beta$,
all these contributions are at most $\eta_i$. They sum to at most
$\alpha$.
Finally, by the choice in \Cref{eq:allocation} we have \(
 \sum_i\eta_i=\alpha,0<\eta_i\leq\alpha,\eta_i\geq\frac{\alpha}{2k},\frac{s_i}{\eta_i}\leq\frac{2\sum_j s_j}{\alpha}.
\)

We now apply \Cref{cor:transfer-parameters} with privacy
$(\epsilon/2,\delta/2)$, accuracy $\eta_i$, and failure
probability $\beta/k$. The cost remains $s_i$ because
the ratio $\delta/\epsilon$ remains unchanged. Learner $i$
therefore needs at most $C\frac{s_i}{\epsilon\eta_i} \log^2\frac{2}{\eta_i}\log\frac{2k}{\beta}$
examples. Using
$s_i/\eta_i\leq2\sum_j s_j/\alpha$ and
$\eta_i\geq\alpha/(2k)$, this is at most
$C\frac{\sum_j s_j}{\alpha\epsilon} \log^2\frac{4k}{\alpha}\log\frac{2k}{\beta}$.
Take one sample of this size and form a transformed copy for
each learner. A learner requiring fewer examples uses only
the first required number of examples from its copy.
This gives enough examples to all learners and proves the
stated sample bound.

\end{proof}

Note that packing is not needed for \Cref{thm:direct} but we need it to relate the sum in its
upper bound to $\icp$. If the same tuple also packs into $\cH$, we obtain the following.

\thmtightproduct*

\begin{proof}
The lower bound follows from \Cref{thm:lower_bound_icp}. As the
tuple is packable, the definition of $\icp$ gives
\begin{equation}\label{eq:sum-icp}
 \sum_{i=1}^k s_i\leq
 \icp_{0.1,\delta/\epsilon}(\cH,\cG).
\end{equation}
Substitute this inequality into the upper bound in \Cref{thm:direct}.
For the product in \Cref{prop:product}, use
$\cX_i=\Phi_i(\cY_i)$ and $\Psi_i=\Phi_i^{-1}$.
\end{proof}

We do not require the tuple to attain the supremum in $\icp$. Its sum
upper bounds the learning cost up to the displayed factors, and is at
most $\icp$ by packing. The lower bound applies to the entire
supremum, giving the two inequalities in \Cref{cor:product_icp}.

\subsection{Boolean aggregation}\label{sec:boolean_aggregation}

For a general covering, the observed label $h(x)$ need not equal
$\widetilde g_i(x)$ and we cannot use
the product learner directly. Instead, we use the closure algorithm of \cite{AlonEtAl2020}, which  learns a Boolean aggregate one class at a time. In each round, it selects labels for
the next class so that the hypotheses learned so far can still be
completed by the remaining classes to an aggregate with small error.
It then learns a hypothesis from these labels.
In our setting, covering represents every $h\in\cH$ as
$\cF(\widetilde g_1,\cdots,\widetilde g_k)$, so we apply this
algorithm to $\widetilde\cG_1,\cdots,\widetilde\cG_k$.
By \Cref{lem:transfer}, their private learning costs are controlled
by the original costs $s_i$. When the same tuple also packs into
$\cH$, we have $\sum_i s_i\leq\icp_{0.1,\delta/\epsilon}(\cH,\cG)$,
which gives the following bound in terms of $\icp$.

\begin{theorem}\label{thm:boolean_aggr}
Let $\cH$ be a non-trivial binary class and $\cG$ a collection of
hypothesis classes. Suppose the tuple $(\cG_1,\ldots,\cG_k)$, with
$\cG_i\in\cG$, both covers $\cH$ as in \Cref{def:cover} and is
independently packable into $\cH$ (\Cref{def:packing_classes}).
Let $0<\epsilon\leq0.1$,
$0<\beta\leq1/3$, and $\alpha\leq\alpha_0/(1536e)$, for sufficiently small $\delta$
\[
 \Om{\frac{\icp_{0.1,\delta/\epsilon}(\cH,\cG)}{\alpha\epsilon}}
 \leq\mpriv{\cH}{\epsilon}{\delta}{\alpha}{\beta}
 \leq\softO{\frac{\VC(\cH)^3\icp_{0.1,\delta/\epsilon}(\cH,\cG)}
                 {\alpha\epsilon}},
\]
where $\softOsymbol$ hides logarithmic factors in $k,1/\alpha,1/\beta$.
\end{theorem}

\begin{proof}
We apply the closure theorem (Theorem 29) of \cite{AlonEtAl2020} to the classes $\widetilde\cG_1,\ldots,\widetilde\cG_k$ and the function $\cF$. It allows improper learners for these classes, and its sample size depends on their sample sizes and on the VC dimension of $\cF(\widetilde\cG_1,\ldots,\widetilde\cG_k)$. We first construct these learners and bound this VC dimension.

To obtain the learners, start with the
$(0.1,\delta/\epsilon)$-DP, $(\alpha_0,1/4)$-PAC learner
for $\widetilde\cG_i$ from \Cref{lem:transfer}, using
$\bigO{s_i}$ examples. Apply \Cref{lem:accuracy} to reduce its error
and failure probability to sufficiently small constant multiples
of $1/k$, using $\bigO{ks_i\log^3(2k)}$ examples. By \Cref{lem:packing_constraints},
\begin{equation}\label{eq:packing-vc}
 k\leq\sum_{i=1}^k\VC(\cG_i)\leq\VC(\cH).
\end{equation}
Covering and \Cref{lem:transfer} also give
\begin{equation}\label{eq:closure-vc}
 \VC(\cH)\leq\VC\bigl(\cF(\widetilde\cG_1,\ldots,\widetilde\cG_k)\bigr)\leq C\log(2k)\sum_i\VC(\cG_i)
              \leq C\log(2k)\sum_i s_i.
\end{equation}
The first inequality holds as $\cH\subseteq\cF(\widetilde\cG_1,\ldots,\widetilde\cG_k)$. For the second inequality, put $d=\sum_i\VC(\cG_i)$.
On any $n\geq d$ points, the number of labelings by an aggregate
is at most the product of the numbers of labelings by its classes.
Sauer's lemma bounds this product by $(enk/d)^d$.
Letting $d_i=\VC(\cG_i)\geq1$, the product is at most
$\prod_i(en/d_i)^{d_i}$, and
$\prod_i(d/d_i)^{d_i}\leq k^d$ by weighted arithmetic--geometric
mean. If these points are shattered, $2^n\leq(enk/d)^d$, which
implies $n\leq Cd\log(2k)$. The case $n<d$ satisfies the same
bound. The last inequality follows from
the usual constant-accuracy PAC lower bound
$\VC(\cG_i)\leq Cs_i$, which also holds for improper learners.

We now apply Theorem 29 of \cite{AlonEtAl2020} with sufficiently small constant accuracy and failure parameters.
The PAC learners constructed above are also
$(1,\delta/\epsilon)$-DP, and by \Cref{eq:packing-vc,eq:closure-vc}
\(
 \VC\bigl(\cF(\widetilde\cG_1,\ldots,\widetilde\cG_k)\bigr)
 \leq C\log(2k)\VC(\cH).
\)
The theorem then uses $\bigO{k^3\VC(\cF(\widetilde\cG_1,\ldots,\widetilde\cG_k))+k^2\log(2k)}$ examples together with the sample costs of the learners, which is at most $n_0$ for sufficiently large $C_0$.
It gives an
$(\bigO{1},\bigO{\delta/\epsilon})$-DP learner with error at most
$1/4$ and failure probability at most
$\frac18+C(\delta/\epsilon)n_0$.
Covering ensures that every target in $\cH$ belongs to the
aggregate class, so this guarantee applies to learning $\cH$.
The condition $(\delta/\epsilon)n_0\leq c_0$ makes the failure
probability at most $1/4$ for sufficiently small $c_0$.
Thus we obtain a private $(1/4,1/4)$-PAC learner using at most
$n_0$ examples.

We now adjust privacy parameters to $(\epsilon,\delta)$.
From a sample of $\lceil Cn_0/\epsilon\rceil$ examples,
select $n_0$ uniformly without replacement, in uniformly random
order, and run the learner obtained above on them.
Using amplification by subsampling \cite{balle2018privacy} with sufficiently large constant
$C$ gives $(\epsilon,\delta)$-privacy.
The selected examples remain independent samples from the original
distribution, so the learner retains its $(1/4,1/4)$-PAC guarantee.
Finally, apply \Cref{lem:accuracy} to this learner.
Since $\epsilon\leq0.1$, the lemma applies and preserves
$(\epsilon,\delta)$-privacy while reducing the error and failure
probability to $\alpha$ and $\beta$. The resulting sample size is
\[
 \bigO{
 \frac{n_0}{\alpha\epsilon}
 \log^2\frac{2}{\alpha}\log\frac{2}{\beta}
 }.
\]

\Cref{eq:closure-vc,eq:sum-icp} give the upper bound
in terms of $k^3\icp$, and \Cref{eq:packing-vc} gives the upper
bound in terms of $\VC(\cH)^3\icp$.
The lower bound follows from \Cref{thm:lower_bound_icp}.
\end{proof}

%% file: appendix/icp_product_union.tex
This appendix proves \Cref{thm:two-families}. \Cref{app:dimensions} computes the VC and Littlestone dimensions of the two classes, \Cref{app:private_upper} gives their private learners, and \Cref{app:product_union_proof} proves the lower bounds by threshold packing and completes the proof. \Cref{app:icp_comparison} proves properties of packings and $\icp$ that we use throughout the paper. We begin with the dimensions of a threshold class.

\begin{claim}
\label[claim]{lem:threshold-dimensions}
For every \(N\geq1\),
\[
  \VC(\Thr{N})=1
  \text{ and }
  \ldim(\Thr{N})=\floor{\log_2(N+1)}.
\]
\end{claim}

\begin{proof}
It is easy to see that a threshold can shatter one point. However, it cannot shatter two points: it cannot label a smaller point zero
and a larger point one. Thus \(\VC(\Thr{N})=1\).

There are \(N+1\) possible threshold positions. A complete mistake
tree of depth \(d\) requires at least \(2^d\) distinct hypotheses, so
\(\ldim(\Thr{N})\leq\floor{\log_2(N+1)}\). For the reverse inequality,
use induction on the remaining depth \(s\). An interval containing at
least \(2^s\) threshold positions supports a complete tree of depth
\(s\): query a point that leaves at least \(2^{s-1}\) positions after
either label, and apply the induction hypothesis to both branches.
Starting with all \(N+1\) positions gives a complete tree of depth
\(\floor{\log_2(N+1)}\).\looseness-1
\end{proof}

\subsection{Product Class and Union Class}\label{app:dimensions}

The product and union have the same VC and Littlestone dimensions for different reasons. Intuitively, in~\(\ProdClass_{v,D}\), the contributions of each disjoint \(X_i\) add. In \(\UnionClass_{v,D}\), the subdomain \(A\) gives the VC dimension and \(B\) gives the Littlestone dimension.

\begin{lemma}
\label{lem:product-dimensions}
The class \(\ProdClass_{v,D}\) in \Cref{eq:product-class} satisfies
\[
  \VC(\ProdClass_{v,D})=v
  \text{ and }
  \ldim(\ProdClass_{v,D})=D.
\]
\end{lemma}

\begin{proof}
Choose one point from each \(X_i\). As their labels are chosen
independently by moving the corresponding thresholds, these \(v\)
points can be shattered. Conversely, by pigeonhole principle, any set
of \(v+1\) points contains at least two points in the same \(X_i\),
and a threshold cannot shatter two points. Thus,
\(\VC(\ProdClass_{v,D})=v\).

By \Cref{lem:threshold-dimensions}, the threshold on \(X_i\) has a
complete mistake tree of depth \(r_i\). Concatenating these trees
gives a complete mistake tree of depth \(\sum_i r_i=D\), because the
\(v\) thresholds can be chosen independently. For the upper bound, run
an optimal threshold strategy separately on each \(X_i\). There are at
most \(r_i\) mistakes on \(X_i\), and therefore at most \(\sum_i
r_i=D\) mistakes in total.This proves that the Littlestone dimension is at most \(D\).
\end{proof}

\begin{lemma}
\label{lem:union-dimension}
The class \(\UnionClass_{v,D}\) in \Cref{eq:union-class} satisfies
\[
  \VC(\UnionClass_{v,D})=v\;
  \text{ and } \;
  \ldim(\UnionClass_{v,D})=D.
\]
\end{lemma}

\begin{proof}
The full Boolean class shatters \(A\), so \(\VC(\UnionClass_{v,D})\ge v\).  A set contained in \(A\) has size at most \(v\), and a
threshold cannot shatter two points in \(B\). A set intersecting both
\(A\) and \(B\) is not shattered either, because no hypothesis labels
a point in each set by one. Therefore \(\VC(\UnionClass_{v,D})=v\).

The threshold class on \(B\) has \(2^D\) threshold positions, so its
Littlestone dimension is \(D\) and thus,
\(\ldim\br{\UnionClass_{v,D}}\ge D\). On the other hand, the two
subclasses share the all-zero hypothesis, and hence
\[
  \ldim\br{\UnionClass_{v,D}} \leq \log_2\abs{\UnionClass_{v,D}}
  =\log_2\br{2^v+2^D-1}
  <D+1.
\]
\end{proof}

\subsection{Private Learners for Product and Union Class}\label{app:private_upper}

For the product, we learn one threshold on each \(X_i\) and combine the
outputs. For the union, we learn one hypothesis from \(\cF_v\) and one
threshold on \(B\), then privately select between them using the private selection of \Cref{lem:priv_selection_union}.
Let \(\ell_N=\logstar\br{\max\bc{2,N}}\).

\begin{lemma}[Private threshold learner]
\label[lemma]{lem:private-threshold-upper}
There is a universal constant \(C\) such that \(\Thr{N}\) has a proper,
realizable, \((\epsilon,\delta)\)-DP learner with error \(\eta\), failure
probability \(\zeta\), and sample size at most
\begin{equation}
  C\frac{\ell_N\log^2\br{2\ell_N/(\zeta\delta)}}
    {\epsilon\eta}.
  \label{eq:threshold-upper}
\end{equation}
The same bound, up to a universal factor, holds after adjoining a domain
point that every target labels zero.
\end{lemma}

\begin{proof}
The sample bound follows from \citet[Theorem~3.9]{CohenEtAl2023}; the
normalization by \(\ell_N\) changes its iterated-logarithm factor by at most
two. For the second statement, place the new point last in
the order and run the same learner on the enlarged domain. If it returns the
threshold that labels every point one, move the threshold just before the new
point. This changes no prediction on the original domain and changes the
prediction at the new point from one to zero. The modification is
postprocessing and can only reduce the error. Finally,
\(\ell_{N+1}=\Theta(\ell_N)\), so the sample bound changes by at most a
universal factor.
\end{proof}

\begin{lemma}[Private learner for the product]
\label{lem:product-upper}
For all \(\epsilon,\delta,\alpha,\beta\in(0,1)\),
\[
  \mpriv{\ProdClass_{v,D}}{\epsilon}{\delta}{\alpha}{\beta}
  \leq
  C\frac{vL_{\mathrm P}}
          {\epsilon\alpha}
  \log^2\br{\frac{vL_{\mathrm P}}{\beta\delta}}
\]
for a universal constant \(C\).
\end{lemma}

\begin{proof}
For each \(i\), add a point \(z_i\) outside \(X_{\mathrm P}\) and extend
every threshold on \(X_i\) to label \(z_i\) by zero. Given a sample \(S\),
construct \(S_i\) by keeping each example whose input lies in \(X_i\) and
replacing every other example by \((z_i,0)\). Run the learner from
\Cref{lem:private-threshold-upper} on \(S_i\) with parameters
\(\br{\epsilon/2,\delta/2,\alpha/v,\beta/v}\).
Do this independently for every \(i\), and combine the returned thresholds.

If one sample in \(S\) changes, only the samples for its old and new sets
\(X_i\) can change. Thus at most two threshold learners receive neighboring
inputs. Since each is \((\epsilon/2,\delta/2)\)-DP, basic composition shows
that their combined output is \((\epsilon,\delta)\)-DP.

Let \(P\) be the original distribution. Drawing from \(P\) and applying the
transformation that defines \(S_i\) gives a realizable distribution for the
threshold on \(X_i\), extended by zero at \(z_i\). With probability at least
\(1-\beta\), all \(v\) returned thresholds have error at most \(\alpha/v\)
on their transformed distributions. Since \(z_i\) is always labeled
correctly, this error is precisely the probability of a mistake on \(X_i\)
under \(P\). The \(X_i\)'s are disjoint, so the errors sum to at most
\(\alpha\).
Finally, \(\ell_{N_i}=\bigO{L_{\mathrm P}}\) for every \(i\). Substituting the
parameters above into \Cref{eq:threshold-upper} gives the stated sample bound.
\end{proof}

\begin{lemma}\cite{McSherryT07}
\label{prop:exponential_mechanism}
    Let $\cE(\cdot)$ be an exponential mechanism sampling in some finite space $\cC$ with probabilities proportional to $\exp\br{ \epsilon s(D,c) / (2\mathsf{sens}_s)}$, where $\mathsf{sens}_s$ bounds change in $s(D,c)$ under neighboring datasets. $\cE(\cdot)$ is $(\epsilon,0)$-differentially private. Further, for any input dataset $D$ and $t>0$, let $\textsc{OPT} = \max_{c\in \cC}s(D,c)$, then
    \[
    \Pr\bs{s(D,\cE(D)) \leq \textsc{OPT} - \frac{2\mathsf{sens}_s}{\epsilon} (\log (\abs{\cC}) + t)} \leq \exp(-t).
    \]
    In particular,
    \[
    \bE\bs{s(D,\cE(D))} \geq \textsc{OPT} - \frac{2\mathsf{sens}_s}{\epsilon} (\log (\abs{\cC}) + 1)
    \]
\end{lemma}

The following lemma is used in the learner for the union class, and in (3c) of \Cref{lem:padding}.

\begin{lemma}
\label{lem:priv_selection_union}
Let $0< \epsilon \leq 1$, $0< \alpha \leq 1$, $0< \eta <1$, and let $r \ge 1$ be an
integer. Let $\mu$ be a distribution over a domain $\cX$, let $c: \cX \rightarrow \bc{0,1}$
be a target, and let $h_1,\cdots,h_r : \cX \rightarrow \bc{0,1}$ be hypotheses fixed in
advance with $\min_i\err_{\mu}(h_i) \leq \alpha/8$. Then there is an $(\epsilon,0)$-DP algorithm (the exponential mechanism in the proof, whose privacy holds for every database $S$ and every $h_1, \cdots, h_r$) that on input a database $S$ with $n$ labeled examples outputs $i \in [r]$ such that if $S$ consists of $n \geq 24 \ln(3r/\eta)/(\epsilon\alpha)$ i.i.d records from $\mu$ labeled by $c$, with probability at least $1-\eta$, $\err_{\mu}(h_i) \leq \alpha$.
\end{lemma}

\begin{proof}
On database $S$ with $h_1,\cdots,h_r$, run exponential mechanism with score function $ - n \err_S(h_i)$, which has sensitivity 1. Privacy follows from the guarantees of the exponential mechanism.

Fix $i^\star$ with $\err_{\mu}(h_{i^\star}) \le \alpha/8$, and call $i$ \emph{bad} if $\err_{\mu}(h_i)>\alpha$. The random variable $n\err_S(h_i)$ is a sum of $n$ independent Bernoullis with mean $n\err_{\mu}(h_i)$. Then Chernoff's bound, applied with the upper bound $\alpha n/8$ on the mean, gives us $\Pr\bs{\err_S(h_{i^\star}) \geq \alpha/4} \leq e^{-\alpha n/24}.$ For each bad $i$, using Chernoff's bound with the lower bound $\alpha n$ on the mean, we have $\Pr\bs{\err_S(h_i) \leq \alpha/2} \leq e^{-\alpha n/8}.$
By a union bound over $i^\star$ and at most $r$ bad indices, the event $E$ that
$\err_S(h_{i^\star}) < \alpha/4$ and $\err_S(h_i)> \alpha/2$ for every bad $i$
fails with probability at most $(r+1)e^{-\alpha n/24}$.

By the utility guarantee of exponential mechanism, for every $t > 0$ the output $i$ satisfies $n\err_S(h_i) \geq \min_{i'} n\err_S(h_{i'}) + t$ with probability at most $r e^{-\epsilon t/2}$. Applying this with $t = \alpha n/4$ and using $\min_{i'} \err_S(h_{i'}) \leq \err_S(h_{i^\star})$, the output $i$ satisfies
$\err_S(h_i) < \err_S(h_{i^\star}) + \alpha/4$, except with probability at most
$r\,e^{-\epsilon\alpha n/8}$. When both, this event and $E$ occur,
$\err_S(h_i) < \alpha/2$, so $i$ is not bad, and $\err_{\mu}(h_i) \leq \alpha$.
As $\epsilon \leq 1$ and $n \geq 24 \ln(3r/\eta)/(\epsilon\alpha)$, the failure probability is at most
\[
(r+1)e^{-\alpha n/24}+r \cdot e^{-\epsilon\alpha n/8} \leq 3r e^{-\epsilon\alpha n/24} \leq \eta.
\]
\end{proof}

\begin{lemma}[Private union learner]
    \label{lem:union_learner}
    Let $0 < \epsilon \leq 1, 0 \leq \delta < 1, 0 < \alpha \leq 1$ and $0 < \beta < 1$. For $j \in \bc{1, 2}$, suppose $\cA_j$ is a realizable $(\epsilon, \delta)$-DP learner for $\cH_j$, with error $\alpha/8$, failure probability $\beta/4$, and sample size $m_j$. Then $\cH_1 \cup \cH_2$ has a realizable $(\epsilon, \delta)$-DP learner with error $\alpha$, failure probability $\beta$, and sample size
    \[
    m_1 + m_2 + \ceil{\frac{24\ln(8/\beta)}{\epsilon\alpha}}.
    \]
\end{lemma}

\begin{proof}
Let $n = \ceil{24\ln(8/\beta)/(\epsilon\alpha)}$. The learner partitions its sample into two training blocks of sizes $m_1$ and $m_2$ and a block $V$ of size $n$. It runs $\cA_j$ on training block $j$ with independent randomness, obtaining candidates $h_1$ and $h_2$, and then runs the algorithm of \Cref{lem:priv_selection_union} on $V$ with the candidates $h_1, h_2$.

\emph{Privacy.} Two neighboring inputs differ in one example, which lies in a single block. If it lies in training block $j$, then all other blocks are unchanged, so the output is a randomized post-processing of $\cA_j$ on that block and is $(\epsilon, \delta)$-DP. If it lies in $V$, then the candidates are unaffected, and for every fixed value of them the selection step is $(\epsilon, 0)$-DP by \Cref{lem:priv_selection_union}. Hence the learner is $(\epsilon, \delta)$-DP.

\emph{Utility.} Suppose the target lies in $\cH_j$. With probability at least $1 - \beta/4$, the candidate $h_j$ has population error at most $\alpha/8$. The block $V$ is disjoint from the training blocks, so, conditioned on the two candidates, they are fixed hypotheses and $V$ consists of $n$ i.i.d.\ examples labeled by the target. For $r = 2$ and $\eta = 3\beta/4$ we have $3r/\eta = 8/\beta$, so $n$ satisfies the condition of \Cref{lem:priv_selection_union}. Hence, if $h_j$ has error at most $\alpha/8$, the selected candidate has error at most $\alpha$, except with probability at most $3\beta/4$. By a union bound, the learner fails with probability at most $\beta/4 + 3\beta/4 = \beta$.
\end{proof}

\begin{lemma}[Private learner for the union class]
\label{lem:union-upper}
For all \(\epsilon,\delta,\alpha,\beta\in(0,1)\),
\[
  \mpriv{\UnionClass_{v,D}}{\epsilon}{\delta}{\alpha}{\beta}
  \leq
  C\frac{
    v+L_{\mathrm U}\log^2\br{L_{\mathrm U}/(\beta\delta)}
      +\log(1/\beta)}
    {\epsilon\alpha}.
\]
\end{lemma}

\begin{proof}
On \(\cF_v\), use the exponential mechanism over its \(2^v\) hypotheses,
with score equal to minus the number of mistakes. This is a proper
\((\epsilon,0)\)-DP learner using
\(\bigO{(v+\log(1/\beta))/(\epsilon\alpha)}\) examples.

For the threshold learner, add a point \(z\) outside $A \cup B$ and
extend every threshold on \(B\) to label \(z\) by zero. Keep an example
\((x,y)\) unchanged when \(x\in B\), and replace it by \((z,0)\) when
\(x\in A\). Apply \Cref{lem:private-threshold-upper} to the transformed
sample and extend the returned threshold by zero on \(A\). Since
the target threshold is zero on \(A\), the transformation preserves
realizability, and the error on the transformed distribution equals the
error on \(B\) under the original distribution. Since
\(\ell_{2^D-1}=\Theta(L_{\mathrm U})\), this learner contributes the second
term in the display.

Applying \Cref{lem:union_learner} to these two learners with error $\alpha / 8$ and failure probability $\beta / 4$ and absorbing the $\bigO{\ln(8/\beta) / (\alpha\epsilon)}$ into $v$ and $\log(1/\beta)$ completes the proof.

\end{proof}

\subsection{Proof of the product-union theorem}\label{app:product_union_proof}

The lower bounds follow from \Cref{thm:lower_bound_icp}, applied to packings of threshold classes, together with the lower bound for threshold classes of \Cref{cor:threshold_sc}. We first give the packings.

\begin{lemma}[Threshold packings of the product and union classes]
    \label{lem:product_union_packings}
    Let $1 \leq v \leq D$.
    \begin{enumerate}[]
        \item $(\Thr{N_1}, \cdots, \Thr{N_v}) \sqsubseteq \ProdClass_{v,D}$.
        \item $(\Thr{1})^v \sqsubseteq \UnionClass_{v,D}$ and $(\Thr{2^D - 1}) \sqsubseteq \UnionClass_{v,D}$.
    \end{enumerate}
\end{lemma}

\begin{proof}
For (1), let $\Phi_i(x) = (i, x)$ for $i \in [v]$ and $x \in [N_i]$. These maps are injective with pairwise disjoint images $X_1, \cdots, X_v$. Given threshold positions $t_1, \cdots, t_v$ with $0 \leq t_i \leq N_i$, the hypothesis $h_t \in \ProdClass_{v,D}$ satisfies $h_t(\Phi_i(x)) = \one\bc{x \leq t_i}$ for all $i$ and $x$, which is \Cref{def:packing_classes} for this tuple.

For (2), let $a_1, \cdots, a_v$ be the points of $A$, and let $\Phi_i(1) = a_i$ for $i \in [v]$. Given $t_1, \cdots, t_v \in \bc{0, 1}$, the hypothesis $h \in \cF_v$ that labels $a_i$ by $t_i$ for every $i$ satisfies $h(\Phi_i(1)) = t_i = \one\bc{1 \leq t_i}$. For the second packing, let $\Phi_1$ be the identity map on $B = [2^D - 1]$. Every threshold on $B$, extended by zero on $A$, lies in $\UnionClass_{v,D}$, which gives the packing.
\end{proof}

We now prove \Cref{thm:two-families}.

\begin{proof}
Set $\epszero = 0.1$, $\alphazero = (1/192)/(1536e)$ and $\betazero = 1/3$, and let $c_0 > 0$ be a small universal constant chosen below, so that $\delta_{v,D} = c_0/(v + L_U) \leq c_0$. We apply \Cref{thm:lower_bound_icp} with $1/192$ as the accuracy parameter of the packed classes, which is the parameter called $\alphazero$ in \Cref{thm:lower_bound_icp}, so $\alphazero = (1/192)/(1536e)$ is the largest accuracy that it allows. Its conditions on the privacy and accuracy parameters are $\epszero \leq 0.1$, $\delta_{v,D} \leq \epszero/(192 \cdot 64)$, the condition on $\alphazero$ just mentioned, and $\betazero \leq 1/3$. They hold once $c_0 \leq 1/122880$.

\emph{The threshold classes.} Let $c_\delta$ be the constant of \Cref{cor:threshold_sc}, and let $N$ be the length of one of the threshold classes in \Cref{lem:product_union_packings}, so $N \in \bc{1, N_1, \cdots, N_v, 2^D - 1}$ and $N \leq 2^D$. We check that \Cref{cor:threshold_sc} applies to $\Thr{N}$ with $\delta_{v,D}/\epszero$ in place of $\delta$. As $N \leq 2^D$, we have $1 + \logstar N \leq 2 + \logstar D \leq 3L_U$, and $\delta_{v,D}/\epszero = 10c_0/(v + L_U) \leq 10c_0/L_U$. So $\delta_{v,D}/\epszero \leq \min\bc{1/64, c_\delta/(1 + \logstar N)}$ once $c_0 \leq \min\bc{1/640, c_\delta/30}$, and \Cref{cor:threshold_sc}, together with the bound $2$ of \Cref{sec:prelims}, gives
\begin{equation}
\label{eq:product_union_threshold}
\mpriv{\Thr{N}}{0.1}{\delta_{v,D}/\epszero}{1/192}{1/4} \geq \max\bc{2, \Om{1 + \logstar N}}.
\end{equation}
In particular, every packed class below has sample complexity at least $2$. Both classes (product and union) are non-trivial (\Cref{def:non_trivial}).

\emph{The product class.} By (1) of \Cref{lem:product_union_packings} and \Cref{def:icp}, the collection $\cG = \bc{\Thr{N_1}, \cdots, \Thr{N_v}}$ has $\icp_{0.1,\delta_{v,D}/\epszero}(\ProdClass_{v,D}, \cG)$ at least the sum of the sample complexities of its $v$ entries. So \Cref{thm:lower_bound_icp} and \Cref{eq:product_union_threshold} give
\[
n \geq \Om{\frac{1}{\alphazero\epszero}\sum_{i=1}^v (1 + \logstar N_i)} = \Om{\sum_{i=1}^v (1 + \logstar N_i)},
\]
as $\alphazero$ and $\epszero$ are universal constants. Moreover $1 + \logstar N_i \geq L_P$ for every $i$. If $D/v < 2$, then $L_P = 1$. Otherwise $r_i \geq \floor{D/v} \geq 2$, so $\logstar N_i = 1 + \logstar r_i \geq 1 + \logstar\floor{D/v} \geq \logstar(D/v) = L_P$, where the last inequality uses $\logstar(x+1)\leq\logstar(2^x)=1+\logstar x$ for $x\geq1$. Hence $n \geq \Om{vL_P}$.

\emph{The union class.} With the collection $\cG = \bc{\Thr{1}}$ and the packing $(\Thr{1})^v \sqsubseteq \UnionClass_{v,D}$ of \Cref{lem:product_union_packings}, \Cref{thm:lower_bound_icp} and \Cref{eq:product_union_threshold} give $n \geq \Om{v}$, as each of the $v$ entries has sample complexity at least $2$. With the collection $\cG = \bc{\Thr{2^D - 1}}$ and the packing $(\Thr{2^D - 1}) \sqsubseteq \UnionClass_{v,D}$, they give $n \geq \Om{1 + \logstar(2^D - 1)}$. Moreover $1 + \logstar(2^D - 1) \geq L_U$. This is clear for $D = 1$, as then $L_U = 1$, and for $D \geq 2$ we have $\logstar(2^D - 1) \geq \logstar(2^{D-1}) = 1 + \logstar(D - 1) \geq \logstar D = L_U$. Hence $n \geq \Om{\max\bc{v, L_U}} = \Om{v + L_U}$. Both lower bounds hold for improper learners, as \Cref{thm:lower_bound_icp} does.

\emph{The upper bounds.} The dimension statements follow from \Cref{lem:product-dimensions,lem:union-dimension}, and the upper bounds follow from \Cref{lem:product-upper,lem:union-upper}. For the chosen value of $\delta_{v,D}$, these bounds are $vL_P$ times a polylogarithmic factor in $v + L_P$, and $v + L_U$ times a polylogarithmic factor in $v + L_U$, respectively. Here $L_U \leq L_P + \logstar(\max\bc{2, v}) + \bigO{1} = \bigO{v + L_P}$. This completes the proof.
\end{proof}

%% file: appendix/compare_icp.tex
\section{Additional properties of $\icp$}
\label{app:icp_comparison}

We show two constraints that every packing satisfies in \Cref{def:icp}.

\begin{lemma}[Constraints on packings]
    \label{lem:packing_constraints}
    Let $(\cG_1, \cdots, \cG_k) \sqsubseteq \cH$. Then
    \[
    \sum_{i=1}^k \VC(\cG_i) \leq \VC(\cH) \qquad \text{and} \qquad \sum_{i=1}^k \ldim(\cG_i) \leq \ldim(\cH).
    \]
    Moreover, let $0 \leq \delta' \leq 1/64$ and $\alphazero \in (0, 1/4]$. Then every $\cG_i$ either consists of a single concept, and then $\mpriv{\cG_i}{0.1}{\delta'}{\alphazero}{1/4} = 0$, or it has VC dimension at least $1$ and $\mpriv{\cG_i}{0.1}{\delta'}{\alphazero}{1/4} \geq 2$. In particular, at most $\VC(\cH)$ of the classes $\cG_i$ have nonzero sample complexity.
\end{lemma}

\begin{proof}
Let $\Phi_1, \cdots, \Phi_k$ be the maps witnessing the packing. For the first inequality, let $Z_i \subseteq \cY_i$ be a set shattered by $\cG_i$ of size $\VC(\cG_i)$. For any labeling of $\Phi_1(Z_1) \cup \cdots \cup \Phi_k(Z_k)$, choose for every $i$ a concept $g_i \in \cG_i$ with the corresponding labels on $Z_i$. By \Cref{def:packing_classes}, some $h \in \cH$ satisfies $h \circ \Phi_i = g_i$ for every $i$, so $h$ has the given labeling. As the images are pairwise disjoint, this set has size $\sum_i \VC(\cG_i)$ and is shattered by $\cH$.

For the second inequality, let $D_i = \ldim(\cG_i)$. We build a complete mistake tree for $\cH$ of depth $\sum_i D_i$. Start with a complete mistake tree for $\cG_1$ of depth $D_1$ on $\cY_1$, with every query $y$ replaced by $\Phi_1(y)$. At each of its leaves, continue with a complete mistake tree for $\cG_2$ of depth $D_2$ with queries mapped by $\Phi_2$, and so on. Every root-to-leaf path is consistent with some $g_i \in \cG_i$ on the queries in $\Phi_i(\cY_i)$, for every $i$. By \Cref{def:packing_classes}, some $h \in \cH$ satisfies $h \circ \Phi_i = g_i$ for every $i$, so $h$ is consistent with the whole path. This gives a complete mistake tree of depth $\sum_i D_i$ for $\cH$.

For the last part, if $\cG_i$ consists of a single concept, then the learner that outputs it without examples is $(0.1, 0)$-DP and $(0, 0)$-PAC, so its sample complexity is $0$. Otherwise, two concepts of $\cG_i$ differ at some point $y$, so $\bc{y}$ is shattered and $\VC(\cG_i) \geq 1$. The point masses on $(y, 0)$ and $(y, 1)$ are both realizable by $\cG_i$, and the argument in \Cref{sec:prelims}, which only uses these point masses and $\delta' \leq 1/64$, gives sample complexity at least $2$. Finally, the classes with nonzero sample complexity have VC dimension at least $1$, so by the first inequality there are at most $\VC(\cH)$ of them.
\end{proof}

$\icp$ for a class $\cH$ depends on the collection $\cG$ of classes that may be packed. When $\cG$ consists of classes of VC dimension one, which includes all threshold classes, $\icp$ can be bounded in terms of the $\VC$ and $\ldim$ of $\cH$. Let $\cG_{\leq 1}$ be the collection of all nonempty classes on finite domains with VC dimension $1$, and let $\ThrCollection = \bc{\Thr{N} : N \geq 1} \subseteq \cG_{\leq 1}$.

\begin{lemma}
    \label{lem:icp_vc_ldim}
     Let $\cH$ be a class on a finite domain with $v = \VC(\cH) \geq 1$ and $D = \ldim(\cH)$, let $\alphazero \in (0, 1/192]$, and let $0 < \delta' \leq \min\bc{1/64, c_{\delta}/(1 + \logstar D)}$, where $c_{\delta}$ is the constant of \Cref{cor:threshold_sc}. Then
    \[
    \Om{v + 1 + \logstar D} \leq \icp_{0.1,\delta'}(\cH, \ThrCollection) \leq \icp_{0.1,\delta'}(\cH, \cG_{\leq 1}) \leq \softO{v\br{\logstar(\max\bc{2, D/v})}}
    \]
    where $\softO{.}$ hides polylog factors in $(1 + \logstar D)$ and a factor that depends only on $\delta'$ and $\alphazero$.
\end{lemma}

\begin{proof}

Let $s(\cG) = \mpriv{\cG}{0.1}{\delta'}{\alphazero}{1/4}$, $\tau_0 = 2$ and $\tau_{j+1} = 2^{\tau_j}$. Applying $\log_2$ to $\tau_j$ exactly $j$ times gives $2$, so for $x \geq 1$ we have $\logstar x \geq j + 1$ if and only if $x \geq \tau_j$. Hence $\logstar x$ is the number of indices $j \geq 0$ with $x \geq \tau_j$. Consider integers $D_1, \cdots, D_k \geq 1$ such that $k \leq v$ and $\sum_i D_i \leq D$.
We have, $\sum_{i=1}^k \logstar D_i = \sum_{j \geq 0} \abs{\bc{i \in [k] : D_i \geq \tau_j}}$.
For every $j$, the set $\bc{i \in [k] : D_i \geq \tau_j}$ has at most $k \leq v$ elements, and, as $\sum_i D_i \leq D$, it has at most $D/\tau_j$ elements. Let $\ell$ be the smallest index with $\tau_\ell \geq D/v$.
For each index $j < \ell$, we bound $\abs{\bc{i : D_i \geq \tau_j}}$ by $v$, so these terms contribute at most $v\ell$ to the sum, and $\ell \leq \logstar(\max\bc{2, D/v})$, as $\tau_0, \cdots, \tau_{\ell-1} < D/v$. For the indices $j \geq \ell$ we use the bound $D/\tau_j$; as $\tau_{j+1} \geq 2\tau_j$, these terms add up to at most $2D/\tau_\ell \leq 2v$. Combining we get

\begin{equation}
\label{eq:log_star_grouping}
\sum_{i=1}^k \br{1 + \logstar D_i} \leq 3v\br{1 + \logstar(\max\bc{2, D/v})}.
\end{equation}

We first show that packings compose. If $(\cG_1, \cdots, \cG_k) \sqsubseteq \cH$ via $\Phi_1, \cdots, \Phi_k$, and $\Thr{N_i} \sqsubseteq \cG_i$ via $\psi_i$ for every $i$, then $(\Thr{N_1}, \cdots, \Thr{N_k}) \sqsubseteq \cH$ via $\Phi_1 \circ \psi_1, \cdots, \Phi_k \circ \psi_k$. As the maps are injective with pairwise disjoint images, given thresholds $h_{t_1}, \cdots, h_{t_k}$, the packings in the $\cG_i$ give $g_i \in \cG_i$ with $g_i \circ \psi_i = h_{t_i}$, and the packing in $\cH$ gives $h \in \cH$ with $h \circ \Phi_i = g_i$, so $h \circ \Phi_i \circ \psi_i = h_{t_i}$ for every $i$.

As $\Thr{N}$ has VC dimension $1$, we have $\ThrCollection \subseteq \cG_{\leq 1}$. Let $x_1, \cdots, x_v$ be a set shattered by $\cH$, and let $\Phi_i(1) = x_i$. Every labeling of the $x_i$ is realized by $\cH$, so $(\Thr{1})^v \sqsubseteq \cH$, and as $s(\Thr{1}) \geq 2$, we get $\icp_{0.1,\delta'}(\cH, \ThrCollection) \geq 2v$.

It remains to show $\icp_{0.1,\delta'}(\cH, \ThrCollection) \geq \Om{1 + \logstar D}$. If $D < 2^{16}$, then $\logstar D \leq 4$, and this follows from the bound $2v \geq 2$. Let $D \geq 2^{16}$ and $r = \floor{\log_2 D}$. Shelah's theorem gives points $x_1, \cdots, x_r$ and hypotheses $h_1, \cdots, h_r \in \cH$ with $h_i(x_j) = \one\bc{i \leq j}$. Let $\Phi(s) = x_{r-s}$ for $s \in [r-1]$. For every $t \in \bc{0, \cdots, r-1}$, the hypothesis $h_{r-t}$ satisfies $h_{r-t}(\Phi(s)) = \one\bc{s \leq t}$, so $(\Thr{r-1}) \sqsubseteq \cH$. As $r - 1 \leq D$, we have $\delta' \leq c_\delta/(1 + \logstar(r-1))$, and \Cref{cor:threshold_sc} gives $s(\Thr{r-1}) \geq \Om{1 + \logstar(r-1)}$. To compare $\logstar(r-1)$ with $\logstar D$, let $y = \log_2 D \geq 16$. Then $r - 1 \geq y - 2 \geq \sqrt{y} \geq \log_2 y$, so
\[
\logstar(r-1) \geq \logstar(\log_2 y) = \logstar(\log_2\log_2 D) = \logstar D - 2,
\]
and as $\logstar D \geq 4$, this gives $1 + \logstar(r-1) \geq (1 + \logstar D)/5$.

We now prove the upper bound. Fix a packing $(\cG_1, \cdots, \cG_k) \sqsubseteq \cH$ with entries in $\cG_{\leq 1}$ and $s(\cG_i) \geq 2$ for every $i$. By \Cref{lem:packing_constraints}, $k \leq v$, every $\cG_i$ has VC dimension exactly $1$ and Littlestone dimension $D_i \geq 1$, and $\sum_i D_i \leq D$. The learner of \citet{Yan2025} learns a class of VC dimension $1$ and Littlestone dimension $D_i$ with $\softO{\logstar\br{\max\bc{2,D_i}}}$ examples, so $s(\cG_i) \leq C_{\delta'}(1 + \logstar D_i)\,\polylog{1 + \logstar D_i}$, where $C_{\delta'}$ depends only on $\delta'$ and $\alphazero$. As $D_i \leq D$, \Cref{eq:log_star_grouping} gives
\begin{align*}
\sum_{i=1}^k s(\cG_i) &\leq C_{\delta'}\,\polylog{1 + \logstar D} \sum_{i=1}^k \br{1 + \logstar D_i} \\
&\leq 3C_{\delta'}\, v\br{1 + \logstar(\max\bc{2, D/v})}\polylog{1 + \logstar D}.
\end{align*}

Finally, taking supremum over packings completes the proof.

\end{proof}

%% file: appendix/independent_class_packings_proofs.tex
\section{Proofs for \Cref{sec:icp}}
\label{app:icp-proofs}

We first show that any sufficiently accurate private empirical learner is also an accurate PAC learner.

\begin{restatable}{lem}{lemEmpToPopError}
    \label{lem:emp_to_pop_error}
    Let $q \geq 1$ and let $\cM$ be an $(\epsilon,\delta)$-DP
algorithm on databases of size $q$ whose output is a binary
hypothesis. For every distribution $\mu$ over labeled examples and $S
\sim \mu^q$, we have
    \[
    \bE_{S, \cM}\bs{\err_{\mu}(\cM(S))} \leq e^\epsilon \cdot \bE_{S, \cM}\bs{\err_{S}(\cM(S))} + \delta.
    \]
\end{restatable}

\begin{proof}[Proof of~\Cref{lem:emp_to_pop_error}]
    Let $S = (z_1, \cdots, z_q) \sim \mu^q$, where $z_i = (x_i,b_i)$.
    Fix \(j\in \bc{1,\ldots,q}\) and draw a fresh example $z = (x,b)$
    from $\mu$ independently of $S$. Let $S^j$ be obtained by
    replacing $z_j$ with $z$, and let $F_z = \bc{h : h(x) \neq b}$.
    Conditioning on $S$ and $z$, the databases $S$ and $S^j$ are
    neighbouring. Then,
    \[
    \Pr\bs{\cM(S) \in F_z} \leq e^\epsilon \Pr\bs{\cM(S^j) \in F_z} + \delta.
    \]

    We now average over $S$ and $z$. As $z$ is independent of
    $S$, the left side becomes
    $\bE_{S,\cM}\bs{\err_\mu(\cM(S))}$.
    On the right, $(S^j,z)$ has the same distribution as
    $(S,z_j)$, since $S^j$ consists of $q$ independent examples
    from $\mu$ with $z$ in position $j$. Thus,
    \[
    \bE_{S,\cM}\bs{\err_\mu(\cM(S))}
    \leq e^\epsilon
    \bE_S \Pr_{\cM}\bs{\cM(S)(x_j) \neq b_j}
    + \delta.
    \]
    Averaging over $j \in [q]$ gives the claimed bound.
\end{proof}

We use the following resampling lemma of \citet{BunEtAl2015} to prove \Cref{lem:resample}.

\begin{restatable}[Resampling \citep{BunEtAl2015}]{lem}{lemSecrecyOfSample}
    \label{lem:secrecy_of_sample}
    Let $0 < \epsilon \leq 1$, and let $\cM$ be an $(\epsilon, \delta)$-DP algorithm on databases of size $n$. For $M \geq 2n$, the algorithm that, on a database of size $M$, draws $n$ examples independently and uniformly with replacement from it and runs $\cM$ on them is $(\epsilon', \delta')$-DP, with
    \[
    \epsilon' = \frac{6\epsilon n}{M} \qquad \text{and} \qquad \delta' = e^{6\epsilon n/M}\,\frac{4n\delta}{M}.
    \]
\end{restatable}

\begin{proof}[Proof of~\Cref{lem:resample}]

    Let $S$ be the database of labeled examples of size $M \geq 9n$. Draw $n$ samples (call this $S'$) independently and uniformly with replacement from $S$ and run $\cM$ on these examples. Now, $S'$ is an i.i.d.\ sample from the uniform distribution on $S$, which is realizable, and the population error with respect to this distribution is the empirical error on $S$, which gives us the $(\alpha,\beta)$ empirical error guarantee. For privacy, by \Cref{lem:secrecy_of_sample}, this mechanism satisfies DP with parameters
    \[
    \epsilon' = \frac{6\epsilon n}{M} \leq \frac{2\epsilon}{3} \leq \epsilon \qquad \delta' = e^{6\epsilon n / M} \frac{4n\delta}{M} \leq e^{2\epsilon/3} \frac{4\delta}{9} \leq \delta.
    \]

    For the second part, partition the $Tn$ examples into $T$ groups, run $\cM$ independently on each group, and return their pointwise majority. The mechanism is $(\epsilon, \delta)$-DP by parallel composition and post-processing. As the mechanism is independent across groups with failure probability at most $\beta \leq 1/3$, by Hoeffding's inequality
    \[
    \Pr\bs{\text{at least $5T / 12$ runs fail}} \leq \exp(-2T (5/12 - 1/3)^2) = \exp(-T/72) \leq \eta.
    \]

    Conditioning on the complement of this event, at least $7T / 12$ groups have error at most $\alpha$; let this set of groups be $C$. If on an example the majority is wrong, at least $T/12$ groups among these predict incorrectly, i.e. for an example $(x,y)$
    \[
    \frac{T}{12} \one\bc{\text{Maj}(x) \neq y} \leq \sum_{i \in C} \one\bc{h_i(x) \neq y}.
    \]

    Averaging over a realizable distribution $\mu$,
    \[
    \frac{T}{12} \bE_{\mu}\bs{\one\bc{\text{Maj}(x) \neq y}} \leq \sum_{i \in C} \bE_{\mu}\bs{\one\bc{h_i(x) \neq y}} \Rightarrow \frac{T}{12} \err_\mu(\text{Maj}) \leq \sum_{i \in C} \err_{\mu}(h_i) \leq \abs{C} \alpha \leq T \alpha.
    \]
    Thus, $\err_\mu(\text{Maj}) \leq 12\alpha$ and this completes the proof.
\end{proof}

\begin{proof}[Proof of~\Cref{lem:duplicate}]
    As $\epsilon \leq 0.1$, we have $\ell \geq 1$ and $\ell\epsilon \leq 0.1$. Let $S'$ be an input database of size $m$, and let $S$ be obtained from $S'$ by repeating every example $\ell$ times. If $S'$ is realizable, then $S$ is realizable by the same hypothesis, and as every example is repeated the same number of times, the empirical error of any hypothesis on $S'$ equals its empirical error on $S$. This gives the utility guarantee. Changing one example of $S'$ changes $\ell$ examples of $S$, so by group privacy $\cM'$ is $(\ell\epsilon, \delta\sum_{s=0}^{\ell-1}e^{s\epsilon})$-DP. As $\ell\epsilon \leq 0.1$ and
    \[
    \delta\sum_{s=0}^{\ell-1}e^{s\epsilon} \leq e^{0.1}\ell\delta \leq \frac{0.1e^{0.1}\delta}{\epsilon} \leq \frac{\delta}{\epsilon},
    \]
    $\cM'$ is $(0.1, \delta/\epsilon)$-DP.
\end{proof}

\begin{restatable}{lem}{lemNonTrivialBound}
\label{lem:non_trivial_bound}
Let $\cH$ be non-trivial (\Cref{def:non_trivial}), $\epsilon>0$, $0 \leq \delta \leq \epsilon$,
$0<\alpha\le1/4$ and $0<\beta\le1/3$. Every $(\epsilon,\delta)$-DP,
$(\alpha,\beta)$-PAC learner for $\cH$, possibly improper, using $n$
examples satisfies
\[
n>\frac{1}{288\,\alpha\epsilon}.
\]
\end{restatable}

\begin{proof}[Proof of~\Cref{lem:non_trivial_bound}]

Let $\cM$ be such a learner, and let $h_1,h_2,x,x'$ be the witnesses for non-triviality, so $h_1(x) = h_2(x)$ and $h_1(x') \neq h_2(x')$. For $b\in\bc{1,2}$, let $P_b$ put probability mass $1-2\alpha$ on $(x,h_1(x))$ and $2\alpha$ on $(x',h_b(x'))$. As $h_b(x) = h_1(x)$, the distribution $P_b$ is realized by $h_b$, and $P_1$ and $P_2$ differ only in the label of the rare point $x'$. Let $E=\bc{h:h(x')=h_1(x')}$. A hypothesis outside $E$ misclassifies $x'$ under $P_1$, which has mass $2\alpha > \alpha$, so every hypothesis with error at most $\alpha$ under $P_1$ lies in $E$. Similarly, every hypothesis in $E$ misclassifies $x'$ under $P_2$, so every hypothesis with error at most $\alpha$ under $P_2$ lies outside $E$. As $\cM$ is $(\alpha,\beta)$-PAC with $\beta \leq 1/3$, it outputs an element of $E$ with probability at least $2/3$ on $S\sim P_1^n$ and at most $1/3$ on $S\sim P_2^n$.

We now couple the two samples so that they differ in few rows. Put each index of $[n]$ in a random set $R$ independently with probability $2\alpha$, and let $S^b$ have the example $(x',h_b(x'))$ at the indices in $R$ and $(x,h_1(x))$ elsewhere. Then $S^b\sim P_b^n$, and $S^1$ and $S^2$ differ exactly in the $\abs{R}$ rows indexed by $R$. Let $r=\floor{24\alpha n}$. As $\bE\bs{\abs{R}} = 2\alpha n$, Markov's gives us $\Pr\bs{\abs{R}>r}\le\Pr\bs{\abs{R}\ge24\alpha n}\le1/12$, where the first inequality holds as $\abs{R} > r$ implies $\abs{R} \geq r + 1 > 24\alpha n$. Fix $R$ with $\abs{R} \leq r$. The databases $S^1$ and $S^2$ then differ in at most $r$ rows. By the group privacy guarantee
\[
\Pr\bs{\cM(S^1) \in E \mid R} \leq e^{r\epsilon}\Pr\bs{\cM(S^2) \in E \mid R} + \delta\sum_{s=0}^{r-1}e^{s\epsilon} \leq e^{r\epsilon}\br{\Pr\bs{\cM(S^2) \in E \mid R} + r\delta}.
\]
Averaging over $R$, and using the trivial bound $1$ when $\abs{R} > r$,
\[
\tfrac23 \leq \Pr\bs{\cM(S^1) \in E} \leq \tfrac{1}{12} + e^{r\epsilon}\br{\Pr\bs{\cM(S^2) \in E} + r\delta} \leq \tfrac1{12}+e^{r\epsilon}\br{\tfrac13+r\delta}.
\]
If $r\epsilon\le1/4$ and $r\delta\le1/12$, the right side is at most
$1/12+e^{1/4}\cdot5/12<2/3$, a contradiction. Hence $r\epsilon>1/4$ or
$r\delta>1/12$.  Since $\delta\leq\epsilon$, either case gives
$r\epsilon>1/12$. As $r\leq24\alpha n$, we obtain
$n>1/(288\alpha\epsilon)$.
\end{proof}

%% file: appendix/04_threshold_lower_bound_proofs.tex
This appendix gives the full proofs of \Cref{lem:skew-matrix} and \Cref{thm:threshold-m-rows}.

\pairwisepredictiongap*
\begin{proof}
The claim is immediate for \(s=1\), so assume \(s\geq2\).  Write
\(p_i(\omega)=P_i(\bc{\omega})\), let \(p=p_1\), and replace \(p_i\) by
\(\widetilde p_i(\omega)=\min\bc{p_i(\omega),e^\epsilon p(\omega)}\).
This changes \(p_i\) only on
\(E_i=\bc{\omega:p_i(\omega)>e^\epsilon p(\omega)}\).  Applying the assumed
inequality to \(E_i\) gives
\[
  \sum_{\omega\in\Omega}\br{p_i(\omega)-\widetilde p_i(\omega)}
  =P_i(E_i)-e^\epsilon P_1(E_i)\leq\delta.
\]
Thus the replacement lowers the expectation of any \(\bs{0,1}\)-valued
function by at most \(\delta\).  Set
\(\widetilde M_{ij}=\sum_{\omega\in\Omega}\widetilde p_i(\omega)v_j(\omega)\).
For \(i<j\), replacing \(p_i\) lowers the first expectation by at most
\(\delta\).  Replacing \(p_j\) lowers the expectation that is subtracted,
so it can only increase the difference.  Hence
\(\widetilde M_{ij}-\widetilde M_{ji}\geq\Gamma-\delta\).

Since \(\widetilde p_i\leq e^\epsilon p\), it vanishes wherever \(p\) does.
We may therefore write \(\widetilde p_i=u_i p\), setting
\(u_i(\omega)=0\) when \(p(\omega)=0\), and view \(u_i\) and \(v_i\) in
\(L_2(P_1)\).  Then
\(\widetilde M_{ij}=\ip{u_i}{v_j}\).  Moreover,
\(0\leq u_i\leq e^\epsilon\) and
\(\bE_{P_1}u_i\leq1\), so
\(\norm{u_i}_2\leq e^{\epsilon/2}\) and \(\norm{v_i}_2\leq1\).

Weight the gap for each pair \(i<j\) by \(1/(j-i)\), and set
\[
  T=\sum_{i<j}
  \frac{\ip{u_i}{v_j}-\ip{u_j}{v_i}}{j-i}.
\]
For each \(d=j-i\), there are \(s-d\) pairs at distance \(d\), so
\(T\geq(\Gamma-\delta)\sum_{d=1}^{s-1}(s-d)/d\).
To bound \(T\) from above, let \(K\) be the \(s\times s\) matrix with
\(K_{ii}=0\) and \(K_{ij}=1/(j-i)\) for \(i\ne j\).  Pairing the terms
indexed by \((i,j)\) and \((j,i)\) gives
\(T=\sum_{i,j}K_{ij}\ip{u_i}{v_j}\).  The discrete Hilbert
inequality bounds its \(\ell_2\) operator norm by
\(\norm{K}_{\mathrm{op}}\leq\pi\)
\citep{MontgomeryVaughan1974}, and hence
\[
  \abs T
  \leq \norm{K}_{\mathrm{op}}
  \br{\sum_i\norm{u_i}_2^2}^{1/2}
  \br{\sum_j\norm{v_j}_2^2}^{1/2}
  \leq\pi e^{\epsilon/2}s.
\]
Since
\[
  \sum_{d=1}^{s-1}\frac{s-d}{d}
  =s\sum_{d=1}^{s-1}\frac1d-(s-1)\geq s(\ln s-1),
\]
the preceding bound gives
\((\Gamma-\delta)(\ln s-1)\leq\pi e^{\epsilon/2}\).  Rearranging proves the
claim.
\end{proof}

\input{appendix/04_threshold_theorem_proof}

%% file: appendix/04_threshold_theorem_proof.tex
\thresholdlowerbound*
\begin{proof}
We first reduce to suffix thresholds, which label an initial segment zero
and the remaining points one.  Complement the labels of a suffix-threshold
database, run the prefix-threshold algorithm, and complement its output.
This preserves privacy and leaves the set of misclassified samples
unchanged, so we continue to call the transformed algorithm \(A\).  We may
also randomly permute the samples before running \(A\).  This lets us treat
the databases below as unordered labeled sets and does not change privacy or
accuracy.

If \(N<m+1\), then \(\logstar N\leq\logstar(m+1)\), and the conclusion
follows after increasing \(C\).  Assume \(N\geq m+1\).  Set
\(r_0=\floor{m/2}\) and
\(r_1=\ceil{m/2}\), so \(r_0+r_1=m\).  Fix
\(U=\bc{x_1<\cdots<x_{m+1}}\).  For each \(i\), let \(S_i\) be the database
on \(U\setminus\bc{x_i}\) whose first \(r_0\) points have label zero and whose
last \(r_1\) points have label one.  Define
\(a_i(U)=\Pr\bs{A(S_i)(x_i)=1}\).
The point \(x_i\) is absent from \(S_i\), so this prediction is not counted
in \(\err_{S_i}(A(S_i))\).

Consider \(T=S_{r_0+1}\), and set \(\eta=31/256\).  The accuracy guarantee
and the fact that empirical error is at most one give \(
  \bE\bs{\err_T(A(T))}
  \leq\frac{15}{16}\frac1{16}+\frac1{16}=\eta.
\)
Averaging separately over the zero- and one-labeled blocks gives indices
\(i_-\leq r_0\) and \(i_+\geq r_0+2\) such that
\[
  \begin{aligned}
    \Pr\bs{A(T)(x_{i_-})=1}&\leq\frac{m}{r_0}\eta\quad\text{and}\quad
    \Pr\bs{A(T)(x_{i_+})=0}&\leq\frac{m}{r_1}\eta.
  \end{aligned}
\]
The database \(S_{i_-}\) is obtained from \(T\) by replacing
\((x_{i_-},0)\) with \((x_{r_0+1},0)\).  Similarly, \(S_{i_+}\) replaces
\((x_{i_+},1)\) with \((x_{r_0+1},1)\).  Applying privacy to the events
\(\bc{h:h(x_{i_-})=1}\) and \(\bc{h:h(x_{i_+})=0}\) gives
\[
  \begin{aligned}
    a_{i_-}(U)&\leq e^{0.1}\frac{m}{r_0}\eta+\delta\quad\text{and}\quad
    1-a_{i_+}(U)&\leq e^{0.1}\frac{m}{r_1}\eta+\delta.
  \end{aligned}
\]
Since \(m(1/r_0+1/r_1)\leq9/2\), set
\(\kappa_0=1-(9/2)e^{0.1}\eta>0\).  The two bounds above give
\(a_{i_+}(U)-a_{i_-}(U)\geq\kappa_0-2\delta\).  Set
\(\kappa=\kappa_0/2\) and choose the theorem's constant \(c\leq\kappa/2\).
Since \(\delta\leq c/m\), the difference is at least \(\kappa\), and
\(\delta\leq\kappa/(2m)\).
By telescoping, this difference is the sum of at most \(m\) adjacent
differences.  To disambiguate the choice, let \(\ell(U)\in\bs m\) be the
smallest index
\begin{equation}
  a_{\ell(U)+1}(U)-a_{\ell(U)}(U)\geq\frac{\kappa}{m}.
  \label{eq:threshold-adjacent-gap}
\end{equation}

Color each \((m+1)\)-subset \(U\subseteq\bs N\) by \(\ell(U)\in\bs m\).  The
quantitative Ramsey bound of \citet[Theorem~19]{AlonEtAl2022} gives a set
\(H\subseteq\bs N\) and an index
\(\ell\in\bs m\) such that every \((m+1)\)-subset of \(H\) has color
\(\ell\), and, for a universal constant \(C_0\),
\begin{equation}
  \abs H\geq\frac{\iterlog{m}{N}}{C_0m\log(m+1)}.
  \label{eq:threshold-ramsey}
\end{equation}
Here \(\iterlog{m}{N}\) is the \(m\)-fold base-two logarithm, truncated below
at one.  If the right side of \Cref{eq:threshold-ramsey} is smaller than
\(m+1\), then
\(\iterlog{m}{N}<C_0m(m+1)\log(m+1)\).  Applying another
\(\logstar m+\bigO{1}\) logarithms already proves the theorem.  We may therefore
assume \(\abs H\geq m+1\).

It remains to upper-bound \(\abs H\).  Let \(B_-\) contain the
\(\ell-1\) smallest points of \(H\), let \(B_+\) contain the
\(m-\ell\) largest points, and set
\(B=B_-\cup B_+\) and \(J=H\setminus B\).
Every \(y\in J\) has exactly \(\ell-1\) points of \(B\) below it and
\(m-\ell\) points of \(B\) above it.  Thus \(y\) has position \(\ell\) in
\(B\cup\bc{y}\).  Let \(S_y\) be the database on this set whose first
\(r_0\) points have label zero and whose last \(r_1\) points have label one.
For distinct \(x,y\in J\), the databases \(S_x\) and \(S_y\) share the
\(m-1\) labeled samples on \(B\) and differ only in their remaining sample.
They are therefore neighbors.

Replace each output by its values on \(\bs N\).  This preserves privacy and
all prediction probabilities and makes the output space finite.  Denote this
space by \(\Omega\), and enumerate \(J=\bc{y_1<\cdots<y_s}\).  Let \(P_i\)
be the output distribution of \(A(S_{y_i})\), and define
\(v_i(h)=\one\bc{h(y_i)=1}\).

Fix \(i<j\) and apply \Cref{eq:threshold-adjacent-gap} to
\(U=B\cup\bc{y_i,y_j}\).  The points \(y_i,y_j\) have positions
\(\ell,\ell+1\).  Deleting \(y_j\) gives \(S_{y_i}\), while deleting \(y_i\)
gives \(S_{y_j}\).  Hence
\begin{equation}
  \bE_{P_i}v_j-\bE_{P_j}v_i\geq\frac{\kappa}{m}.
  \label{eq:threshold-pair-gap}
\end{equation}

For every \(i>1\), the databases \(S_{y_i}\) and \(S_{y_1}\) are neighbors,
so privacy gives \(P_i(E)\leq e^{0.1}P_1(E)+\delta\)
for every \(E\subseteq\Omega\).
The same inequality is immediate for \(i=1\).  Together with
\Cref{eq:threshold-pair-gap}, this verifies the assumptions of
\Cref{lem:skew-matrix} with \(\epsilon=0.1\) and \(\Gamma=\kappa/m\).
The lemma gives \(\abs J\leq\exp\br{\bigO m}\).

Finally, \(\abs H=\abs J+m-1\).  Combining this identity with
\Cref{eq:threshold-ramsey} gives
\[
  \iterlog{m}{N}\leq\exp\br{\bigO m}.
\]
Applying \(\log_2\) another \(\logstar m+\bigO{1}\) times reduces the right side
below two.  Since \(\iterlog{m}{N}\) already applies \(m\) logarithms to
\(N\), we obtain
\(\logstar N\leq m+\logstar m+\bigO{1}\), as required.
\end{proof}

%% file: appendix/upper_bound_proofs.tex
\section{Proofs for \Cref{sec:upper_bounds}}
\label{app:upper_bound}

\begin{proof}[Proof of~\Cref{lem:transfer}]
First, note that any learner can always return a hypothesis equal to $o_i$ outside $\cX_i$. Now, suppose $\cX_i=\cX$. Replace each example $(x,\ell)$ by $(\Psi_i(x),\ell)$, run the learner for $\cG_i$ to obtain $\widehat g_i$, and return $x\mapsto\widehat g_i(\Psi_i(x))$. The transformed examples are independent samples from a distribution realizable by $\cG_i$. The returned hypothesis has exactly the learner's error on that distribution. A replacement changes at most one transformed record. Thus the same sample size and learning parameters suffice.

For general $\cX_i$, \Cref{lem:accuracy} gives a
$(0.1,\delta/\epsilon)$-DP learner for $\cG_i$ with error
$\alphazero$, failure probability $1/8$, and sample size $r\leq Cs_i$.
Here $\alphazero\leq1/4$ is fixed.
From $N=\ceil{32r/\alphazero}$ examples, keep the first at most
$r$ whose inputs lie in $\cX_i$, and replace each retained
$(x,\ell)$ by $(\Psi_i(x),\ell)$. Fill any missing positions with
a fixed labeled point of $\cY_i$. Uniformly permute these $r$
examples and run the learner. Return its prediction composed with
$\Psi_i$ on $\cX_i$, and return $o_i$ outside $\cX_i$.
Thus we have an algorithm even when no input lies in $\cX_i$.

Replacing one original record changes the retained and filled multiset
by at most one example. The permutation argument in the proof of
\Cref{lem:accuracy} therefore preserves
$(0.1,\delta/\epsilon)$-privacy.

For utility, fix a distribution $P$ and a target
$\widetilde g_i\in\widetilde\cG_i$. If $P(\cX_i)\leq\alphazero$,
the returned hypothesis has error at most $\alphazero$ regardless of
its predictions inside $\cX_i$. If $P(\cX_i)>\alphazero$, the
expected number retained before truncation exceeds $32r$.
A Chernoff bound makes the probability of fewer than $r$ such
examples at most $e^{-4r}<1/8$.
Conditional on having enough examples, the first $r$ are independent
samples from $P$ conditioned on $\cX_i$. This follows by conditioning
on which sample positions lie in $\cX_i$, as in the acceptance
argument of \Cref{lem:accuracy}. Their images under $\Psi_i$ have
a distribution realizable by $g_i$. Except with probability $1/8$,
the learner's error on this conditional distribution is at most
$\alphazero$. The overall error is this conditional error multiplied by
$P(\cX_i)$, and is therefore at most $\alphazero$. The two failure
probabilities sum to at most $1/4$, and $N\leq Cs_i$.

Finally, a set shattered by $\widetilde\cG_i$ lies inside $\cX_i$,
since labels outside are fixed. Two of its points cannot have the same
image under $\Psi_i$, since they would always receive the same label.
Its image is therefore a set of the same size shattered by $\cG_i$.
This proves the VC bound and completes the proof.
\end{proof}

\begin{proof}[Proof of~\Cref{lem:accuracy}]
Let $\cA$ be the given learner. Its expected error on every realizable
distribution is at most $(3/4)(1/4)+1/4=7/16$.
We first obtain error $\eta/12$ with failure probability at most $1/4$.

Let $T$ be the smallest positive odd integer with
$T\geq128\ln(96/\eta)$, and set $N=\ceil{96mT/\eta}$.
Define the margin $M_t(x,\ell)=\sum_{j=1}^t
(2\one\bc{f_j(x)=\ell}-1)$, with $M_0=0$, and let
\[
 V_t(s)=\min\bc{1,(127/128)^{T-t}(7/8)^s},
 \qquad w_t(s)=V_t(s-1)-V_t(s+1)\in[0,1].
\]
In round $t$, accept each example $(x,\ell)$ from a fresh batch of $N$
examples independently with probability $w_t(M_{t-1}(x,\ell))$.
Keep the first at most $m$ accepted examples, filling any missing
positions with a fixed labeled point. Uniformly permute the resulting
$m$ examples and run $\cA$ to obtain $f_t$.
Return the pointwise majority of $f_1,\ldots,f_T$.

\emph{Privacy.}
Fix the earlier outputs and couple the acceptance coins on neighboring
batches position by position. Replacing one example changes the retained
multiset, including filling examples, by at most one example.
Uniform permutation couples the inputs to $\cA$ to differ in at most
one position, so this round is $(\epsilon,\delta)$-DP.
A change in batch $t$ leaves the earlier outputs unchanged in
distribution. Later rounds use unchanged batches and are postprocessing
of these outputs and $f_t$. Thus the full sequence, and its majority,
are $(\epsilon,\delta)$-DP.

\emph{Accuracy.}
Fix a realizable labeled distribution $P$, and let
$\overline V_t=\bE_P[V_t(M_t(x,\ell))]$.
For round $t$, condition on the earlier outputs and write
$a_t=\bE_P[w_t(M_{t-1}(x,\ell))]$.
If $a_t>0$, the distribution $\mu_t$ of an accepted example is realizable
by the same target as $P$. The number of acceptances is
$K_t\sim\operatorname{Bin}(N,a_t)$. Conditional on the acceptance
decisions, the accepted examples are independent samples from $\mu_t$,
so conditional on $K_t\geq m$ the first $m$ of them are as well. Hence
$\bE[\err_{\mu_t}(f_t)]\leq7/16+\Pr[K_t<m]$.
Adding one independent trial and using exchangeability gives
\[
 a_t\Pr[K_t<m]
 =\frac{\bE[B_t\one\bc{B_t\leq m}]}{N+1}
 \leq\frac{m}{N+1},
 \qquad B_t\sim\operatorname{Bin}(N+1,a_t).
\]
Since $(9/16)(7/8)+(7/16)(8/7)=127/128$, we have
\[
 V_t(s+1)+\frac{7}{16}w_t(s)
 =\frac{9}{16}V_t(s+1)+\frac{7}{16}V_t(s-1)
 \leq V_{t-1}(s).
\]
Indeed, the middle expression is at most $1$ and at most the expression
defining $V_{t-1}(s)$ before taking the minimum.
With $s=M_{t-1}(x,\ell)$, a correct prediction of $f_t$ increases
the margin by one and an incorrect prediction decreases it by one, so
$V_t(M_t(x,\ell))=V_t(s+1)+w_t(s)\one\bc{f_t(x)\neq\ell}$.
Thus the conditional expected increase in $\overline V_t$ is at most
$a_t(\bE[\err_{\mu_t}(f_t)]-7/16)\leq m/(N+1)$.
If $a_t=0$, the weights vanish $P$-almost surely and the increase
is nonpositive.

Initially, $\overline V_0=(127/128)^T\leq\eta/96$, and $V_T(s)=1$ for
$s\leq0$, so $\overline V_T$ bounds the error of the final majority.
Taking expectations and summing over the rounds gives
\[
 \bE[\err_P(\operatorname{Maj}(f_1,\ldots,f_T))]
 \leq\frac{\eta}{96}+\frac{Tm}{N+1}\leq\frac{\eta}{48}.
\]
Markov's inequality gives error at most $\eta/12$ with probability
at least $3/4$. The sample size is $TN=\bigO{(m/\eta)\log^2(2/\eta)}$.
Finally, the confidence amplification part of \Cref{lem:resample}
gives error $\eta$ and failure probability $\gamma$, using
$\bigO{\log(2/\gamma)}$ independent copies and preserving privacy.
\end{proof}

\begin{proof}[Proof of~\Cref{cor:transfer-parameters}]
Apply \Cref{lem:accuracy} to the learner in \Cref{lem:transfer}.
It uses $r\leq C(s_i/\eta)\log^2(2/\eta)\log(2/\gamma)$
examples and remains $(0.1,\delta/\epsilon)$-DP.
From $\ceil{ r/\epsilon}$ examples, select $r$ uniformly
without replacement and in uniformly random order with probability $p\leq\epsilon$. Privacy amplification \cite{balle2018privacy} under
replacement adjacency gives parameters
$\log\br{1+p(e^{0.1}-1)}\leq\epsilon$ and $p\delta/\epsilon\leq\delta$.
Conditional on any choice of distinct sample indices, the selected
examples have the original product distribution. They therefore remain
independent samples after averaging over the indices, so utility is
unchanged. Finally, setting all predictions outside $\cX_i$ to $o_i$
is postprocessing and cannot increase the error on a realizable
distribution.
\end{proof}

%% file: appendix/interval_upper_bound.tex
\section{Private Learning of Unions of Intervals}
\label{app:interval-upper}

Recall that $\cI_{N,k}$ is the class of unions of at most $k$
intervals on $\bs{N}$, where $N\geq2k$ and $k\geq1$. We privately
release threshold queries on the labeled sample, use them to estimate
the empirical error of each hypothesis, and return a minimizer.
Write $\ell_M=\logstar\br{\max\bc{2,M}}$.

\begin{lemma}[Threshold query release]
\label{lem:interval-threshold-release}
Let $M\geq2$, $0<\epsilon\leq1$ and $0<\delta,\eta,\zeta\leq1/4$.
For a database $T=\br{z_1,\ldots,z_n}\in\bs{M}^n$, define
$F_T(t)=\frac1n\sum_{i=1}^n\one\bc{z_i\leq t}$ for
$t\in\bc{0,\ldots,M}$. There is an $(\epsilon,\delta)$-DP algorithm
that, with probability at least $1-\zeta$, returns answers satisfying
$\max_{0\leq t\leq M}\abs{\widehat F(t)-F_T(t)}\leq\eta$, provided
\begin{equation}
\label{eq:interval-threshold-release-size}
 n\geq\frac{C}{\epsilon\eta}\bs{
   \ell_M\log^2\br{\frac{C\ell_M}{\eta\zeta\delta}}
   +\log\br{\frac C\zeta}\log^{5/2}\br{\frac C\eta}
 },
\end{equation}
where $C\geq2$ is universal.
\end{lemma}

\begin{proof}
The interior-point algorithm of \citet{CohenEtAl2023} uses
$m=\bigO{\frac{\ell_M}{\epszero}
\log^2\br{\frac{C\ell_M}{\gamma\delta_0}}}$ examples for privacy
$(\epszero,\delta_0)$ and failure probability $\gamma$.
Apply \citet[Theorem~4.6(2)]{BunEtAl2015} with
$\epszero=\epsilon/5$, $\delta_0=\delta/\br{1+e^{\epszero}}$ and
$\gamma=\eta\zeta/24$. This releases threshold queries with accuracy
$\eta$, failure probability $\zeta$ and privacy $(\epsilon,\delta)$,
using
\[
 \max\bc{
    \frac{6m}{\eta},\,
    \frac{25\log\br{24/\zeta}\log^{5/2}\br{6/\eta}}{\eta\epszero}
 }
\]
examples. Finally, substitution gives \eqref{eq:interval-threshold-release-size}.
\end{proof}

\begin{theorem}[Private learner for unions of intervals]
\label{thm:interval-upper}
Let $N\geq2k$, $k\geq1$, $0<\epsilon\leq1$ and
$0<\alpha,\beta,\delta\leq1/4$. There is a proper, realizable,
$(\epsilon,\delta)$-DP, $(\alpha,\beta)$-PAC learner for $\cI_{N,k}$
with sample complexity at most
\begin{equation}
\label{eq:interval-upper}
 \frac{Ck}{\alpha\epsilon}
 \bs{
   \ell_N\log^2\br{\frac{Ck\ell_N}{\alpha\beta\delta}}
   +\log\br{\frac C\beta}\log^{5/2}\br{\frac{Ck}{\alpha}}
 },
\end{equation}
where $C\geq2$ is universal.
\end{theorem}

\begin{proof}
Given $S=\br{(x_1,y_1),\ldots,(x_n,y_n)}$, encode each example as
$z_i=x_i+Ny_i\in\bs{2N}$, so negative examples precede positive ones.
Let $T=\br{z_1,\ldots,z_n}$ and $F=F_T$. Apply
\Cref{lem:interval-threshold-release} to $T$ with privacy
$(\epsilon,\delta)$, failure probability $\zeta=\beta/2$ and accuracy
$\eta=\alpha/\br{8(4k+1)}$. For $h\in\cI_{N,k}$, let $\bs{a_1,b_1},\ldots,\bs{a_r,b_r}$ be its
maximal positive intervals, with $r\leq k$. Define
\begin{equation}
\label{eq:interval-released-error}
 \widehat L(h)=1-\widehat F(N)
 +\sum_{j=1}^r\bs{
    \widehat F(b_j)-\widehat F(a_j-1)
    -\widehat F(N+b_j)+\widehat F(N+a_j-1)
 }.
\end{equation}
Return $\widehat h\in\argmin_{h\in\cI_{N,k}}\widehat L(h)$. A
minimizer exists since $\cI_{N,k}$ is finite. Replacing one example
changes only one entry of $T$, and then choosing $\widehat h$ is
postprocessing of the released answers. Thus the learner is
$(\epsilon,\delta)$-DP on every database, including nonrealizable
ones. Next, to bound the error, note that $1-F(N)$ is the fraction of
positive examples. Predicting one on $\bs{a,b}$ adds the negative mass
$F(b)-F(a-1)$ and removes the positive mass $F(N+b)-F(N+a-1)$. Hence
 \[\err_S(h)=1-F(N)
 +\sum_{j=1}^r\bs{
    F(b_j)-F(a_j-1)-F(N+b_j)+F(N+a_j-1)
 }.\]
With probability at least $1-\beta/2$, every released answer is within
$\eta$ of its true value. On this event, every $h\in\cI_{N,k}$ satisfies
\[
 \abs{\widehat L(h)-\err_S(h)}
 \leq\br{4r+1}\eta\leq\br{4k+1}\eta=\alpha/8.
\]
If $S$ is realizable by $h^\star\in\cI_{N,k}$, then
\[
 \err_S(\widehat h)
 \leq\widehat L(\widehat h)+\alpha/8
 \leq\widehat L(h^\star)+\alpha/8
 \leq\err_S(h^\star)+\alpha/4
 =\alpha/4.
\]

Fix a realizable distribution $\mu$ and let $S\sim\mu^n$.
Since $v=\VC\br{\cI_{N,k}}=2k$, the standard relative VC bound
\citep[Theorem~A3.1]{BlumerEtAl1989} gives, with probability at least $1-\beta/2$,
$\err_\mu(h)\leq\alpha$ for every $h\in\cI_{N,k}$ with
$\err_S(h)\leq\alpha/4$,
provided
\begin{equation}
\label{eq:interval-generalization-size}
 n\geq\frac{C_1}{\alpha}
       \bs{v\log\br{\frac2\alpha}+\log\br{\frac2\beta}},
\end{equation}
where $C_1$ is universal.

A union bound gives $\err_\mu(\widehat h)\leq\alpha$ with probability
at least $1-\beta$. Substituting our parameters into
\eqref{eq:interval-threshold-release-size} and using
$\ell_{2N}\leq\ell_N+1\leq2\ell_N$ gives \eqref{eq:interval-upper}.
This also dominates \eqref{eq:interval-generalization-size} since
$k\geq1$ and $\epsilon\leq1$, after increasing $C$.
\end{proof}